\def\ArxivPreprint{}
\PassOptionsToPackage{table}{xcolor}
\documentclass{article}

\PassOptionsToPackage{table}{xcolor}
\ifdefined\ArxivPreprint\else
\documentclass{article}
\fi
\usepackage[T1]{fontenc}

\usepackage{microtype}
\usepackage{graphicx}
\usepackage{subcaption}
\usepackage{booktabs} 

\usepackage{hyperref}
\usepackage{xurl} 

\ifdefined\ArxivPreprint
  \usepackage[preprint]{icml2026}
\else
  \usepackage{icml2026}
\fi

\usepackage{amsmath}
\usepackage{amssymb}
\usepackage{mathtools}
\usepackage{amsthm}
\usepackage{hyperref}   
\usepackage{fontawesome5}

\usepackage[capitalize,noabbrev]{cleveref}

\theoremstyle{plain}

\theoremstyle{definition}

\theoremstyle{remark}

\usepackage[textsize=tiny]{todonotes}

\usepackage{xcolor}
\usepackage{tabularx}
\usepackage{wrapfig}
\definecolor{tabhead}{HTML}{E3E9F0}
\definecolor{tabrow}{HTML}{F2F5F8}
\definecolor{tabrule}{HTML}{34475A}
\arrayrulecolor{tabrule}
\newcolumntype{R}{>{\raggedleft\arraybackslash}X}
\newcolumntype{L}{>{\raggedright\arraybackslash}X}
\newcolumntype{C}{>{\centering\arraybackslash}X}

\icmltitlerunning{Understanding Mixture-of-Experts Language Models Beyond Expert Boundaries}

\begin{document}

\twocolumn[
  \icmltitle{Understanding Mixture-of-Experts Language Models Beyond Expert Boundaries}



  \icmlsetsymbol{equal}{*}
  \icmlsetsymbol{corr}{$\dagger$}

  \begin{icmlauthorlist}
    \icmlauthor{Zhenhong Zhou}{ntu,equal}
    \icmlauthor{Mao Zheng}{tencent,equal}
    \icmlauthor{Mingyang Song}{tencent,equal}
    \icmlauthor{Qiankun Li}{ntu}
    \icmlauthor{Kun Wang}{ntu}
    \icmlauthor{Junfeng Fang}{nus,corr}
  \end{icmlauthorlist}

  \icmlaffiliation{ntu}{Nanyang Technological University}
  \icmlaffiliation{tencent}{Foundation Model Department, Tencent}
  \icmlaffiliation{nus}{National University of Singapore}

  \icmlcorrespondingauthor{Junfeng Fang}{fangjf@nus.edu.sg}

  \icmlkeywords{Machine Learning, ICML}

  \vskip 0.3in
]



\newcommand{\githubbtn}[1]{%
  \href{#1}{%
    \tikz[baseline=-0.5ex]{\node[draw, rounded corners=2pt, inner sep=2pt, anchor=base] {\faGithub\ GitHub};}%
  }%
}

\printAffiliationsAndNotice{\icmlEqualContribution
\textsuperscript{$\dagger$}Corresponding author.
}

\begin{abstract}
In Mixture-of-Experts (MoE) models, the nature of expert specialization remains controversial.
Reports of domain-associated expert preferences contrast with analyses attributing routing primarily to token identity or syntactic patterns.
To clarify these differing accounts, we look beyond expert-level routing to investigate the neuron-level organization of domain-specialized computation using Routing-Aware Magnitude Profiling (RAMP), a measure developed for this analysis.
Using mixed-domain calibration data, we identify domain-specialized neuron sets with 58--74\% overlap across multi-source samples of the same domain, compared with 15--24\% across domains.
To understand how this organization evolves during training, we trace neuron importance throughout pre-training and post-training across three fully open MoE families.
Neuron importance continues to reorganize between pre-training checkpoints despite comparatively stable expert-level importance rankings, while post-training largely preserves neuron rankings.
Building on these neuron-level perspectives, we further introduce RAMP-based Pruning (RAMPr) to select neurons across experts for balanced multi-domain capability preservation.
At 75\% neuron retention, RAMPr preserves average 91--98\% of original full performance across six models.
At 50\% retention, our pruning method can reduce resident GPU memory by 36--43\% and improves batched prefill throughput by 17--39\% effectively.
\githubbtn{https://github.com/ydyjya/Neuron-Level-Specialization-in-MoE}
\end{abstract}

\begin{figure*}[t]
\centering
\includegraphics[width=\textwidth]{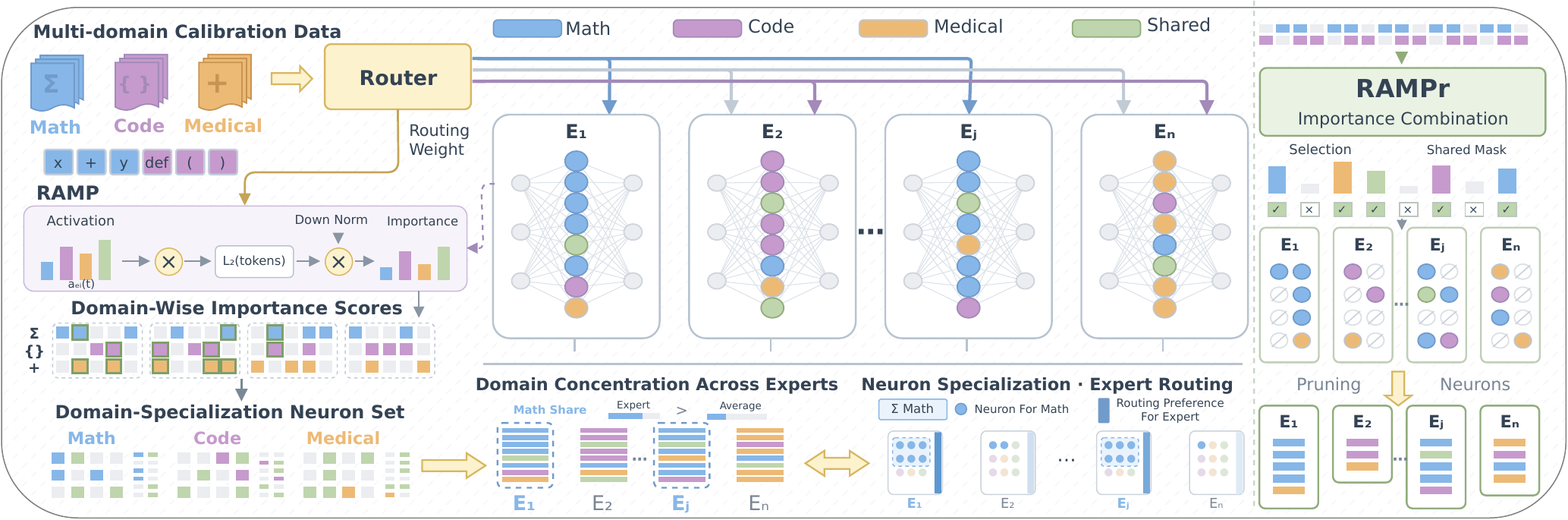}
\caption{
Routing-weighted activation magnitudes and output projection norms quantify domain-wise neuron importance, revealing domain-specialized sets that span mixed experts. 
Their enrichment correlates with domain routing preferences. 
Aggregating these important scores guides multi-domain neuron pruning across experts, retaining different numbers of neurons in each expert.}
\vspace{-15pt}
\label{fig:overview}
\end{figure*}

\section{Introduction}
\label{sec:introduction}

Mixture-of-Experts (MoE) architectures increase model capacity through a collection of experts separately, only a subset of which processes each token \citep{openai2025gptoss120bgptoss20bmodel, deepseekai2026deepseekv4highlyefficientmilliontoken, qiu2026designqwen38nextarchitectureevaluation}.
This design naturally fosters the intuition that each expert serves as a unit of computational specialization in MoE models.
This intuition is reflected in work that treats expert specialization as a defining property of MoE architectures and seeks to induce or identify distinct expert functions \citep{dai-etal-2024-deepseekmoe, muennighoff2025olmoe, pmlr-v306-herbst26a}. 
However, a contrasting line of work questions whether routing preferences alone provide reliable evidence of functional specialization, showing that routing decisions can be strongly tied to token identity and hidden-state geometry rather than directly revealing domain-level expertise \citep{pmlr-v235-xue24c,wang2026mythexpertspecializationmoes}. 
Furthermore, recent interpretability work suggests that an expert's routing profile may reflect a combination of fine-grained features rather than specialization in a single domain\citep{pmlr-v306-lasy26a}.
This raises an open research question: \emph{to what extent is computational specialization in MoE models aligned with input domains?}

Recent studies suggest that neurons within MoE experts contribute unequally to model computation, motivating a finer-grained view of MoE specialization \citep{cheng2025mone,he2026less}.
We investigate how domain specialization is organized within MoE models using Routing-Aware Magnitude Profiling (RAMP), which measures each neuron's contribution magnitude weighted by its expert's routing coefficient.
Across six models, the resulting neuron sets exhibit domain-dependent membership, with 58--74\% overlap within domains and 15--24\% across domains at 6.25\% selection. 
Examining how these neurons are distributed across experts reveals that 62.64--82.50\% of domain-exclusive neurons reside in experts where they are disproportionately represented. 
We further find that this enrichment correlates positively with expert-level routing preferences across all models, with median Spearman correlations of 0.42--0.71, providing a neuron-level account of the domain-specific routing patterns reported in prior work \citep{hu2026whatgetsactivated}.
Meanwhile, the distribution of domain-related neurons across mixed experts explains why expert-level routing profiles need not reveal a clear domain-level organization of computation \citep{jiang2024mixtral}.
Building on prior findings, our analysis reconciles the debate over expert specialization by explaining how domain-associated routing can coexist with the absence of clear expert-level domain boundaries.

To understand how this neuron-level organization develops, we investigate its evolution during training, revisiting existing observations of early expert routing stabilization during pre-training \citep{pmlr-v235-xue24c}.
We track neuron importance alongside expert-level statistics in two parallel analyses, one following the OLMoE full training trajectory \citep{muennighoff2025olmoe} and the other combining training stages in OpenMoE \citep{pmlr-v235-xue24c} and Instella \citep{liu2026instellamoe}.
Comparisons between adjacent checkpoints reveal continued neuron-level reorganization during pre-training even when expert-level importance patterns appear stable.
Comparing adjacent-checkpoint changes within pre-training with those within post-training yields median neuron-importance rank correlations of 0.842--0.939 and 0.997--0.998, respectively, extending prior studies of module and neuron stability \citep{zhang2023emergent, wang2026deconstructing, qiu2025demons}.

Beyond explaining specialization, this neuron-level view provides a basis for optimizing computational capacity in MoE.
Domain-wise profiling reveals the relative importance of neurons across experts, motivating RAMP-based Pruning (RAMPr).
The observed cross-domain sharing and domain-dependent importance motivate preserving both broadly important neurons and those important to particular domains.
These considerations lead to layer-wise pruning by cumulative importance (RAMPr-L) and domain-balanced pruning by normalized importance (RAMPr-D), each retaining a shared neuron set across domains.
At 75\% retention on GLM-4.7-Flash, RAMPr-D offers a more favorable balance between knowledge and generation performance than the expert-pruning baseline REAP, improving the eleven-task macro-average over it by 2.91 percentage points.
Pruning at 50\% retention level reduces resident GPU memory by 36--43\% and improves batched prefill throughput by 17--39\%.

Our contributions are summarized as follows.
\begin{list}{}{
  \setlength{\leftmargin}{0pt}
  \setlength{\labelwidth}{1.25em}
  \setlength{\labelsep}{0.4em}
  \setlength{\itemindent}{1.65em}
  \setlength{\listparindent}{0pt}
  \setlength{\topsep}{2pt}
  \setlength{\parsep}{0pt}
  \renewcommand{\makelabel}[1]{#1\hfil}
}

\item[\textcircled{\scriptsize\textit{1}}]
\textbf{\underline{Neuron-level specialization.}}
We find there are domain-specialized neuron sets in MoE spanning mixed experts.
This finding reconciles the debate over expert specialization by locating domain-related organization at the neuron level.

\item[\textcircled{\scriptsize\textit{2}}]
\textbf{\underline{Training-time evolution.}}
We reveal continued neuron-level reorganization during pre-training even when expert-level patterns appear stable.
Post-training largely preserves neuron importance rankings.
Expert-level stability provides an incomplete picture of how specialization evolves.

\item[\textcircled{\scriptsize\textit{3}}]
\textbf{\underline{Specialization-informed pruning.}}
Domain specialized neurons provide a basis for selecting what to retain across expert boundaries, motivating RAMP-based Pruning.
The results show that this neuron-level perspective helps preserve capabilities while reducing GPU memory usage.
\end{list}

\section{Related Work}
\label{sec:related-work}

\subsection{Expert Specialization in MoE Models}
Architectural approaches organize conditional computation through expert granularity and token allocation, relating the partition of model capacity to computational cost and language-model performance \citep{shazeer2017outrageously,zhou2022expertchoice,ludziejewski2024finegrained}.
Analyses of the resulting specialization connect expert responses to input patterns and routing geometry, revealing both architecture-dependent differences in specialization and associations between router scores and internal activations \citep{zoph2022stmoe,ahrac2026routers}.
Training-oriented approaches address the interaction between routing and expert learning through stabilized assignments, router-logit regularization and component-specific optimization schedules \citep{dai2022stablemoe,ludziejewski2025decoupled}.
Compression studies use routed contributions to identify dispensable computation at different granularities, from whole-expert removal to connection-level sparsification within experts \citep{li2024mcsmoe,chen2025hcsmoe,jaiswal2025fantastic,lasby2026reap,song2026razor,dong2025easyep,jha2026halfexperts,xie2024moepruner}.
Another direction reorganizes or selects neuron-level computation through expert reconstruction, distillation and depth-dependent retention, extending compression beyond the expert selection\citep{lee2025stun,zhou2025dern,li2025slimmoe,li2026heapr,xu2026camera,he2026tenp}.

\subsection{Neuron and Feature Interpretability}
Neuron-level interpretations connect activations and parameters to input patterns and factual associations through memory-based analyses, attribution and targeted editing \citep{geva2021transformer,dai2022knowledge,meng2022locating}.
This view of feed-forward computation supports conditional execution by grouping neurons into routed modules, linking analyses of their input responses with architectural decomposition \citep{zhang2022moefication}.
Sparse dictionary learning instead studies features in a learned representational basis, decomposing activations into interpretable directions that can span multiple original neuron coordinates \citep{bricken2023monosemanticity,cunningham2023sparse,templeton2024scaling}.
Circuit-level approaches examine interactions among components through interventions and automated pathway discovery, identifying the computations supporting specific model behaviors \citep{wang2023interpretability,conmy2023automated}.

\section{Domain-Specialized Neuron Organization}
\label{sec:neuron-organization}

In MoE, experts can participate in multiple domains \citep{muennighoff2025olmoe, hu2026whatgetsactivated}, while other studies associate routing with token identity and hidden-state geometry, questioning whether these preferences reflect domain specialization \citep{pmlr-v235-xue24c,wang2026mythexpertspecializationmoes}.
In this section, we address this tension through a neuron-level analysis of domain specialization within and across experts.

\begin{figure*}[t]
\centering
\includegraphics[width=\textwidth]{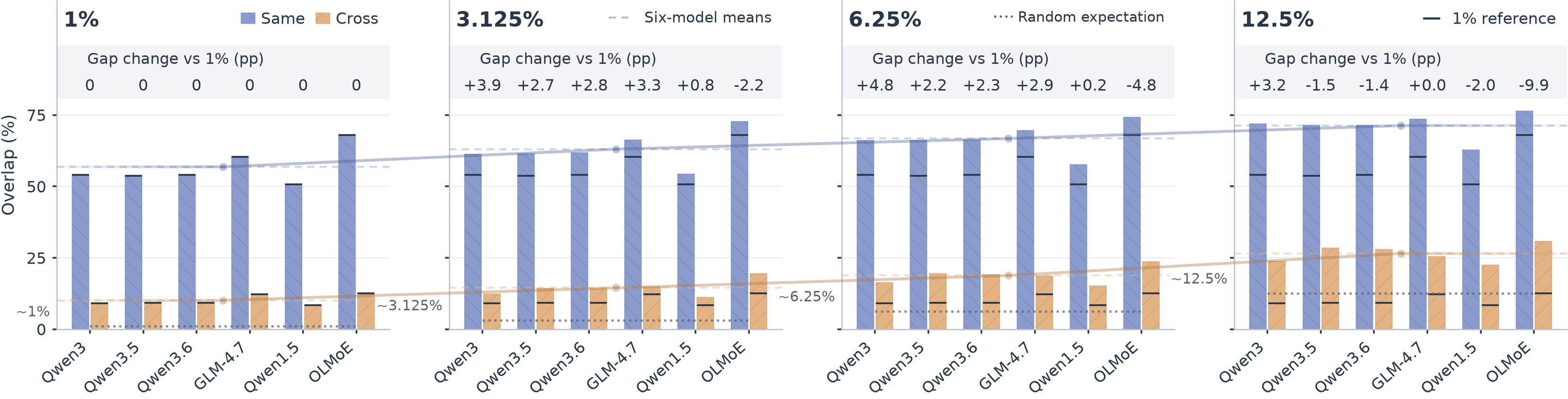}
\caption{Neuron-set overlap at four selection rates across six models.
Blue and orange bars show same-domain and cross-domain overlap.
Dashed lines and connected center points denote six-model means, while gray dotted lines mark random-set expectations.
Dark ticks show each model's overlap at 1\% selection, and upper strips give the change in the same--cross gap relative to that setting, in percentage points.}
\label{fig:discovery-replication}
\vspace{-15pt}
\end{figure*}

\subsection{Models and Domain Comparisons.}
\label{sec:discovery-setup}

The analysis covers six models, Qwen3-30B-A3B \citep{qwen2025qwen3}, Qwen3.5-35B-A3B \citep{qwen2026qwen35}, Qwen3.6-35B-A3B \citep{qwen2026qwen36}, GLM-4.7-Flash \citep{glm2025glm45}, Qwen1.5-MoE-A2.7B-Chat \citep{qwen2024qwen15moe} and OLMoE-1B-7B-0125-Instruct \citep{muennighoff2025olmoe}, and more details are in Appendix~\ref{app:discovery-protocol}.
To explore domain-related neuron organization across varied inputs, we draw calibration text from multiple datasets covering chat, mathematics, medicine, Chinese text and code.
Generation and multiple-choice data provide different task formats, while code is represented by generation tasks, yielding nine groups, and their source composition is detailed in Appendix~\ref{app:discovery-protocol}.
Each group contains 128 documents, divided equally into discovery and validation sets.

\subsection{Routing-Aware Magnitude Profiling}
\label{sec:ramp}
In MoE, a neuron's activation magnitude alone does not fully capture its contribution magnitude, which also depends on the norm of the corresponding down-projection column and the expert's routing weight.
Within an MoE layer with $E$ routed experts of width $N$, expert $e$ maps hidden state $x(t)$ to the gated activation vector $a^e(t)$ as
\begin{equation}
a^e(t)=\operatorname{SiLU}\!\left(W_{\mathrm{gate}}^e x(t)\right)
\odot W_{\mathrm{up}}^e x(t).
\end{equation}
Let $w_i^{e,\mathrm{down}}$ denote the down-projection column for neuron $i$ in expert $e$.
The routed branch output is
\begin{equation}
y^{\mathrm{routed}}(t)=\sum_{e=1}^{E}p^e(t)
\sum_{i=1}^{N}a_i^e(t)w_i^{e,\mathrm{down}},
\label{eq:routed-output}
\end{equation}
where $p^e(t)$ is the routing weight assigned to expert $e$ for token $t$, set to zero when the expert is not selected.

Equation~\eqref{eq:routed-output} assigns neuron $i$ in expert $e$ the routed contribution vector $c_i^e(t)=p^e(t)a_i^e(t)w_i^{e,\mathrm{down}}$.
For calibration text $D_d$ from domain $d$, Routing-Aware Magnitude Profiling (RAMP) assigns neuron $i$ in expert $e$ the importance score
\begin{equation}
 I_i^e(D_d)=\left(\sum_{x\in D_d}\sum_{t\in x}
 [p^e(t)a_i^e(t)]^2\right)^{1/2}
 \left\|w_i^{e,\mathrm{down}}\right\|_2.
 \label{eq:ramp-weighted}
\end{equation}
This is the square root of the summed squared norms of the neuron's routed contribution vectors across tokens.
Accumulation and tie-breaking details appear in Appendix \ref{app:discovery-protocol}.

\subsection{Domain-Specialized Neuron}
\label{sec:discovery-replication}

Using the importance score $I_i^e(D_d)$, we identify the neurons most relevant to each domain by selecting the top $r$ fraction of routed expert--neuron coordinates within each layer.
We use $r\in\{1\%,3.125\%,6.25\%,12.5\%\}$ to form domain-specific neuron sets, allowing a neuron to belong to multiple sets.
We compare these sets across samples from the same domain and across different domains to examine how their membership varies with the input domain.

\textbf{Domain-dependent membership.}
To explore whether neuron sets are domain-specialized, we compare their overlap across samples from the same domain and across different domains.
For discovery and validation neuron sets $T_d^{\mathrm{disc}}$ and $T_d^{\mathrm{val}}$, each of size $k$, overlap is respectively defined as
\begin{equation}
\begin{aligned}
 O_{\mathrm{same}}(d)&=\frac{|T_d^{\mathrm{disc}}\cap T_d^{\mathrm{val}}|}{k},\\
 O_{\mathrm{cross}}(d,c)&=\frac{|T_d^{\mathrm{disc}}\cap T_c^{\mathrm{disc}}|}{k}.
\end{aligned}
 \label{eq:discovery-overlap}
\end{equation}
Higher same-domain than cross-domain overlap would indicate that the selected neurons are associated with domain identity beyond individual text samples.
Independent random selection gives an expected overlap of $k/(E\times N)\approx r$, providing a reference for assessing sharing across domains.
At 1\% selection, with RAMP, the overlap is 50.79--68.07\% within domains and 8.47--12.64\% across domains (\cref{fig:discovery-replication}).
As selection expands to 6.25\%, within-domain and cross-domain overlaps reach 57.77--74.33\% and 15.23--23.68\%, respectively.
Same-domain overlap exceeds cross-domain overlap in all 24 model--budget configurations, supporting domain specialization at the neuron level.
Full analysis are in Appendix~\ref{app:discovery-budgets}.
Ablating a domain's top neurons further shows that the model relies on them (Appendix~\ref{app:domain-ablation}).

\textbf{Neuron specialization for task formats.}
Task format may also constitute an additional dimension of neuron specialization. 
To test this intuition, we compare neuron sets identified from generation and multiple-choice inputs within the same content or language categories.
Averaged over the four selection rates and six models, the within-bucket discovery--validation overlap is 67.31\% for multiple-choice and 60.93\% for generation, with model-wise averages across the four rates ranging from 58.51--73.81\% and 52.09--70.36\%, respectively.
Cross-format overlap between the paired discovery sets averages 35.08\%, ranging from 26.92--43.83\% across models.
Within-bucket overlap exceeds cross-format overlap for both formats in all 24 model--budget configurations, suggesting that neuron-level specialization reflects both content domain and task format.
We further examine whether individual neurons retain consistent domain preferences across different text samples.
Most neurons with a clear domain preference retain that preference on validation text, complementing the set-level comparisons above.
Further details are in Appendix~\ref{app:cross-format} and Appendix~\ref{app:identity}.

\subsection{Neuron Organization Across Expert Boundaries}
\label{sec:discovery-spatial}
The identified domain neuron sets offer a finer-grained perspective on specialization.
We next discuss how these neurons are distributed across experts to relate neuron-level specialization to the expert specialization discussed in prior work \citep{muennighoff2025olmoe,pmlr-v306-herbst26a}.
For each domain and layer, we rank experts by their counts $c_e^{\mathrm{disc}}$ of selected neurons in the discovery top set $T_d^{\mathrm{disc}}$, then measure how much of that entire set the highest-count experts cover.
The fraction covered by the top 10\% of experts directly indicates how concentrated the domain's selected neurons are, compare with approximately 10\% under uniform allocation.
At 1\% selection, the top 10\% of experts cover 53.27--83.70\% of each domain's selected neurons across the six models.
Coverage decreases as the selection rate grows but remains 35.98--62.27\% at 12.5\%, so the selected neurons concentrate in a small subset of experts.
Appendix~\ref{app:spatial} reports the more detailed figures, tables and analysis.

This concentration alone may produce part of the same-domain overlap, because $T_d^{\mathrm{disc}}$ and $T_d^{\mathrm{val}}$ share neurons whenever they draw on the same experts.
To remove this effect, we fix the number of neurons each expert contributes to $T_d^{\mathrm{disc}}$ and $T_d^{\mathrm{val}}$, denoted $c_e^{\mathrm{disc}}$ and $c_e^{\mathrm{val}}$, and draw neurons uniformly, which gives the expected baseline overlap
\begin{equation}
O_{\mathrm{count}}=\frac1{k}\sum_e\frac{c_e^{\mathrm{disc}}c_e^{\mathrm{val}}}{N}.
\label{eq:expert-count-null}
\end{equation}
Observed overlap exceeds this baseline in every layer--bucket cell at both budgets.
At 6.25\% selection, the model-average excess ranges from 22.39 to 40.11 percentage points (\cref{fig:spatial-reference}b), and more statistics are in Appendix~\ref{app:spatial}.
The discovery and validation sets therefore agree not only on which experts hold the selected neurons but also on which neurons are selected within those experts.

\begin{figure*}[t]
\centering
\includegraphics[width=\textwidth]{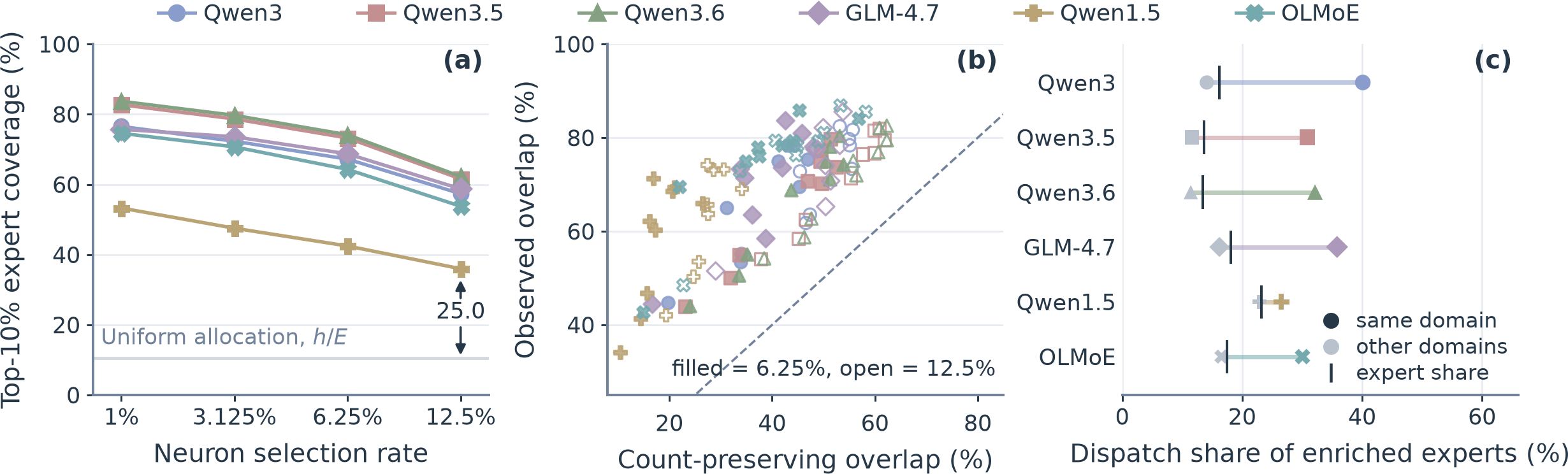}
\caption{
(a) Share of each domain's selected neurons held by the top $\lceil0.1E\rceil$ experts, with the gray line marking uniform allocation and the arrow giving the gap in percentage points.
(b) Observed overlap against the count-preserving expectation in \cref{eq:expert-count-null}, with one marker per model, domain and selection rate, and the dashed line marking parity.
(c) Validation dispatch share of each domain's enriched experts, from the same domain in color and from other domains in gray, with black ticks marking their share of all experts.}
\label{fig:spatial-reference}
\end{figure*}

\subsection{Domain Enrichment Within Mixed Experts}
\label{sec:exclusive-enrichment}
The previous coverage is measured for each domain, so it cannot tell whether each domain concentrates in its own experts or all domains concentrate in the same heavily selected experts.
Telling them apart requires comparing domains within each expert, and neurons selected for several domains would perturb this comparison, so we use the discovery neurons selected for exactly one domain,
\begin{equation}
 U_d=T_d^{\mathrm{disc}}\setminus\bigcup_{c\ne d}T_c^{\mathrm{disc}}.
\end{equation}
At 6.25\% selection, exclusive neurons constitute 27.75--39.51\% of their parent top sets on average.

If neuron-level specialization merely followed experts that are heavily selected overall, each domain's share within an expert would match its share of all exclusive neurons.
We test this alternative with the enrichment of domain $d$ in expert $e$, the ratio of these two shares,
\begin{equation}
 F_d^e=\frac{x_d^e/s^e}{m_d/S},
 \label{eq:composition-enrichment}
\end{equation}
where $x_d^e$ counts $d$'s exclusive neurons in $e$, $s^e=\sum_d x_d^e$, $m_d=\sum_e x_d^e$ and $S=\sum_e s^e$.
A value of $F_d^e=1$ matches this alternative, while $F_d^e>1$ means that $d$ takes a larger share of the expert than of the pool.
Since an expert's domain shares sum to one, no expert can have $F_d^e>1$ for every domain, so a high overall count alone does not produce enrichment.
We call an expert enriched for $d$ when $F_d^e>1$ and the BH-adjusted hypergeometric $q_d^e<0.05$, with reference count $m_ds^e/S$.
Enriched experts appear in every domain--layer unit at all four selection rates (Appendix~\ref{app:exclusive-enrichment}).
They comprise only 13.39--23.26\% of experts on average at 6.25\% selection yet carry a share $M$ of 66.40--83.40\% of the domain's exclusive neurons (\cref{tab:exclusive-enrichment}), and $M$ stays between 58.18\% and 85.89\% across the four rates.
Therefore, the previous concentration does reflect experts where each domain takes a larger share than it does overall.

If neuron-level specialization relates to the expert-level domain preferences reported in prior work, enrichment measured from neuron composition alone may correspond to routing, which we test on validation text.
Let $C_d^e$ count dispatched token--expert edges and define
\begin{equation}
\begin{aligned}
 r_d^e=\frac{C_d^e}{\sum_{e'}C_d^{e'}}, \quad
 b_d^e=\frac18\sum_{c\ne d}r_c^e, \quad
 \Delta r_d^e=r_d^e-b_d^e.
\end{aligned}
\label{eq:domain-routing-preference}
\end{equation}
$\Delta r_d^e$ measures an expert's preference for $d$ beyond its average usage in the other domains, matching the relative form of $F$.
Within each domain and layer, discovery $F$ and validation $\Delta r$ are positively correlated across experts at 6.25\% selection, with median Spearman correlations of 0.3682--0.6668 (\cref{tab:exclusive-enrichment}).
Taken as a group, the enriched experts $H_d=\{e:F_d^e>1,\ q_d^e<0.05\}$ fixed on discovery text also receive a validation dispatch share 3.65--26.01 pp higher for their own domain than for other domains (\cref{fig:spatial-reference}c).
Both associations hold at all four selection rates (Appendix~\ref{app:router-link}).
To ensure that these patterns do not follow merely from how often each expert is used, we normalize every neuron's score by its expert's dispatch count, under which the same-domain advantage, the concentration of selected neurons and the positive association with routing all persist in every model, with median correlations of 0.16--0.38 (Appendix~\ref{app:dispatch-normalized}).

\begin{table}[t]
\caption{Enrichment and validation routing at 6.25\% selection.
$|H_d|/E$ is the share of experts enriched for a domain, $M$ is their share of the domain's exclusive neurons, $\widetilde\rho$ is the median Spearman correlation between $F$ and $\Delta r$, and $\Delta_H$ is the target-minus-other-domain dispatch-share difference, with all but $\widetilde\rho$ in percent.}
\label{tab:exclusive-enrichment}
\centering\small
\setlength{\tabcolsep}{3pt}
\rowcolors{2}{tabrow}{}
\begin{tabularx}{\linewidth}{lRRRR}
\toprule
\rowcolor{tabhead}Model & $|H_d|/E$ & $M$ & $\widetilde\rho$ & $\Delta_H$ \\
\midrule
Qwen3 &16.18&83.11&0.6668&26.01\\
Qwen3.5 &13.61&83.02&0.5671&19.19\\
Qwen3.6 &13.39&83.40&0.5677&20.71\\
GLM-4.7 &18.12&83.29&0.6498&19.61\\
Qwen1.5 &23.26&66.40&0.3682&3.65\\
OLMoE &17.47&72.43&0.6357&13.51\\
\bottomrule
\end{tabularx}
\vspace{-15pt}
\end{table}

\begin{figure*}[t]
\centering
\includegraphics[width=\textwidth]{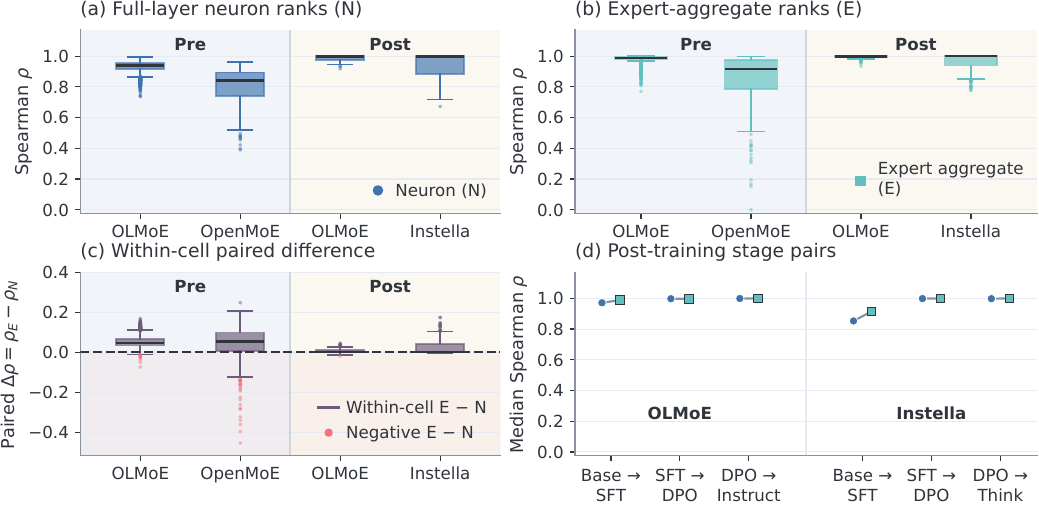}
\caption{Importance-rank correlations between adjacent checkpoints within each domain--layer--pair cell on validation text.
(a) Neuron correlations $\rho_N$ and (b) expert-aggregate correlations $\rho_E$ for pre- and post-training.
(c) Paired differences $\Delta\rho=\rho_E-\rho_N$, with red dots where neurons are more stable.
(d) Neuron and expert medians for each post-training stage pair on the scale of (a) and (b).}
\label{fig:evolution-stage-level}
\end{figure*}

Experts thus serve several domains yet are preferentially routed to the domains whose neurons they disproportionately hold.
This neuron-level account accommodates both sides of the tension that opened this section, since token types shared by all domains drive part of the cross-domain sharing, consistent with token-level accounts of routing \citep{pmlr-v235-xue24c,wang2026mythexpertspecializationmoes}, while the same--cross gap largely persists on these shared tokens and corresponds to domain-associated expert observations \citep{muennighoff2025olmoe,pmlr-v306-herbst26a,pmlr-v306-lasy26a} (Appendix~\ref{app:shared-token}).
Because RAMP includes both dispatch and routing-weight magnitude, we further show that the domain dependence of selected neurons extends beyond what routing alone can produce (for details, see Appendix~\ref{app:routing-contribution}).

\section{Training Dynamics of Neuron Organization}
\label{sec:evolution}

Expert routing has been reported to stabilize early in pre-training \citep{pmlr-v235-xue24c}, and this stability need not extend to the neuron-level.
Thus, we follow neuron importance through training, using expert aggregates as a reference.

\textbf{Models and comparisons.}
\label{sec:evolution-design}
A single training trajectory may reflect the recipe of one model, so we observe each training stage in multiple model families.
One follows OLMoE through pre- and post-training \citep{muennighoff2025olmoe}, and the other uses OpenMoE pre-training \citep{pmlr-v235-xue24c} with Instella post-training \citep{liu2026instellamoe}.
Checkpoints are saved at different intervals across stages and models, so we take an adjacent-checkpoint pair as the common unit of change, from successive saved steps in pre-training and successive released checkpoints in post-training such as Base-to-SFT.
All pairs are evaluated on the same validation text (Appendix~\ref{app:evolution-coverage}).
We measure change with the importance $I_{li}^e(D_d)$ in layer $l$ that defines the domain neuron sets, and sum it within each expert to obtain the expert-level view as a reference.
Within each domain--layer--pair cell, Spearman correlations between the two checkpoints give $\rho_N$ for the full-layer neuron ranking and $\rho_E$ for the expert ranking, and their difference $\Delta\rho=\rho_E-\rho_N$ is positive when experts are more stable.
We report medians across cells for each model and training stage (See Appendix~\ref{app:evolution-stage-level}).

\subsection{Neuron Reorganization During Pre-training}
\label{sec:neuron-pretraining}
\label{sec:expert-dynamics}
Between adjacent pre-training checkpoints, median $\rho_N$ is 0.9393 for OLMoE and 0.8418 for OpenMoE, below the expert medians of 0.9898 and 0.9164 (\cref{fig:evolution-stage-level}a,b).
Within the same cells, the median gaps are 0.0481 and 0.0532, with expert rankings more stable in 5,775/5,808 (99.43\%) and 201/264 (76.14\%) cells, respectively (\cref{fig:evolution-stage-level}c).
Accumulated over training, these steps lower the correlation between the first and last pre-training checkpoints to 0.337 for OLMoE and 0.450 for OpenMoE (Appendix~\ref{app:evolution-intervals}).

Neuron changes could simply follow shifts in routing.
To separate the two, we select adjacent pre-training transitions on discovery text in which at most 5\% of the routing load shifts between experts while neuron-rank correlation stays below 0.90.
On validation text, 209 of 240 OLMoE and 22 of 27 OpenMoE transitions meet both criteria again, so neurons reorganize even where routing barely moves (Appendix~\ref{app:evolution-frozen}).
Decomposing each change into redistribution within experts and shifts of experts' total importance, the within-expert component is larger in every one of these cells, with median norm shares of 0.7830 and 0.6491 (Appendix~\ref{app:evolution-decomposition}).
The largest changes also hold across texts, as the direction of the largest 1\% of OLMoE changes agrees on validation for 89.08\% of neurons, compared with 71.22\% over all changes (Appendix~\ref{app:evolution-replication}).
Stable routing and expert aggregates therefore leave room for pre-training to keep reorganizing which neurons matter inside each expert.
This mirrors \cref{sec:discovery-spatial}, where domain specialization lies in which neurons are selected within experts.
Domain neuron sets already overlap far more within than across domains after 20B tokens and stay so (Appendix~\ref{app:evolution-sets}).

\subsection{Neuron Importance During Post-training}
\label{sec:neuron-posttraining}

Post-training median neuron-importance correlations exceed 0.99 at both levels in OLMoE and Instella, with median gaps of only 0.0007 and 0.0016.
Experts remain more stable in 365/528 (69.13\%) and 737/858 (85.90\%) cells, but by margins small enough that both levels barely change.

The remaining change concentrates at Base-to-SFT (\cref{fig:evolution-stage-level}d), with neuron medians of 0.9715 and 0.8531 and expert medians of 0.9886 and 0.9149 for OLMoE and Instella, respectively.
Instella also releases checkpoints before Base, and this step changes neuron rankings more than the preceding Pretrain-to-Midtrain and Midtrain-to-Base transitions, whose mean correlations are 0.917 and 0.930 against 0.849 at Base-to-SFT (Appendix~\ref{app:evolution-intervals}).
After SFT, both levels' medians exceed 0.9979, and the remaining changes show little consistent direction across texts, with median cross-text correlations of 0.1479 and 0.1750 for Instella's SFT-to-DPO and DPO-to-Think pairs, so later stages add no systematic reorganization (See Appendix~\ref{app:evolution-decomposition}).

The importance underlying neuron-level specialization is thus shaped mainly during pre-training and supervised fine-tuning, while preference optimization largely preserves the structure it inherits.
This timing complements the neuron sets of \cref{sec:neuron-organization}, which are measured on released models whose neuron importance has largely settled.

\section{From Neuron Specialization to Pruning}
\label{sec:pruning}

\Cref{sec:neuron-organization} locates domain specialization in which neurons are selected within and across experts, raising the question of whether this finding can be put to practical use.
Pruning offers a direct test, since if domain capabilities are carried by specific neurons inside experts, removing the remaining neurons should largely preserve them.
We therefore propose RAMP-based Pruning named RAMPr, which aggregates RAMP importance across domains to select one shared neuron set within and across all routed experts.

\subsection{Layer-Wise and Domain-Balanced Selection}
RAMP scores every neuron separately for each domain, and these domain-wise rankings are partly shared and partly domain-dependent (\cref{sec:neuron-organization}), so pruning one model for all domains hinges on how they are combined into a single score.
Summing across domains favors neurons with large overall contributions, but domains with larger total importance can dominate this sum, so we also consider normalizing each domain before selection.
For the importance $I_{li}^e(D_d)$ in layer $l$ (\cref{eq:ramp-weighted}), we define
\begin{equation}
 S^L_{lei}=\sum_d I_{li}^e(D_d),\qquad
 S^D_{lei}=\max_d\frac{I_{li}^e(D_d)}{Z_l(D_d)},
\end{equation}
where $Z_l(D_d)=\sum_{e,i}I_{li}^e(D_d)$ is the total importance of layer $l$ for domain $d$.
The summed rule defines RAMPr-Layer (RAMPr-L), and RAMPr-Domain (RAMPr-D) takes the maximum after normalization so that neurons important to any domain compete on relative contribution.
Both yield one shared neuron set serving all calibrated domains.

Because domain-important neurons concentrate unevenly across experts (\cref{sec:discovery-spatial}), each layer retains its top $k_l=\max(1,\operatorname{round}(rE_lN_l))$ routed neurons across all experts rather than a fixed quota per expert.
The fractions $r\in\{0.5,0.75,0.9\}$ count routed intermediate neuron, not total parameters.
Pruning a neuron removes its rows in the gate and up projections and its down-projection column, keeping shared experts and routers (See Appendix~\ref{app:width-retention}).

\begin{table}[t]
\caption{Mean performance retention (\%) of RAMPr relative to the original model at 50\%, 75\% and 90\% neuron retention.}
\label{tab:pruning-retention}
\centering\small
\setlength{\tabcolsep}{2.5pt}
\rowcolors{3}{}{tabrow}
\begin{tabularx}{\linewidth}{lRRRRRR}
\toprule
\rowcolor{tabhead} & \multicolumn{3}{c}{RAMPr-L} & \multicolumn{3}{c}{RAMPr-D} \\
\cmidrule(lr){2-4}\cmidrule(lr){5-7}
\rowcolor{tabhead}Model & 50 & 75 & 90 & 50 & 75 & 90 \\
\midrule
Qwen3 & 61.04 & 92.64 & 97.44 & 44.47 & 93.99 & 98.74 \\
Qwen3.5 & 83.05 & 96.82 & 98.52 & 77.53 & 98.20 & 99.54 \\
Qwen3.6 & 83.83 & 97.63 & 98.66 & 80.01 & 97.63 & 99.15 \\
GLM-4.7 & 66.16 & 91.12 & 100.77 & 67.33 & 95.61 & 100.67 \\
Qwen1.5 & 88.47 & 93.93 & 95.26 & 85.73 & 92.95 & 103.60 \\
OLMoE & 83.42 & 95.30 & 95.62 & 80.08 & 89.91 & 97.16 \\
\bottomrule
\end{tabularx}
\end{table}

\begin{table}[t]
\caption{Raw-score 11-task macro-average against Fisher and REAP at 50\% neuron retention, with GLM-4.7 also compared at 75\%.
F-G and F-M denote Fisher calibrated on GSM8K and MBPP, and REAP retains the same fraction of experts.}
\label{tab:pruning-main}
\centering\small
\setlength{\tabcolsep}{2.5pt}
\rowcolors{2}{tabrow}{}
\begin{tabularx}{\linewidth}{lRRRRRR}
\toprule
\rowcolor{tabhead}Model & Base & F-G & F-M & REAP & L & D \\
\midrule
Qwen3 &78.45&29.25&29.37&60.77&52.66&40.18\\
Qwen3.5 &85.20&49.75&56.34&--&76.12&74.21\\
Qwen3.6 &84.10&51.65&58.26&--&75.23&75.71\\
GLM-4.7 &74.70&29.87&31.01&--&49.05&52.11\\
Qwen1.5 &45.92&32.59&33.76&--&39.60&39.15\\
OLMoE &46.37&26.89&30.60&--&38.29&38.05\\
\midrule
GLM-4.7 (75\%) &74.70&--&--&68.99&67.86&71.90\\
\bottomrule
\end{tabularx}
\end{table}

\begin{figure*}[t]
\centering
\includegraphics[width=\textwidth]{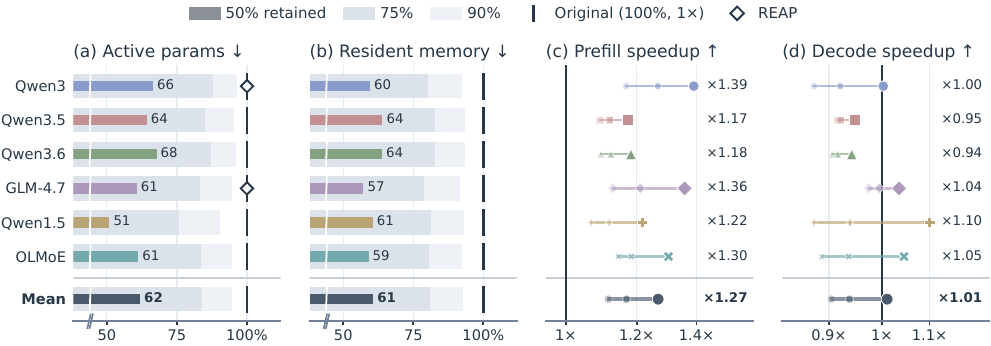}
\caption{RAMPr-L with fixed128 layout at batch 32 and 2048 input tokens, relative to the original model on the same GPU.
The Mean row averages six models arithmetically in (a) and (b) and geometrically in (c) and (d).
(a) Active parameters and (b) resident memory, with colored bars for 50\% retention, gray bands for 75\% and 90\%, ticks at the original and diamonds for REAP.
(c) Prefill and (d) decode speedups on log axes, with points from faint to solid for 90\%, 75\% and 50\% retention and the original model at $1\times$.}
\label{fig:pruning-efficiency}
\vspace{-15pt}
\end{figure*}

\subsection{Experimental Setup}
We evaluate the six models of \cref{sec:discovery-setup} on 13 knowledge and generation tasks, estimating importance from 128 calibration texts in each of eleven domains (Appendix~\ref{app:pruning-protocol}).
We compare with two published pruning methods in their original configurations: Fisher pruning removes neurons with squared-gradient scores and expert-local allocation, calibrated on GSM8K or MBPP \citep{he2026less}, whereas REAP removes whole experts and has checkpoints for two of our models \citep{lasby2026reap}. We report an 11-task raw-score macro-average, excluding the AIME tasks where two models score zero, and mean performance retention, which divides each task score by its Base score to weight tasks of different difficulty comparably (See Appendix~\ref{app:pruning-protocol}).

\subsection{Results and Analysis}
\textbf{Capability retention.}
With a quarter of the routed neurons removed, RAMPr-L keeps 91.12--97.63\% of the original performance, and 95.26--100.77\% at 90\% retention (\cref{tab:pruning-retention}, with per-model curves in Appendix~\ref{app:pruning-metrics}).
At 50\% retention, RAMPr-L still keeps 83.05--88.47\% of performance in four of the six models, with Qwen3 and GLM-4.7 lower at 61.04\% and 66.16\%.
Against other pruning methods, both RAMPr variants exceed both Fisher baselines on the 11-task macro-average in every model at 50\% retention (\cref{tab:pruning-main}).
On GLM-4.7-Flash at 75\% retention, RAMPr-D exceeds REAP by 2.91 points, mainly through MMLU and CMMLU at 69.37 and 68.08 versus 61.62 and 55.44, while HumanEval and MBPP stay within 1.2 points and 13-task retention is nearly equal at 95.61\% versus 95.71\%.
REAP retains more overall on Qwen3 at 50\% (76.74\%), but in both models it preserves generation at the expense of QA tasks, retaining 86.23\% and 67.25\% on them, whereas RAMPr keeps QA retention above 90\% and, on GLM-4.7-Flash, holds both QA and generation tasks at about 95\%.

\textbf{Runtime efficiency.}
We measure pruned RAMPr-L, whose task scores remain unchanged after pruning (Appendix~\ref{app:pruning-pairs}), against the original model, using a fixed block width of 128 rather than the fastest layout per setting.
At 50\% retention, batch 32 and 2048 input tokens, resident memory decreases by 36--43\% and prefill throughput improves by 17--39\%, averaging 60.68\% of the original memory and a 1.27-fold prefill speedup (\cref{fig:pruning-efficiency}).
Decode stays close to the original, with a geometric-mean speedup of 1.01 and per-model values from 0.94 to 1.10.
Savings grow with pruning, as mean memory falls from 92.66\% to 81.07\% and 60.68\% and mean prefill speedup rises from 1.12 to 1.17 and 1.27 at 90\%, 75\% and 50\% retention.
RAMPr also cuts routed-FFN active parameters per token by an estimated 32.29--49.43\% (Appendix~\ref{app:estimated-active}), a saving that REAP cannot offer, since removing whole experts leaves the activated experts of each token unchanged in both number and width.

\textbf{Ablation study.}
\label{sec:pruning-ablation}
Importance is estimated from calibration text, so we vary its size from one to 64 texts per domain in four models, both rules and three retention rates.
Larger calibration sets improve the 50\% RAMPr-L endpoint in each of the four models, with Qwen3 RAMPr-D at 50\% as the only exception to this trend (Appendix~\ref{app:pruning-calibration}).

Calibration composition also offers a simple lever for recovering targeted capabilities without changing the method.
Calibrating Qwen3 at 50\% retention on five generation domains instead of all eleven raises RAMPr-L MBPP from 4.00 to 66.80 and HumanEval from 0.61 to 63.41, lifting generation retention from 35.63\% to 61.64\% and overall retention from 61.04\% to 70.12\%, with multiple-choice retention at 80.00\% (Appendix~\ref{app:pruning-calibration}).
Execution-layout ablations, which fix the neuron masks to isolate implementation effects, are reported separately in Appendix~\ref{app:pruning-efficiency}.

\section{Conclusion}
\label{sec:conclusion}

Looking beyond expert boundaries treats neurons, rather than whole experts, as the unit at which MoE models organize domain specialization.
Domain-specialized neurons overlap across domains, concentrate in a few experts yet recur within them beyond what expert-level counts predict, and their enrichment in particular experts corresponds to routing preferences, reconciling domain-associated routing with the absence of clear expert-level domain boundaries.
Training dynamics show when this organization takes shape, as neuron importance reorganizes inside experts during pre-training while expert-level views stay stable, and it largely settles after supervised fine-tuning.
The same view makes it usable, as RAMP-based Pruning retains one shared multi-domain neuron set that preserves capability better than gradient-based neuron pruning and competes with expert pruning, while reducing memory and computation.
Overall, discovering, tracing and using neurons across experts shows that expert-level analyses capture only part of how MoE language models specialize across domains.


\section*{Impact Statement}
This paper presents work whose goal is to advance the understanding and efficient deployment of Mixture-of-Experts language models.
Insight into how domain-dependent computation is organized at the level of individual neurons can support the inspection of these models, and neuron-level pruning reduces the memory and computation needed to serve them, lowering deployment cost and energy use.
There are many potential societal consequences of our work, none of which we feel must be specifically highlighted here.

\bibliography{custom}

@misc{openai2025gptoss120bgptoss20bmodel,
      title={gpt-oss-120b \& gpt-oss-20b Model Card}, 
      author={OpenAI and : and Sandhini Agarwal and Lama Ahmad and Jason Ai and Sam Altman and Andy Applebaum and Edwin Arbus and Rahul K. Arora and Yu Bai and Bowen Baker and Haiming Bao and Boaz Barak and Ally Bennett and Tyler Bertao and Nivedita Brett and Eugene Brevdo and Greg Brockman and Sebastien Bubeck and Che Chang and Kai Chen and Mark Chen and Enoch Cheung and Aidan Clark and Dan Cook and Marat Dukhan and Casey Dvorak and Kevin Fives and Vlad Fomenko and Timur Garipov and Kristian Georgiev and Mia Glaese and Tarun Gogineni and Adam Goucher and Lukas Gross and Katia Gil Guzman and John Hallman and Jackie Hehir and Johannes Heidecke and Alec Helyar and Haitang Hu and Romain Huet and Jacob Huh and Saachi Jain and Zach Johnson and Chris Koch and Irina Kofman and Dominik Kundel and Jason Kwon and Volodymyr Kyrylov and Elaine Ya Le and Guillaume Leclerc and James Park Lennon and Scott Lessans and Mario Lezcano-Casado and Yuanzhi Li and Zhuohan Li and Ji Lin and Jordan Liss and Lily and Liu and Jiancheng Liu and Kevin Lu and Chris Lu and Zoran Martinovic and Lindsay McCallum and Josh McGrath and Scott McKinney and Aidan McLaughlin and Song Mei and Steve Mostovoy and Tong Mu and Gideon Myles and Alexander Neitz and Alex Nichol and Jakub Pachocki and Alex Paino and Dana Palmie and Ashley Pantuliano and Giambattista Parascandolo and Jongsoo Park and Leher Pathak and Carolina Paz and Ludovic Peran and Dmitry Pimenov and Michelle Pokrass and Elizabeth Proehl and Huida Qiu and Gaby Raila and Filippo Raso and Hongyu Ren and Kimmy Richardson and David Robinson and Bob Rotsted and Hadi Salman and Suvansh Sanjeev and Max Schwarzer and D. Sculley and Harshit Sikchi and Kendal Simon and Karan Singhal and Yang Song and Dane Stuckey and Zhiqing Sun and Philippe Tillet and Sam Toizer and Foivos Tsimpourlas and Nikhil Vyas and Eric Wallace and Xin Wang and Miles Wang and Olivia Watkins and Kevin Weil and Amy Wendling and Kevin Whinnery and Cedric Whitney and Hannah Wong and Lin Yang and Yu Yang and Michihiro Yasunaga and Kristen Ying and Wojciech Zaremba and Wenting Zhan and Cyril Zhang and Brian Zhang and Eddie Zhang and Shengjia Zhao},
      year={2025},
      eprint={2508.10925},
      archivePrefix={arXiv},
      primaryClass={cs.CL},
      url={https://arxiv.org/abs/2508.10925}, 
}

@inproceedings{dai-etal-2024-deepseekmoe,
    title = "{D}eep{S}eek{M}o{E}: Towards Ultimate Expert Specialization in Mixture-of-Experts Language Models",
    author = "Dai, Damai  and
      Deng, Chengqi  and
      Zhao, Chenggang  and
      Xu, R.X.  and
      Gao, Huazuo  and
      Chen, Deli  and
      Li, Jiashi  and
      Zeng, Wangding  and
      Yu, Xingkai  and
      Wu, Y.  and
      Xie, Zhenda  and
      Li, Y.K.  and
      Huang, Panpan  and
      Luo, Fuli  and
      Ruan, Chong  and
      Sui, Zhifang  and
      Liang, Wenfeng",
    editor = "Ku, Lun-Wei  and
      Martins, Andre  and
      Srikumar, Vivek",
    booktitle = "Proceedings of the 62nd Annual Meeting of the {A}ssociation for {C}omputational {L}inguistics (Volume 1: Long Papers)",
    month = aug,
    year = "2024",
    address = "Bangkok, Thailand",
    publisher = "Association for Computational Linguistics",
    url = "https://aclanthology.org/2024.acl-long.70/",
    doi = "10.18653/v1/2024.acl-long.70",
    pages = "1280--1297"
}

@inproceedings{
muennighoff2025olmoe,
title={{OLM}oE: Open Mixture-of-Experts Language Models},
author={Niklas Muennighoff and Luca Soldaini and Dirk Groeneveld and Kyle Lo and Jacob Morrison and Sewon Min and Weijia Shi and Evan Pete Walsh and Oyvind Tafjord and Nathan Lambert and Yuling Gu and Shane Arora and Akshita Bhagia and Dustin Schwenk and David Wadden and Alexander Wettig and Binyuan Hui and Tim Dettmers and Douwe Kiela and Ali Farhadi and Noah A. Smith and Pang Wei Koh and Amanpreet Singh and Hannaneh Hajishirzi},
booktitle={The Thirteenth International Conference on Learning Representations},
year={2025},
url={https://openreview.net/forum?id=xXTkbTBmqq}
}

@InProceedings{pmlr-v306-herbst26a,
  title = 	 {The Expert Strikes Back: Interpreting Mixture-of-Experts Language Models at Expert Level},
  author =       {Herbst, Jeremy and Wermter, Stefan and Lee, Jae Hee},
  booktitle = 	 {Proceedings of the 43rd International Conference on Machine Learning},
  pages = 	 {42781--42820},
  year = 	 {2026},
  editor = 	 {Zhang, Tong and Dudik, Miroslav and Jaggi, Martin and Agarwal, Alekh and Li, Sharon and Schuurmans, Dale and Zhu, Jerry and Berkenkamp, Felix and Dong, Hanze and Bietti, Alberto},
  volume = 	 {306},
  series = 	 {Proceedings of Machine Learning Research},
  month = 	 {06--11 Jul},
  publisher =    {PMLR},
  url = 	 {https://proceedings.mlr.press/v306/herbst26a.html}
}

@InProceedings{pmlr-v235-xue24c,
  title = 	 {{O}pen{M}o{E}: An Early Effort on Open Mixture-of-Experts Language Models},
  author =       {Xue, Fuzhao and Zheng, Zian and Fu, Yao and Ni, Jinjie and Zheng, Zangwei and Zhou, Wangchunshu and You, Yang},
  booktitle = 	 {Proceedings of the 41st International Conference on Machine Learning},
  pages = 	 {55625--55655},
  year = 	 {2024},
  editor = 	 {Salakhutdinov, Ruslan and Kolter, Zico and Heller, Katherine and Weller, Adrian and Oliver, Nuria and Scarlett, Jonathan and Berkenkamp, Felix},
  volume = 	 {235},
  series = 	 {Proceedings of Machine Learning Research},
  month = 	 {21--27 Jul},
  publisher =    {PMLR},
  url = 	 {https://proceedings.mlr.press/v235/xue24c.html}
}

@misc{wang2026mythexpertspecializationmoes,
      title={The Myth of Expert Specialization in MoEs: Why Routing Reflects Geometry, Not Necessarily Domain Expertise}, 
      author={Xi Wang and Soufiane Hayou and Eric Nalisnick},
      year={2026},
      eprint={2604.09780},
      archivePrefix={arXiv},
      primaryClass={cs.AI},
      url={https://arxiv.org/abs/2604.09780}, 
}

@InProceedings{pmlr-v306-lasy26a,
  title = 	 {{R}outer{I}nterp: Understanding Superposed Specialisation in Mixture of Experts Routing},
  author =       {Lasy, Ilya and Cai, Nora Yinuo and Ayonrinde, Kola},
  booktitle = 	 {Proceedings of the 43rd International Conference on Machine Learning},
  pages = 	 {62858--62883},
  year = 	 {2026},
  editor = 	 {Zhang, Tong and Dudik, Miroslav and Jaggi, Martin and Agarwal, Alekh and Li, Sharon and Schuurmans, Dale and Zhu, Jerry and Berkenkamp, Felix and Dong, Hanze and Bietti, Alberto},
  volume = 	 {306},
  series = 	 {Proceedings of Machine Learning Research},
  month = 	 {06--11 Jul},
  publisher =    {PMLR},
  url = 	 {https://proceedings.mlr.press/v306/lasy26a.html}
}

@misc{deepseekai2026deepseekv4highlyefficientmilliontoken,
      title={DeepSeek-V4: Towards Highly Efficient Million-Token Context Intelligence}, 
      author={DeepSeek-AI and Anyi Xu and Bangcai Lin and Bing Xue and Bingxuan Wang and Bingzheng Xu and Bochao Wu and Bowei Zhang and Chaofan Lin and Chen Dong and Chenchen Ling and Chengda Lu and Chenggang Zhao and Chengqi Deng and Chengyu Hou and Chenhao Xu and Chenze Shao and Chong Ruan and Conner Sun and Damai Dai and Daya Guo and Dejian Yang and Deli Chen and Donghao Li and Dongjie Ji and Erhang Li and Fang Wei and Fangyun Lin and Fangzhou Yuan and Feiyu Xia and Fucong Dai and Guangbo Hao and Guanting Chen and Guoai Cao and Guolai Meng and Guowei Li and Han Yu and Han Zhang and Hanwei Xu and Hao Li and Haofen Liang and Haoling Zhang and Haoming Luo and Haoran Wei and Haotian Yuan and Haowei Zhang and Haowen Luo and Haoyu Chen and Haozhe Ji and Hengqing Zhang and Honghui Ding and Hongxuan Tang and Huanqi Cao and Huazuo Gao and Hui Qu and Hui Zeng and J Yang and JQ Zhu and Jia Luo and Jia Song and Jia Yu and Jialiang Huang and Jialu Cai and Jian Liang and Jiangting Zhou and Jiasheng Ye and Jiashi Li and Jiaxin Xu and Jiewen Hu and Jieyu Yang and Jin Chen and Jin Yan and Jingchang Chen and Jingli Zhou and Jingting Xiang and Jingyang Yuan and Jingyuan Cheng and Jingzi Zhou and Jinhua Zhu and Jiping Yu and Joseph Sun and Jun Ran and Junguang Jiang and Junjie Qiu and Junlong Li and Junmin Zheng and Junxiao Song and Kai Dong and Kaige Gao and Kang Guan and Kexing Zhou and Kezhao Huang and Kuai Yu and Lean Wang and Lecong Zhang and Lei Wang and Leyi Xia and Li Zhang and Liang Zhao and Lihua Guo and Lingxiao Luo and Linwang Ma and Linyan Zhu and Litong Wang and Liyu Cai and Liyue Zhang and Longhao Chen and MS Di and MY Xu and Max Mei and Miaojun Wang and Mingchuan Zhang and Minghua Zhang and Minghui Tang and Mingming Li and Mingxu Zhou and Minmin Han and Ning Wang and Panpan Huang and Panpan Wang and Peixin Cong and Peiyi Wang and Peng Zhang and Qiancheng Wang and Qihao Zhu and Qingyang Li and Qinyu Chen and Qiushi Du and Qiwei Jiang and Rui Tian and Ruifan Xu and Ruijie Lu and Ruiling Xu and Ruiqi Ge and Ruisong Zhang and Ruizhe Pan and Runji Wang and Runqian Chen and Runqiu Yin and Runxin Xu and Ruomeng Shen and Ruoyu Zhang and Ruyi Chen and SH Liu and Shanghao Lu and Shangmian Sun and Shangyan Zhou and Shanhuang Chen and Shaofei Cai and Shaoheng Nie and Shaoqing Wu and Shaoyuan Chen and Shengding Hu and Shengyu Liu and Shiqiang Hu and Shirong Ma and Shiyu Wang and Shuiping Yu and Shunfeng Zhou and Shuting Pan and Shuying Yu and Songyang Zhou and Tao Ni and Tao Yun and Tian Jin and Tian Pei and Tian Ye and Tianle Lin and Tianran Ji and Tianyi Cui and Tianyuan Yue and Tingting Yu and Tun Wang and W Zhang and WL Xiao and Wangding Zeng and Wei An and Weilin Zhao and Wen Liu and Wenfeng Liang and Wenjie Pang and Wenjing Luo and Wenjing Yao and Wenjun Gao and Wenkai Yang and Wenlve Huang and Wenqing Hou and Wentao Zhang and Wenting Ma and Xi Gao and Xiang He and Xiangwen Wang and Xianzu Wang and Xiao Bi and Xiaodong Liu and Xiaohan Wang and Xiaokang Chen and Xiaokang Zhang and Xiaotao Nie and Xiaowen Sun and Xiaoxiang Wang and Xin Cheng and Xin Liu and Xin Xie and Xingchao Liu and Xingchen Liu and Xingkai Yu and Xingyou Li and Xinyu Yang and Xinyu Zhang and Xu Chen and Xuanyu Wang and Xuecheng Su and Xueyin Chen and Xuheng Lin and Xuwei Fu and YC Yan and YQ Wang and YW Ma and Yanfeng Luo and Yang Zhang and Yanhong Xu and Yanru Ma and Yanwen Huang and Yao Li and Yao Li and Yao Xu and Yao Zhao and Yaofeng Sun and Yaohui Wang and Yi Qian and Yi Shao and Yi Yu and Yichao Zhang and Yifan Ding and Yifan Shi and Yijia Wu and Yiliang Xiong and Yiling Ma and Ying He and Ying Tang and Ying Zhou and Yingjia Luo and Yinmin Zhong and Yishi Piao and Yisong Wang and Yixiang Zhang and Yixiao Chen and Yixuan Tan and Yixuan Wei and Yiyang Ma and Yiyuan Liu and Yonglun Yang and Yongqiang Guo and Yongtong Wu and Yu Wu and YuKun Li and Yuan Cheng and Yuan Ou and Yuanfan Xu and Yuanhao Li and Yuduan Wang and Yuehan Yang and Yuer Xu and Yuhan Wu and Yuhao Meng and Yuheng Zou and Yukun Zha and Yunfan Xiong and Yupeng Chen and Yuping Lin and Yuqian Cao and Yuqian Wang and Yushun Zhang and Yuting Yan and Yutong Lin and Yuxian Gu and Yuxiang Luo and Yuxiang You and Yuxuan Liu and Yuxuan Zhou and Yuyang Zhou and Yuzhen Huang and ZF Wu and Zehao Wang and Zehua Zhao and Zehui Ren and Zekai Zhang and Zhangli Sha and Zhe Fu and Zhe Ju and Zhean Xu and Zhenda Xie and Zhengyan Zhang and Zheren Gao and Zhewen Hao and Zhibin Gou and Zhicheng Ma and Zhigang Yan and Zhihong Shao and Zhixian Huang and Zhixuan Chen and Zhiyu Wu and Zhizhou Ren and Zhongyu Wu and Zhuoshu Li and Zhuping Zhang and Zian Xu and Zihao Wang and Zihua Qu and Zihui Gu and Zijia Zhu and Zilin Li and Zipeng Zhang and Ziwei Xie and Ziyi Gao and Ziyi Wan and Zizheng Pan and Zongqing Yao},
      year={2026},
      eprint={2606.19348},
      archivePrefix={arXiv},
      primaryClass={cs.CL},
      url={https://arxiv.org/abs/2606.19348}, 
}

@misc{qiu2026designqwen38nextarchitectureevaluation,
      title={On the Design of Qwen3.8-Next Architecture: Evaluation, Efficiency, and Training Stability}, 
      author={Zihan Qiu and Zekun Wang and Xiao Li and Yanpeng Li and Yang Xu and Yixuan Wang and Huaqing Zhang and Rui Men and Bochao Mao and Chengruidong Zhang and Fan Zhou and Hao Luo and Haofeng Huang and Haoran Lian and Haoyan Huang and Hongqing Chen and Jianwei Zhang and Jing Xu and Junjie Wang and Langshi Chen and Liangyu Wang and Linlang Jiang and Man Yuan and Minmin Sun and Peng Jin and Siqi Zhang and Siyu Wang and Xingzhang Ren and Yakai Wang and Yi Zhang and Yiming Dong and Yizhong Cao and Yubo Ma and Yunfei Mao and Bo Zheng and Dayiheng Liu},
      year={2026},
      eprint={2608.30320},
      archivePrefix={arXiv},
      primaryClass={cs.CL},
      url={https://arxiv.org/abs/2608.30320}, 
}

@article{jiang2024mixtral,
      title={Mixtral of Experts}, 
      author={Albert Q. Jiang and Alexandre Sablayrolles and Antoine Roux and Arthur Mensch and Blanche Savary and Chris Bamford and Devendra Singh Chaplot and Diego de las Casas and Emma Bou Hanna and Florian Bressand and Gianna Lengyel and Guillaume Bour and Guillaume Lample and Lélio Renard Lavaud and Lucile Saulnier and Marie-Anne Lachaux and Pierre Stock and Sandeep Subramanian and Sophia Yang and Szymon Antoniak and Teven Le Scao and Théophile Gervet and Thibaut Lavril and Thomas Wang and Timothée Lacroix and William El Sayed},
      year={2024},
      eprint={2401.04088},
      archivePrefix={arXiv},
      primaryClass={cs.LG},
      url={https://arxiv.org/abs/2401.04088}, 
}

@inproceedings{wang2026deconstructing,
  title={Deconstructing pre-training: Knowledge attribution analysis in moe and dense models},
  author={Wang, Bo and Li, Junzhuo and Chen, Hong and Chu, Yuanlin and Fan, Yuxuan and Hu, Xuming},
  booktitle={Proceedings of the AAAI Conference on Artificial Intelligence},
  volume={40},
  number={39},
  pages={33359--33367},
  year={2026}
}

@inproceedings{qiu2025demons,
  title={Demons in the detail: On implementing load balancing loss for training specialized mixture-of-expert models},
  author={Qiu, Zihan and Huang, Zeyu and Zheng, Bo and Wen, Kaiyue and Wang, Zekun and Men, Rui and Titov, Ivan and Liu, Dayiheng and Zhou, Jingren and Lin, Junyang},
  booktitle={Proceedings of the 63rd Annual Meeting of the Association for Computational Linguistics (Volume 1: Long Papers)},
  pages={5005--5018},
  year={2025}
}

@inproceedings{zhang2023emergent,
  title = {Emergent Modularity in Pre-trained Transformers},
  author = {Zhang, Zhengyan and Zeng, Zhiyuan and Lin, Yankai and Xiao, Chaojun and Wang, Xiaozhi and Han, Xu and Liu, Zhiyuan and Xie, Ruobing and Sun, Maosong and Zhou, Jie},
  booktitle = {Findings of the Association for Computational Linguistics: ACL 2023},
  year = {2023},
  pages = {4066--4083},
  publisher = {Association for Computational Linguistics},
  doi = {10.18653/v1/2023.findings-acl.250},
  url = {https://aclanthology.org/2023.findings-acl.250/}
}

@article{cheng2025mone,
  title = {Mixture of Neuron Experts},
  author = {Cheng, Runxi and Guan, Yuchen and Ding, Yucheng and Hu, Qingguo and Wei, Yongxian and Yuan, Chun and Shen, Yelong and Chen, Weizhu and Gong, Yeyun},
  journal = {arXiv preprint arXiv:2510.05781},
  year = {2025},
  url = {https://arxiv.org/abs/2510.05781}
}

@article{he2026less,
  title = {Less is {MoE}: Trimming Experts in Domain-Specialist Language Models},
  author = {He, Haoze and Zou, Xinkai and Jiang, Xuan and Ding, Xingyuan and Qu, Ao and Li, Juncheng Billy and Miller, Heather},
  journal = {arXiv preprint arXiv:2606.05538},
  year = {2026},
  url = {https://arxiv.org/abs/2606.05538}
}

@inproceedings{zhou2025dern,
  title = {Dropping Experts, Recombining Neurons: Retraining-Free Pruning for Sparse Mixture-of-Experts {LLMs}},
  author = {Zhou, Yixiao and Zhao, Ziyu and Cheng, Dongzhou and Wu, Zhiliang and Gui, Jie and Yang, Yi and Wu, Fei and Cheng, Yu and Fan, Hehe},
  booktitle = {Findings of the Association for Computational Linguistics: EMNLP 2025},
  year = {2025},
  pages = {15169--15186},
  publisher = {Association for Computational Linguistics},
  doi = {10.18653/v1/2025.findings-emnlp.820},
  url = {https://aclanthology.org/2025.findings-emnlp.820/}
}

@inproceedings{lasby2026reap,
  title = {{REAP} the Experts: Why Pruning Prevails for One-Shot {MoE} compression},
  author = {Lasby, Mike and Lazarevich, Ivan and Sinnadurai, Nish and Lie, Sean and Ioannou, Yani and Thangarasa, Vithursan},
  booktitle = {International Conference on Learning Representations},
  year = {2026},
  volume = {2026},
  pages = {146883--146912},
  url = {https://proceedings.iclr.cc/paper_files/paper/2026/file/ed93b2b5722acc2341d421b8916404a1-Paper-Conference.pdf}
}

@article{liu2026instellamoe,
  title={Instella-MoE Technical Report},
  author={Liu, Jiang and Ranjan, Sudhanshu and Mishra, Prakamya and Dukler, Yonatan and Ramesh, Gowtham and Wu, Jialian and Sun, Ximeng and Xie, Wen and Hou, Chaojun and Appia, Vikram and others},
  journal={arXiv preprint arXiv:2609.00791},
  year={2026}
}

@misc{qwen2026qwen35,
  title = {{Qwen3.5}: Towards Native Multimodal Agents},
  author = {{Qwen Team}},
  year = {2026},
  month = feb,
  url = {https://qwen.ai/blog?id=qwen3.5}
}

@misc{qwen2026qwen36,
  title = {{Qwen3.6-35B-A3B}: Agentic Coding Power, Now Open to All},
  author = {{Qwen Team}},
  year = {2026},
  month = apr,
  url = {https://qwen.ai/blog?id=qwen3.6-35b-a3b}
}

@misc{glm2025glm45,
  title = {{GLM-4.5}: Agentic, Reasoning, and Coding ({ARC}) Foundation Models},
  author = {{GLM Team} and Zeng, Aohan and Lv, Xin and Zheng, Qinkai and others},
  year = {2025},
  eprint = {2508.06471},
  archivePrefix = {arXiv},
  url = {https://arxiv.org/abs/2508.06471}
}

@misc{qwen2024qwen15moe,
  title = {{Qwen1.5-MoE-A2.7B-Chat}},
  author = {{Qwen Team}},
  year = {2024},
  howpublished = {Model card},
  url = {https://huggingface.co/Qwen/Qwen1.5-MoE-A2.7B-Chat}
}

@misc{qwen2025qwen3,
  title = {{Qwen3} Technical Report},
  author = {{Qwen Team}},
  year = {2025},
  eprint = {2505.09388},
  archivePrefix = {arXiv},
  primaryClass = {cs.CL},
  url = {https://arxiv.org/abs/2505.09388}
}

@inproceedings{sakaguchi2020winogrande,
  title = {{WinoGrande}: An Adversarial {Winograd} Schema Challenge at Scale},
  author = {Sakaguchi, Keisuke and Le Bras, Ronan and Bhagavatula, Chandra and Choi, Yejin},
  booktitle = {Proceedings of the AAAI Conference on Artificial Intelligence},
  year = {2020},
  volume = {34},
  pages = {8732--8740},
  doi = {10.1609/aaai.v34i05.6399},
  url = {https://ojs.aaai.org/index.php/AAAI/article/view/6399}
}

@inproceedings{zellers2019hellaswag,
  title = {{HellaSwag}: Can a Machine Really Finish Your Sentence?},
  author = {Zellers, Rowan and Holtzman, Ari and Bisk, Yonatan and Farhadi, Ali and Choi, Yejin},
  booktitle = {Proceedings of the 57th Annual Meeting of the Association for Computational Linguistics},
  year = {2019},
  pages = {4791--4800},
  publisher = {Association for Computational Linguistics},
  doi = {10.18653/v1/P19-1472},
  url = {https://aclanthology.org/P19-1472/}
}

@inproceedings{huang2023ceval,
  title = {{C-Eval}: A Multi-Level Multi-Discipline {Chinese} Evaluation Suite for Foundation Models},
  author = {Huang, Yuzhen and Bai, Yuzhuo and Zhu, Zhihao and Zhang, Junlei and Zhang, Jinghan and Su, Tangjun and Liu, Junteng and Lv, Chuancheng and Zhang, Yikai and Lei, Jiayi and Fu, Yao and Sun, Maosong and He, Junxian},
  booktitle = {Advances in Neural Information Processing Systems},
  year = {2023},
  url = {https://arxiv.org/abs/2305.08322}
}

@inproceedings{li2024cmmlu,
  title = {{CMMLU}: Measuring massive multitask language understanding in {Chinese}},
  author = {Li, Haonan and Zhang, Yixuan and Koto, Fajri and Yang, Yifei and Zhao, Hai and Gong, Yeyun and Duan, Nan and Baldwin, Timothy},
  booktitle = {Findings of the Association for Computational Linguistics: ACL 2024},
  year = {2024},
  pages = {11260--11285},
  publisher = {Association for Computational Linguistics},
  doi = {10.18653/v1/2024.findings-acl.671},
  url = {https://aclanthology.org/2024.findings-acl.671/}
}

@inproceedings{hendrycks2021mmlu,
  title = {Measuring Massive Multitask Language Understanding},
  author = {Hendrycks, Dan and Burns, Collin and Basart, Steven and Zou, Andy and Mazeika, Mantas and Song, Dawn and Steinhardt, Jacob},
  booktitle = {International Conference on Learning Representations},
  year = {2021},
  url = {https://openreview.net/forum?id=d7KBjmI3GmQ}
}

@inproceedings{jin2019pubmedqa,
  title = {{PubMedQA}: A Dataset for Biomedical Research Question Answering},
  author = {Jin, Qiao and Dhingra, Bhuwan and Liu, Zhengping and Cohen, William and Lu, Xinghua},
  booktitle = {Proceedings of the 2019 Conference on Empirical Methods in Natural Language Processing and the 9th International Joint Conference on Natural Language Processing (EMNLP-IJCNLP)},
  year = {2019},
  pages = {2567--2577},
  publisher = {Association for Computational Linguistics},
  doi = {10.18653/v1/D19-1259},
  url = {https://aclanthology.org/D19-1259/}
}

@article{austin2021mbpp,
  title = {Program Synthesis with Large Language Models},
  author = {Austin, Jacob and Odena, Augustus and Nye, Maxwell and Bosma, Maarten and Michalewski, Henryk and Dohan, David and Jiang, Ellen and Cai, Carrie and Terry, Michael and Le, Quoc and Sutton, Charles},
  journal = {arXiv preprint arXiv:2108.07732},
  year = {2021},
  url = {https://arxiv.org/abs/2108.07732}
}

@article{chen2021humaneval,
  title = {Evaluating Large Language Models Trained on Code},
  author = {Chen, Mark and Tworek, Jerry and Jun, Heewoo and Yuan, Qiming and de Oliveira Pinto, Henrique Ponde and Kaplan, Jared and Edwards, Harri and Burda, Yuri and Joseph, Nicholas and Brockman, Greg and Ray, Alex and Puri, Raul and Krueger, Gretchen and Petrov, Michael and Khlaaf, Heidy and Sastry, Girish and Mishkin, Pamela and Chan, Brooke and Gray, Scott and Ryder, Nick and Pavlov, Mikhail and Power, Alethea and Kaiser, Lukasz and Bavarian, Mohammad and Winter, Clemens and Tillet, Philippe and Such, Felipe Petroski and Cummings, Dave and Plappert, Matthias and Chantzis, Fotios and Barnes, Elizabeth and Herbert-Voss, Ariel and Guss, William Hebgen and Nichol, Alex and Paino, Alex and Tezak, Nikolas and Tang, Jie and Babuschkin, Igor and Balaji, Suchir and Jain, Shantanu and Saunders, William and Hesse, Christopher and Carr, Andrew N. and Leike, Jan and Achiam, Josh and Misra, Vedant and Morikawa, Evan and Radford, Alec and Knight, Matthew and Brundage, Miles and Murati, Mira and Mayer, Katie and Welinder, Peter and McGrew, Bob and Amodei, Dario and McCandlish, Sam and Sutskever, Ilya and Zaremba, Wojciech},
  journal = {arXiv preprint arXiv:2107.03374},
  year = {2021},
  url = {https://arxiv.org/abs/2107.03374}
}

@article{cobbe2021gsm8k,
  title = {Training Verifiers to Solve Math Word Problems},
  author = {Cobbe, Karl and Kosaraju, Vineet and Bavarian, Mohammad and Chen, Mark and Jun, Heewoo and Kaiser, Lukasz and Plappert, Matthias and Tworek, Jerry and Hilton, Jacob and Nakano, Reiichiro and Hesse, Christopher and Schulman, John},
  journal = {arXiv preprint arXiv:2110.14168},
  year = {2021},
  url = {https://arxiv.org/abs/2110.14168}
}

@inproceedings{hendrycks2021math,
  title = {Measuring Mathematical Problem Solving With the {MATH} Dataset},
  author = {Hendrycks, Dan and Burns, Collin and Kadavath, Saurav and Arora, Akul and Basart, Steven and Tang, Eric and Song, Dawn and Steinhardt, Jacob},
  booktitle = {Proceedings of the Neural Information Processing Systems Track on Datasets and Benchmarks},
  year = {2021},
  volume = {1},
  url = {https://datasets-benchmarks-proceedings.neurips.cc/paper/2021/hash/be83ab3ecd0db773eb2dc1b0a17836a1-Abstract-round2.html}
}

@inproceedings{rein2024gpqa,
  title = {{GPQA}: A Graduate-Level Google-Proof {Q\&A} Benchmark},
  author = {Rein, David and Hou, Betty Li and Stickland, Asa Cooper and Petty, Jackson and Pang, Richard Yuanzhe and Dirani, Julien and Michael, Julian and Bowman, Samuel R.},
  booktitle = {First Conference on Language Modeling},
  year = {2024},
  url = {https://openreview.net/forum?id=Ti67584b98}
}

@inproceedings{ding2023ultrachat,
  title = {Enhancing Chat Language Models by Scaling High-quality Instructional Conversations},
  author = {Ding, Ning and Chen, Yulin and Xu, Bokai and Qin, Yujia and Hu, Shengding and Liu, Zhiyuan and Sun, Maosong and Zhou, Bowen},
  booktitle = {Proceedings of the 2023 Conference on Empirical Methods in Natural Language Processing},
  year = {2023},
  pages = {3029--3051},
  publisher = {Association for Computational Linguistics},
  doi = {10.18653/v1/2023.emnlp-main.183},
  url = {https://aclanthology.org/2023.emnlp-main.183/}
}

@inproceedings{kopf2023openassistant,
  title = {{OpenAssistant} Conversations - Democratizing Large Language Model Alignment},
  author = {K\"{o}pf, Andreas and Kilcher, Yannic and von R\"{u}tte, Dimitri and Anagnostidis, Sotiris and Tam, Zhi Rui and Stevens, Keith and Barhoum, Abdullah and Nguyen, Duc and Stanley, Oliver and Nagyfi, Rich\'{a}rd and ES, Shahul and Suri, Sameer and Glushkov, David and Dantuluri, Arnav and Maguire, Andrew and Schuhmann, Christoph and Nguyen, Huu and Mattick, Alexander},
  booktitle = {Advances in Neural Information Processing Systems},
  year = {2023},
  volume = {36},
  pages = {47669--47681},
  publisher = {Curran Associates, Inc.},
  doi = {10.52202/075280-2064},
  url = {https://papers.nips.cc/paper_files/paper/2023/hash/949f0f8f32267d297c2d4e3ee10a2e7e-Abstract-Datasets_and_Benchmarks.html}
}

@inproceedings{wei2024magicoder,
  title = {{Magicoder}: Empowering Code Generation with {OSS}-Instruct},
  author = {Wei, Yuxiang and Wang, Zhe and Liu, Jiawei and Ding, Yifeng and Zhang, Lingming},
  booktitle = {Proceedings of the 41st International Conference on Machine Learning},
  year = {2024},
  volume = {235},
  series = {Proceedings of Machine Learning Research},
  pages = {52632--52657},
  publisher = {PMLR},
  url = {https://proceedings.mlr.press/v235/wei24h.html}
}

@inproceedings{yu2024metamath,
  title = {{MetaMath}: Bootstrap Your Own Mathematical Questions for Large Language Models},
  author = {Yu, Longhui and Jiang, Weisen and Shi, Han and Yu, Jincheng and Liu, Zhengying and Zhang, Yu and Kwok, James T. and Li, Zhenguo and Weller, Adrian and Liu, Weiyang},
  booktitle = {The Twelfth International Conference on Learning Representations},
  year = {2024},
  url = {https://openreview.net/forum?id=N8N0hgNDRt}
}

@inproceedings{ling2017aqua,
  title = {Program Induction by Rationale Generation: Learning to Solve and Explain Algebraic Word Problems},
  author = {Ling, Wang and Yogatama, Dani and Dyer, Chris and Blunsom, Phil},
  booktitle = {Proceedings of the 55th Annual Meeting of the Association for Computational Linguistics (Volume 1: Long Papers)},
  year = {2017},
  pages = {158--167},
  publisher = {Association for Computational Linguistics},
  doi = {10.18653/v1/P17-1015},
  url = {https://aclanthology.org/P17-1015/}
}

@inproceedings{welbl2017sciq,
  title = {Crowdsourcing Multiple Choice Science Questions},
  author = {Welbl, Johannes and Liu, Nelson F. and Gardner, Matt},
  booktitle = {Proceedings of the 3rd Workshop on Noisy User-generated Text},
  year = {2017},
  pages = {94--106},
  publisher = {Association for Computational Linguistics},
  doi = {10.18653/v1/W17-4413},
  url = {https://aclanthology.org/W17-4413/}
}

@inproceedings{pal2022medmcqa,
  title = {{MedMCQA}: A Large-scale Multi-Subject Multi-Choice Dataset for Medical domain Question Answering},
  author = {Pal, Ankit and Umapathi, Logesh Kumar and Sankarasubbu, Malaikannan},
  booktitle = {Proceedings of the Conference on Health, Inference, and Learning},
  year = {2022},
  volume = {174},
  series = {Proceedings of Machine Learning Research},
  pages = {248--260},
  publisher = {PMLR},
  url = {https://proceedings.mlr.press/v174/pal22a.html}
}

@article{jin2021medqa,
  title = {What Disease Does This Patient Have? A Large-Scale Open Domain Question Answering Dataset from Medical Exams},
  author = {Jin, Di and Pan, Eileen and Oufattole, Nassim and Weng, Wei-Hung and Fang, Hanyi and Szolovits, Peter},
  journal = {Applied Sciences},
  year = {2021},
  volume = {11},
  number = {14},
  pages = {6421},
  doi = {10.3390/app11146421},
  url = {https://www.mdpi.com/2076-3417/11/14/6421}
}

@inproceedings{talmor2019commonsenseqa,
  title = {{CommonsenseQA}: A Question Answering Challenge Targeting Commonsense Knowledge},
  author = {Talmor, Alon and Herzig, Jonathan and Lourie, Nicholas and Berant, Jonathan},
  booktitle = {Proceedings of the 2019 Conference of the North American Chapter of the Association for Computational Linguistics: Human Language Technologies, Volume 1 (Long and Short Papers)},
  year = {2019},
  pages = {4149--4158},
  publisher = {Association for Computational Linguistics},
  doi = {10.18653/v1/N19-1421},
  url = {https://aclanthology.org/N19-1421/}
}

@inproceedings{sap2019socialiqa,
  title = {Social {IQ}a: Commonsense Reasoning about Social Interactions},
  author = {Sap, Maarten and Rashkin, Hannah and Chen, Derek and Le Bras, Ronan and Choi, Yejin},
  booktitle = {Proceedings of the 2019 Conference on Empirical Methods in Natural Language Processing and the 9th International Joint Conference on Natural Language Processing (EMNLP-IJCNLP)},
  year = {2019},
  pages = {4463--4473},
  publisher = {Association for Computational Linguistics},
  doi = {10.18653/v1/D19-1454},
  url = {https://aclanthology.org/D19-1454/}
}

@inproceedings{huang2019cosmosqa,
  title = {Cosmos {QA}: Machine Reading Comprehension with Contextual Commonsense Reasoning},
  author = {Huang, Lifu and Le Bras, Ronan and Bhagavatula, Chandra and Choi, Yejin},
  booktitle = {Proceedings of the 2019 Conference on Empirical Methods in Natural Language Processing and the 9th International Joint Conference on Natural Language Processing (EMNLP-IJCNLP)},
  year = {2019},
  pages = {2391--2401},
  publisher = {Association for Computational Linguistics},
  doi = {10.18653/v1/D19-1243},
  url = {https://aclanthology.org/D19-1243/}
}

@article{li2023chatdoctor,
  title = {{ChatDoctor}: A Medical Chat Model Fine-Tuned on a Large Language Model {Meta-AI} ({LLaMA}) Using Medical Domain Knowledge},
  author = {Li, Yunxiang and Li, Zihan and Zhang, Kai and Dan, Ruilong and Jiang, Steve and Zhang, You},
  journal = {Cureus},
  year = {2023},
  volume = {15},
  number = {6},
  pages = {e40895},
  doi = {10.7759/cureus.40895},
  url = {https://doi.org/10.7759/cureus.40895}
}

@article{xie2024moepruner,
  title = {{MoE-Pruner}: Pruning Mixture-of-Experts Large Language Model using the Hints from Its Router},
  author = {Xie, Yanyue and Zhang, Zhi and Zhou, Ding and Xie, Cong and Song, Ziang and Liu, Xin and Wang, Yanzhi and Lin, Xue and Xu, An},
  journal = {arXiv preprint arXiv:2410.12013},
  year = {2024},
  url = {https://arxiv.org/abs/2410.12013}
}

@inproceedings{
li2025slimmoe,
title={SlimMoE: Structured Compression of Large MoE Models via Expert Slimming and Distillation},
author={Zichong Li and Chen Liang and Zixuan Zhang and Ilgee Hong and Young Jin Kim and Weizhu Chen and Tuo Zhao},
booktitle={Second Conference on Language Modeling},
year={2025},
url={https://openreview.net/forum?id=oaCUsn391F}
}

@article{ahrac2026routers,
  title = {Routers Learn the Geometry of Their Experts: Geometric Coupling in Sparse Mixture-of-Experts},
  author = {Ahrac, Sagi and Hochwald, Noya and Geva, Mor},
  journal = {arXiv preprint arXiv:2605.12476},
  year = {2026},
  url = {https://arxiv.org/abs/2605.12476}
}

@article{ludziejewski2025decoupled,
  title = {Decoupled Relative Learning Rate Schedules},
  author = {Ludziejewski, Jan and Ma{\l}a{\'s}nicki, Jan and Pi{\'o}ro, Maciej and Krutul, Micha{\l} and Ciebiera, Kamil and Stefaniak, Maciej and Krajewski, Jakub and Sankowski, Piotr and Cygan, Marek and Adamczewski, Kamil and Jaszczur, Sebastian},
  journal = {arXiv preprint arXiv:2507.03526},
  year = {2025},
  url = {https://arxiv.org/abs/2507.03526}
}

@article{hu2026whatgetsactivated,
  title = {What Gets Activated: Uncovering Domain and Driver Experts in {MoE} Language Models},
  author = {Hu, Guimin and Li, Meng and Peng, Qiwei and Hu, Lijie and Xu, Boyan and Cai, Ruichu},
  journal = {arXiv preprint arXiv:2601.10159},
  year = {2026},
  url = {https://arxiv.org/abs/2601.10159}
}

@article{he2026tenp,
  title = {{TENP}: Trapezoidal Expert Neuron Pruning For Mixture-of-Experts},
  author = {He, Jiangyang and Zhu, Shaolin and Xiong, Deyi},
  journal = {arXiv preprint arXiv:2606.09885},
  year = {2026},
  url = {https://arxiv.org/abs/2606.09885}
}

@article{yu2025opencsg,
  title = {{OpenCSG} Chinese Corpus: A Series of High-quality Chinese Datasets for {LLM} Training},
  author = {Yu, Yijiong and Dai, Ziyun and Wang, Zekun and Wang, Wei and Chen, Ran and Pei, Ji},
  journal = {arXiv preprint arXiv:2501.08197},
  year = {2025},
  url = {https://arxiv.org/abs/2501.08197}
}

@inproceedings{geva2021transformer,
  title = {Transformer Feed-Forward Layers Are Key-Value Memories},
  author = {Geva, Mor and Schuster, Roei and Berant, Jonathan and Levy, Omer},
  booktitle = {Proceedings of the 2021 Conference on Empirical Methods in Natural Language Processing},
  year = {2021},
  pages = {5484--5495},
  publisher = {Association for Computational Linguistics},
  doi = {10.18653/v1/2021.emnlp-main.446},
  url = {https://aclanthology.org/2021.emnlp-main.446/}
}

@inproceedings{dai2022knowledge,
  title = {Knowledge Neurons in Pretrained Transformers},
  author = {Dai, Damai and Dong, Li and Hao, Yaru and Sui, Zhifang and Chang, Baobao and Wei, Furu},
  booktitle = {Proceedings of the 60th Annual Meeting of the Association for Computational Linguistics (Volume 1: Long Papers)},
  year = {2022},
  pages = {8493--8502},
  publisher = {Association for Computational Linguistics},
  doi = {10.18653/v1/2022.acl-long.581},
  url = {https://aclanthology.org/2022.acl-long.581/}
}

@inproceedings{meng2022locating,
  title = {Locating and Editing Factual Associations in {GPT}},
  author = {Meng, Kevin and Bau, David and Andonian, Alex and Belinkov, Yonatan},
  booktitle = {Advances in Neural Information Processing Systems},
  year = {2022},
  volume = {35},
  pages = {17359--17372},
  publisher = {Curran Associates, Inc.},
  doi = {10.52202/068431-1262},
  url = {https://papers.nips.cc/paper_files/paper/2022/hash/6f1d43d5a82a37e89b0665b33bf3a182-Abstract-Conference.html}
}

@article{cunningham2023sparse,
  title = {Sparse Autoencoders Find Highly Interpretable Features in Language Models},
  author = {Cunningham, Hoagy and Ewart, Aidan and Riggs, Logan and Huben, Robert and Sharkey, Lee},
  journal = {arXiv preprint arXiv:2309.08600},
  year = {2023},
  eprint = {2309.08600},
  archivePrefix = {arXiv},
  primaryClass = {cs.LG},
  url = {https://arxiv.org/abs/2309.08600}
}

@article{bricken2023monosemanticity,
  title = {Towards Monosemanticity: Decomposing Language Models With Dictionary Learning},
  author = {Bricken, Trenton and Templeton, Adly and Batson, Joshua and Chen, Brian and Jermyn, Adam and Conerly, Tom and Turner, Nick and Anil, Cem and Denison, Carson and Askell, Amanda and Lasenby, Robert and Wu, Yifan and Kravec, Shauna and Schiefer, Nicholas and Maxwell, Tim and Joseph, Nicholas and Hatfield-Dodds, Zac and Tamkin, Alex and Nguyen, Karina and McLean, Brayden and Burke, Josiah E and Hume, Tristan and Carter, Shan and Henighan, Tom and Olah, Christopher},
  journal = {Transformer Circuits Thread},
  year = {2023},
  url = {https://transformer-circuits.pub/2023/monosemantic-features/index.html}
}

@article{templeton2024scaling,
  title = {Scaling Monosemanticity: Extracting Interpretable Features from {Claude 3 Sonnet}},
  author = {Templeton, Adly and Conerly, Tom and Marcus, Jonathan and Lindsey, Jack and Bricken, Trenton and Chen, Brian and Pearce, Adam and Citro, Craig and Ameisen, Emmanuel and Jones, Andy and Cunningham, Hoagy and Turner, Nicholas L and McDougall, Callum and MacDiarmid, Monte and Freeman, C. Daniel and Sumers, Theodore R. and Rees, Edward and Batson, Joshua and Jermyn, Adam and Carter, Shan and Olah, Chris and Henighan, Tom},
  journal = {Transformer Circuits Thread},
  year = {2024},
  url = {https://transformer-circuits.pub/2024/scaling-monosemanticity/index.html}
}

@inproceedings{wang2023interpretability,
  title = {Interpretability in the Wild: a Circuit for Indirect Object Identification in {GPT-2} small},
  author = {Wang, Kevin and Variengien, Alexandre and Conmy, Arthur and Shlegeris, Buck and Steinhardt, Jacob},
  booktitle = {The Eleventh International Conference on Learning Representations},
  year = {2023},
  url = {https://openreview.net/forum?id=NpsVSN6o4ul}
}

@inproceedings{conmy2023automated,
  title = {Towards Automated Circuit Discovery for Mechanistic Interpretability},
  author = {Conmy, Arthur and Mavor-Parker, Augustine and Lynch, Aengus and Heimersheim, Stefan and Garriga-Alonso, Adri\`{a}},
  booktitle = {Advances in Neural Information Processing Systems},
  year = {2023},
  volume = {36},
  pages = {16318--16352},
  publisher = {Curran Associates, Inc.},
  doi = {10.52202/075280-0719},
  url = {https://proceedings.neurips.cc/paper_files/paper/2023/hash/34e1dbe95d34d7ebaf99b9bcaeb5b2be-Abstract-Conference.html}
}

@article{shazeer2017outrageously,
  title = {Outrageously Large Neural Networks: The Sparsely-Gated Mixture-of-Experts Layer},
  author = {Shazeer, Noam and Mirhoseini, Azalia and Maziarz, Krzysztof and Davis, Andy and Le, Quoc and Hinton, Geoffrey and Dean, Jeff},
  journal = {arXiv preprint arXiv:1701.06538},
  year = {2017},
  eprint = {1701.06538},
  archivePrefix = {arXiv},
  url = {https://arxiv.org/abs/1701.06538}
}

@article{zoph2022stmoe,
  title = {{ST-MoE}: Designing Stable and Transferable Sparse Expert Models},
  author = {Zoph, Barret and Bello, Irwan and Kumar, Sameer and Du, Nan and Huang, Yanping and Dean, Jeff and Shazeer, Noam and Fedus, William},
  journal = {arXiv preprint arXiv:2202.08906},
  year = {2022},
  eprint = {2202.08906},
  archivePrefix = {arXiv},
  url = {https://arxiv.org/abs/2202.08906}
}

@article{zhou2022expertchoice,
  title = {Mixture-of-Experts with Expert Choice Routing},
  author = {Zhou, Yanqi and Lei, Tao and Liu, Hanxiao and Du, Nan and Huang, Yanping and Zhao, Vincent and Dai, Andrew and Chen, Zhifeng and Le, Quoc and Laudon, James},
  journal = {arXiv preprint arXiv:2202.09368},
  year = {2022},
  eprint = {2202.09368},
  archivePrefix = {arXiv},
  url = {https://arxiv.org/abs/2202.09368}
}

@inproceedings{ludziejewski2024finegrained,
  title = {Scaling Laws for Fine-Grained Mixture of Experts},
  author = {Ludziejewski, Jan and Krajewski, Jakub and Adamczewski, Kamil and Pi\'{o}ro, Maciej and Krutul, Micha{\l} and Antoniak, Szymon and Ciebiera, Kamil and Kr\'{o}l, Krystian and Odrzyg\'{o}\'{z}d\'{z}, Tomasz and Sankowski, Piotr and Cygan, Marek and Jaszczur, Sebastian},
  booktitle = {Proceedings of the 41st International Conference on Machine Learning},
  year = {2024},
  volume = {235},
  series = {Proceedings of Machine Learning Research},
  pages = {33270--33288},
  publisher = {PMLR},
  url = {https://proceedings.mlr.press/v235/ludziejewski24a.html}
}

@inproceedings{dai2022stablemoe,
  title = {{StableMoE}: Stable Routing Strategy for Mixture of Experts},
  author = {Dai, Damai and Dong, Li and Ma, Shuming and Zheng, Bo and Sui, Zhifang and Chang, Baobao and Wei, Furu},
  booktitle = {Proceedings of the 60th Annual Meeting of the Association for Computational Linguistics (Volume 1: Long Papers)},
  year = {2022},
  pages = {7085--7095},
  publisher = {Association for Computational Linguistics},
  doi = {10.18653/v1/2022.acl-long.489},
  url = {https://aclanthology.org/2022.acl-long.489/}
}

@inproceedings{zhang2022moefication,
  title = {{MoEfication}: Transformer Feed-forward Layers are Mixtures of Experts},
  author = {Zhang, Zhengyan and Lin, Yankai and Liu, Zhiyuan and Li, Peng and Sun, Maosong and Zhou, Jie},
  booktitle = {Findings of the Association for Computational Linguistics: ACL 2022},
  year = {2022},
  pages = {877--890},
  publisher = {Association for Computational Linguistics},
  doi = {10.18653/v1/2022.findings-acl.71},
  url = {https://aclanthology.org/2022.findings-acl.71/}
}

@article{song2026razor,
  title = {{RAZOR}: Pruning Replaceable Experts in {LLMs}},
  author = {Song, Mingyang and Zheng, Mao},
  journal = {arXiv preprint arXiv:2609.30465},
  year = {2026}
}

@inproceedings{lu2024notallexperts,
  title = {Not All Experts are Equal: Efficient Expert Pruning and Skipping for Mixture-of-Experts Large Language Models},
  author = {Lu, Xudong and Liu, Qi and Xu, Yuhui and Zhou, Aojun and Huang, Siyuan and Zhang, Bo and Yan, Junchi and Li, Hongsheng},
  booktitle = {Proceedings of the 62nd Annual Meeting of the Association for Computational Linguistics (Volume 1: Long Papers)},
  year = {2024}
}

@inproceedings{chowdhury2024provably,
  title = {A Provably Effective Method for Pruning Experts in Fine-tuned Sparse Mixture-of-Experts},
  author = {Chowdhury, Mohammed Nowaz Rabbani and Wang, Meng and El Maghraoui, Kaoutar and Wang, Naigang and Chen, Pin-Yu and Carothers, Christopher},
  booktitle = {Proceedings of the 41st International Conference on Machine Learning},
  series = {Proceedings of Machine Learning Research},
  volume = {235},
  year = {2024}
}

@article{liu2024efficientexpert,
  title = {Efficient Expert Pruning for Sparse Mixture-of-Experts Language Models: Enhancing Performance and Reducing Inference Costs},
  author = {Liu, Enshu and Zhu, Junyi and Lin, Zinan and Ning, Xuefei and Blaschko, Matthew B. and Yan, Shengen and Dai, Guohao and Yang, Huazhong and Wang, Yu},
  journal = {arXiv preprint arXiv:2407.00945},
  year = {2024}
}

@article{jaiswal2025fantastic,
  title = {Finding Fantastic Experts in {MoEs}: A Unified Study for Expert Dropping Strategies and Observations},
  author = {Jaiswal, Ajay and Wang, Jianyu and Li, Yixiao and Li, Pingzhi and Chen, Tianlong and Wang, Zhangyang and Wang, Chong and Pang, Ruoming and Du, Xianzhi},
  journal = {arXiv preprint arXiv:2504.05586},
  year = {2025}
}

@article{jha2026halfexperts,
  title = {Half the Experts, All the Code: One-Shot Domain Pruning of Mixture-of-Experts {LLMs} for Coding},
  author = {Jha, Anik},
  journal = {arXiv preprint arXiv:2607.16721},
  year = {2026}
}

@inproceedings{li2024mcsmoe,
  title = {Merge, Then Compress: Demystify Efficient {SMoE} with Hints from Its Routing Policy},
  author = {Li, Pingzhi and Zhang, Zhenyu and Yadav, Prateek and Sung, Yi-Lin and Cheng, Yu and Bansal, Mohit and Chen, Tianlong},
  booktitle = {The Twelfth International Conference on Learning Representations},
  year = {2024}
}

@inproceedings{chen2025hcsmoe,
  title = {Retraining-Free Merging of Sparse {MoE} via Hierarchical Clustering},
  author = {Chen, I-Chun and Liu, Hsu-Shen and Sun, Wei-Fang and Chao, Chen-Hao and Hsu, Yen-Chang and Lee, Chun-Yi},
  booktitle = {Proceedings of the 42nd International Conference on Machine Learning},
  series = {Proceedings of Machine Learning Research},
  year = {2025}
}

@inproceedings{lee2025stun,
  title = {{STUN}: Structured-Then-Unstructured Pruning for Scalable {MoE} Pruning},
  author = {Lee, Jaeseong and Hwang, Seung-won and Qiao, Aurick and Campos, Daniel F. and Yao, Zhewei and He, Yuxiong},
  booktitle = {Proceedings of the 63rd Annual Meeting of the Association for Computational Linguistics (Volume 1: Long Papers)},
  year = {2025}
}

@inproceedings{yang2024moei2,
  title = {{MoE-I}$^2$: Compressing Mixture of Experts Models through Inter-Expert Pruning and Intra-Expert Low-Rank Decomposition},
  author = {Yang, Cheng and Sui, Yang and Xiao, Jinqi and Huang, Lingyi and Gong, Yu and Duan, Yuanlin and Jia, Wenqi and Yin, Miao and Cheng, Yu and Yuan, Bo},
  booktitle = {Findings of the Association for Computational Linguistics: EMNLP 2024},
  year = {2024}
}

@inproceedings{li2026heapr,
  title = {{HEAPr}: Hessian-based Efficient Atomic Expert Pruning in Output Space},
  author = {Li, Ke and Yang, Zheng and Zhou, Zhongbin and Xue, Feng and Jiang, Zhonglin and Wang, Wenxiao},
  booktitle = {The Fourteenth International Conference on Learning Representations},
  year = {2026}
}

@inproceedings{xu2026camera,
  title = {{CAMERA}: Multi-Matrix Joint Compression for {MoE} Models via Micro-Expert Redundancy Analysis},
  author = {Xu, Yuzhuang and Han, Xu and Zhang, Yuanchi and Wang, Yixuan and Liu, Yijun and Ji, Shiyu and Zhu, Qingfu and Che, Wanxiang},
  booktitle = {Proceedings of the AAAI Conference on Artificial Intelligence},
  year = {2026}
}

@inproceedings{dong2025easyep,
  title = {Domain-Specific Pruning of Large Mixture-of-Experts Models with Few-shot Demonstrations},
  author = {Dong, Zican and Peng, Han and Liu, Peiyu and Zhao, Wayne Xin and Wu, Dong and Xiao, Feng and Wang, Zhifeng},
  booktitle = {Advances in Neural Information Processing Systems},
  year = {2025}
}
\bibliographystyle{icml2026}
\clearpage
\appendix
\onecolumn
\raggedbottom
\renewcommand{\topfraction}{0.9}
\renewcommand{\bottomfraction}{0.8}
\renewcommand{\textfraction}{0.08}
\renewcommand{\floatpagefraction}{0.85}
\setcounter{topnumber}{3}
\setcounter{bottomnumber}{2}
\setcounter{totalnumber}{5}
\makeatletter
\setlength{\@fptop}{0pt}
\setlength{\@fpsep}{12pt}
\setlength{\@fpbot}{0pt plus 1fil}
\makeatother
\section{Neuron-Set Analysis Protocol and Additional Results}
\label{app:discovery-protocol}

The neuron-level analysis in \cref{sec:neuron-organization} connects domain-dependent membership with organization across experts.
Understanding this connection requires separating how neuron sets change with sampled text and selection size from how they overlap across domains and spread within a layer.

\subsection{Models and Calibration Pool}

\begin{table}[!htbp]
\centering
\begin{minipage}[t]{0.40\linewidth}
\vspace{0pt}
\caption{Routed FFN structures used for discovery.
$L$ counts observed MoE layers, $E$ is experts per layer, $N$ is intermediate width, and $K$ is experts selected per token.
GLM excludes decoder layer 0.
All models have hidden dimension 2048.}
\label{tab:discovery-models}
\centering
\small
\rowcolors{2}{tabrow}{}
\begin{tabular}{lrrrr}
\toprule
\rowcolor{tabhead}Model & $L$ & $E$ & $N$ & $K$ \\
\midrule
Qwen3 & 48 & 128 & 768 & 8 \\
Qwen3.5 & 40 & 256 & 512 & 8 \\
Qwen3.6 & 40 & 256 & 512 & 8 \\
GLM-4.7 & 46 & 64 & 1536 & 4 \\
Qwen1.5 & 24 & 60 & 1408 & 4 \\
OLMoE & 16 & 64 & 1024 & 8 \\
\bottomrule
\end{tabular}
\end{minipage}\hfill
\begin{minipage}[t]{0.57\linewidth}
\vspace{0pt}
\caption{Dataset sources of the nine task buckets.
Each bucket contributes 128 randomly sampled documents, split into 64 discovery and 64 validation documents.}
\label{tab:discovery-sources}
\centering
\small
\rowcolors{2}{tabrow}{}
\begin{tabularx}{\linewidth}{lL}
\toprule
\rowcolor{tabhead}Bucket & Actual sources \\
\midrule
chat-gen & UltraChat, OASST1, Dolly \\
chat-mcq & CommonsenseQA, SocialIQA, CosmosQA \\
code-gen & Magicoder-Evol, Magicoder-OSS \\
math-gen & GSM8K, Hendrycks MATH, MetaMathQA \\
math-mcq & AQuA, SciQ \\
med-gen & PubMedQA artificial, ChatDoctor \\
med-mcq & MedMCQA, MedQA, PubMedQA labeled \\
zh-gen & Fineweb-Edu-Chinese, Alpaca-ZH \\
zh-mcq & C-Eval dev, CMMLU dev \\
\bottomrule
\end{tabularx}
\end{minipage}
\end{table}

To compare neuron organization across models on the same inputs, the analysis in \cref{sec:discovery-setup} uses a shared document split while preserving each model's native routed structure.

All observed layers use hidden dimension 2048.
Qwen3 refers to Qwen3-30B-A3B \citep{qwen2025qwen3}, while Qwen3.5 and Qwen3.6 are their respective 35B-A3B checkpoints.
GLM-4.7 refers to GLM-4.7-Flash, Qwen1.5 to Qwen1.5-MoE-A2.7B-Chat, and OLMoE to OLMoE-1B-7B-0125-Instruct \citep{muennighoff2025olmoe}.
The observed decoder layers are 0--47 for Qwen3, 0--39 for Qwen3.5 and Qwen3.6, 1--46 for GLM, 0--23 for Qwen1.5, and 0--15 for OLMoE.
Native top-$K$ normalization is enabled for the first four models and disabled for Qwen1.5 and OLMoE.
The effective mixing coefficients retain all native scaling.

The sources in \cref{tab:discovery-sources} cover both generation and multiple-choice inputs.
Chat generation includes UltraChat \citep{ding2023ultrachat} and OpenAssistant \citep{kopf2023openassistant}, while chat multiple-choice uses CommonsenseQA \citep{talmor2019commonsenseqa}, Social IQa \citep{sap2019socialiqa} and Cosmos QA \citep{huang2019cosmosqa}.
Medical generation additionally uses ChatDoctor \citep{li2023chatdoctor}, and code generation uses Magicoder data \citep{wei2024magicoder}.
Mathematical sources include GSM8K \citep{cobbe2021gsm8k}, MATH \citep{hendrycks2021math}, MetaMathQA \citep{yu2024metamath}, AQuA \citep{ling2017aqua} and SciQ \citep{welbl2017sciq}.
Medical sources include PubMedQA \citep{jin2019pubmedqa}, MedMCQA \citep{pal2022medmcqa} and MedQA \citep{jin2021medqa}.
Chinese multiple-choice data use the development splits of C-Eval \citep{huang2023ceval} and CMMLU \citep{li2024cmmlu}.
Chinese generation uses \path{opencsg/Fineweb-Edu-Chinese-V2.1}, whose dataset card cites the OpenCSG Chinese Corpus report \citep{yu2025opencsg}, and \path{shibing624/alpaca-zh}.
The Dolly source is \path{databricks/databricks-dolly-15k}, and the ChatDoctor variant is \path{lavita/ChatDoctor-HealthCareMagic-100k}.
The Chinese web source uses a subset of Fineweb-Edu-Chinese, and the other sources use their training splits.
These buckets span combined input conditions, since mathematics multiple-choice includes science questions and the Chinese buckets vary language as well as content and source.

The calibration data contain 1152 documents per model from nine buckets of an eleven-bucket collection, excluding finance and law.
Each bucket contributes 128 documents.
A fixed shuffle with seed 1234 assigns 64 to discovery and 64 to validation, giving 576 documents on each side.
All six models use the same whole-document split.

The original eleven-bucket collection contains 1408 distinct whitespace-normalized texts.
The discovery and validation samples share no documents under exact or whitespace-normalized matching.
Inputs are processed as raw text without an additional chat template or special tokens and truncated to at most 2048 tokens per document.
Identical documents can yield different token counts across tokenizers.

The native structures give per-layer neuron populations of 98,304 for Qwen3 and GLM, 131,072 for Qwen3.5 and Qwen3.6, 84,480 for Qwen1.5, and 65,536 for OLMoE.
The corresponding 6.25\% sets contain 6144, 8192, 5280 and 4096 neurons.
The 12.5\% sets contain twice as many neurons.
All recorded layers contribute to the model-level mean.

\subsection{Measurement and Aggregation Details}
\begin{table}[t]
\centering
\begin{minipage}[t]{0.575\linewidth}
\vspace{0pt}
\caption{Routing-weighted RAMP overlap (\%) at four selection fractions.
Same compares discovery and validation sets within a domain; Cross compares discovery sets across domains.}
\label{tab:replication-four-rates}
\centering\footnotesize
\setlength{\tabcolsep}{2.4pt}
\rowcolors{3}{}{tabrow}
\begin{tabularx}{\linewidth}{lRRRRRRRR}
\toprule
\rowcolor{tabhead}& \multicolumn{2}{c}{1\%} & \multicolumn{2}{c}{3.125\%} & \multicolumn{2}{c}{6.25\%} & \multicolumn{2}{c}{12.5\%} \\
\cmidrule(lr){2-3}\cmidrule(lr){4-5}\cmidrule(lr){6-7}\cmidrule(lr){8-9}
\rowcolor{tabhead}Model & Same & Cross & Same & Cross & Same & Cross & Same & Cross \\
\midrule
Qwen3 & 54.01 & 9.02 & 61.30 & 12.39 & 66.13 & 16.33 & 72.05 & 23.81 \\
Qwen3.5 & 53.73 & 9.28 & 61.59 & 14.46 & 66.18 & 19.52 & 71.41 & 28.42 \\
Qwen3.6 & 54.10 & 9.21 & 61.91 & 14.21 & 66.40 & 19.25 & 71.51 & 27.98 \\
GLM-4.7 & 60.32 & 12.23 & 66.32 & 14.93 & 69.72 & 18.74 & 73.63 & 25.51 \\
Qwen1.5 & 50.79 & 8.47 & 54.38 & 11.28 & 57.77 & 15.23 & 62.79 & 22.47 \\
OLMoE & 68.07 & 12.64 & 72.84 & 19.61 & 74.33 & 23.68 & 76.42 & 30.88 \\
\bottomrule
\end{tabularx}
\end{minipage}\hfill
\begin{minipage}[t]{0.395\linewidth}
\vspace{0pt}
\centering
\includegraphics[width=\linewidth]{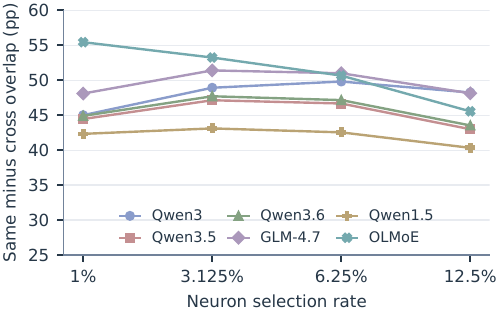}
\captionof{figure}{Same-domain minus cross-domain overlap from \cref{tab:replication-four-rates}; positive gaps persist at every rate.}
\label{fig:app-budget-gap}
\end{minipage}
\end{table}
RAMP in \cref{sec:ramp} combines neuron activations with routing weights.
The unweighted control in Appendix~\ref{app:weighted-control} omits routing-weight magnitude:
\begin{equation}
 I_i^{e,U}(D_d)=\left(\sum_{x\in D_d}\sum_{t\in x}
 R^e(t)a_i^e(t)^2\right)^{1/2}
 \left\|w_i^{e,\mathrm{down}}\right\|_2,
 \label{eq:ramp-unweighted}
\end{equation}
where $R^e(t)$ indicates actual dispatch.
The score therefore retains routing selection and frequency, but not routing-weight magnitude.
Membership, spatial organization and domain enrichment otherwise use routing-weighted RAMP.

For both measurements, forward hooks record squared activations only for experts dispatched by the native model, storing per-document statistics in FP32.
The weighted form accumulates $[pa]^2$, while the unweighted form accumulates $Ra^2$.
Summing these statistics within and across documents before taking the square root and multiplying by the output-projection column norm gives each neuron's accumulated contribution magnitude.
Native shared or dense branches remain active but lie outside the ranked routed-neuron population.

For each domain and layer, the selected set contains $k=\max(1,\operatorname{round}(rEN))$ neurons.
The main RAMP membership comparison uses $r\in\{1\%,3.125\%,6.25\%,12.5\%\}$.
Domain enrichment and fixed-format comparisons use the same four fractions; task matrices and spatial organization retain 6.25\% and 12.5\% only.
The paired scoring control uses 6.25\%.
These selection fractions are distinct from pruning retention rates.
Neurons are ranked by absolute domain scores in stable descending order over flattened expert--neuron coordinates, without cross-domain preference normalization, a minimum routing-count filter or a contribution-quantile threshold.
Tied scores, including zero responses, follow a fixed coordinate order.

Writing $\mathcal L$ for the observed layers, the reported means are
\begin{align}
\overline O_{\mathrm{same}}&=\frac{1}{|\mathcal L|}\sum_{l\in\mathcal L}\frac{1}{9}\sum_d O_{\mathrm{same}}(d,l),\\
\overline O_{\mathrm{cross}}&=\frac{1}{|\mathcal L|}\sum_{l\in\mathcal L}\frac{1}{36}\sum_{d<c}O_{\mathrm{cross}}(d,c,l).
\end{align}
Within each layer, these means average over nine domains for same-domain overlap and 36 unordered pairs for cross-domain overlap, then weight observed layers equally.
Same-domain comparisons use disjoint discovery and validation text, while cross-domain comparisons use discovery sets on both sides.
Normalizing each intersection by one selected set's size makes the two overlaps directly comparable.

\subsection{Collection Configuration}
The four-budget RAMP comparison contains 24 model--budget configurations and uses full-precision aggregate overlap values.
Each model is loaded in BF16 on a single NVIDIA H20 GPU with PyTorch 2.13.0 and Transformers 5.17.0, and squared activation statistics are accumulated in FP32.
All six models process the same frozen 1152 documents, with 64 discovery and 64 validation documents per bucket.

\subsection{Measurement and Selection-Budget Comparisons}
\label{app:discovery-budgets}
The membership comparison in \cref{sec:discovery-replication} depends on both the score definition and how many neurons enter each set.
\Cref{tab:replication-four-rates} reports routing-weighted overlap at all four selection fractions.
Both overlaps grow with selection size in every model while same-domain overlap stays clearly higher, and the gap between them varies only modestly.
From 1\% to 3.125\%, the same--cross gap widens in five models and narrows in OLMoE, from 55.43 to 53.23 percentage points.

Increasing selection from 6.25\% to 12.5\% narrows the same--cross gap in every model (\cref{fig:app-budget-gap}), while the uniform expectation rises with set size.
Larger sets thus share more neurons across domains, but same-domain overlap remains higher at both budgets.

\subsection{Task-Format Comparisons}
\label{app:task-formats}

\subsubsection{Cross-Format Comparisons}
\label{app:cross-format}
\begin{wraptable}{r}{0.4\textwidth}
\vspace{-12pt}
\caption{Task-format overlap (\%) averaged over four selection fractions.
MCQ and Gen compare discovery and validation sets within a bucket; Cross compares the two formats within a category.}
\label{tab:cross-format-averages}
\centering\small
\rowcolors{2}{tabrow}{}
\begin{tabularx}{\linewidth}{lRRR}
\toprule
\rowcolor{tabhead}Model & MCQ & Gen & Cross \\
\midrule
Qwen3 & 66.71 & 60.60 & 34.74 \\
Qwen3.5 & 66.81 & 58.99 & 33.65 \\
Qwen3.6 & 67.23 & 59.56 & 33.78 \\
GLM-4.7 & 70.80 & 63.99 & 37.56 \\
Qwen1.5 & 58.51 & 52.09 & 26.92 \\
OLMoE & 73.81 & 70.36 & 43.83 \\
\bottomrule
\end{tabularx}
\vspace{-8pt}
\end{wraptable}
The task-format analysis in \cref{sec:discovery-replication} compares generation and multiple-choice neuron sets within the same content or language category.
We pair chat-gen with chat-mcq, math-gen with math-mcq, med-gen with med-mcq, and zh-gen with zh-mcq, excluding code because it has no multiple-choice counterpart.
The analysis reuses the existing routing-weighted RAMP overlaps at 1\%, 3.125\%, 6.25\% and 12.5\% selection, without additional model evaluations.

For each model, selection fraction and native layer, we match the generation and multiple-choice buckets by category.
Within-bucket overlap compares discovery and validation sets separately for each format, while cross-format overlap compares the two matched discovery sets.
Each intersection is normalized by the same full-layer selected-set size $k$, as in \cref{eq:discovery-overlap}.
We average the within-bucket overlaps over the four categories separately for generation and multiple choice, and average cross-format overlaps over the four matched pairs, then weight observed layers equally.
\Cref{tab:cross-format-averages} averages each model's results equally over the four selection fractions.
The main-text overall means give the six models equal weight, and the reported ranges are the minimum and maximum of these model-wise four-rate averages, not the extrema across individual rates or layers.

Both formats retain higher within-bucket than cross-format overlap in all 24 model--budget configurations.

\subsubsection{Within-Format Domain Comparisons}
\label{app:fixed-format}
\begin{figure}[t]
\centering
\includegraphics[width=\textwidth]{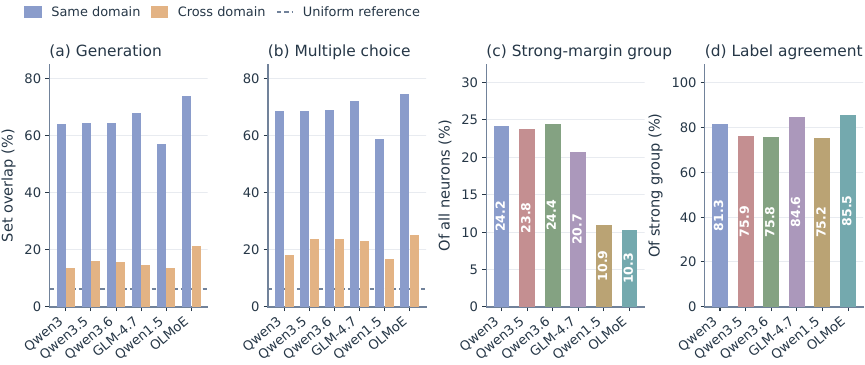}
\caption{Within-format overlap and dominant-domain identity under routing-weighted RAMP.
(a, b) Same-domain replication and cross-domain overlap within generation and multiple-choice inputs at 6.25\% selection, with the dashed line marking the uniform-set reference; \cref{tab:fixed-format-four-rates} reports all four selection fractions.
(c) Size of the strong-margin group relative to all routed neurons.
(d) Validation dominant-label agreement within this group, which is selected on discovery text.}
\label{fig:fixed-format}
\label{fig:dominant-identity}
\end{figure}

The domain comparisons in \cref{sec:discovery-replication} mix generation and multiple-choice inputs.
Restricting comparisons to a single format helps separate domain differences from this format distinction without changing the selected neuron sets.
We summarize the existing routing-weighted sets at 1\%, 3.125\%, 6.25\% and 12.5\% selection, using the same document split without additional model evaluations.

Generation comparisons use chat, code, mathematics, medicine and Chinese generation buckets, while multiple-choice comparisons use chat, mathematics, medicine and Chinese multiple-choice buckets.
For each model, selection fraction and native layer, same-domain discovery--validation overlaps are averaged over five generation or four multiple-choice domains, and cross-domain discovery--discovery overlaps over ten or six unordered within-format pairs, respectively.
Each intersection retains the original full-layer selected-set size $k$ as its denominator, and the resulting layer means are averaged equally over observed layers.
\Cref{tab:fixed-format-four-rates} reports all four fractions, while \cref{fig:fixed-format}a--b shows only the representative 6.25\% setting.

\begin{table}[!htbp]
\caption{Routing-weighted fixed-format overlap (\%) at four selection fractions: six models, four budgets and two formats.
Same compares discovery and validation sets within a domain; Cross compares discovery sets across domains within the same format.
Generation averages five domains or ten unordered pairs per layer, and multiple-choice averages four domains or six unordered pairs, followed by equal layer averaging.}
\label{tab:fixed-format-four-rates}
\centering\small
\setlength{\tabcolsep}{4pt}
\rowcolors{3}{}{tabrow}
\begin{tabularx}{\linewidth}{lRRRRRRRR}
\toprule
\rowcolor{tabhead}& \multicolumn{2}{c}{1\%} & \multicolumn{2}{c}{3.125\%} & \multicolumn{2}{c}{6.25\%} & \multicolumn{2}{c}{12.5\%} \\
\cmidrule(lr){2-3}\cmidrule(lr){4-5}\cmidrule(lr){6-7}\cmidrule(lr){8-9}
\rowcolor{tabhead}Model & Same & Cross & Same & Cross & Same & Cross & Same & Cross \\
\midrule
\multicolumn{9}{c}{Generation} \\
\midrule
Qwen3 & 49.72 & 7.18 & 58.40 & 9.97 & 64.02 & 13.75 & 70.69 & 21.16 \\
Qwen3.5 & 48.39 & 6.31 & 58.24 & 10.62 & 64.34 & 16.02 & 70.49 & 25.69 \\
Qwen3.6 & 48.58 & 5.98 & 58.41 & 10.19 & 64.42 & 15.62 & 70.52 & 25.05 \\
GLM-4.7 & 55.93 & 8.81 & 63.32 & 10.88 & 67.79 & 14.77 & 72.41 & 21.94 \\
Qwen1.5 & 46.95 & 6.36 & 52.59 & 9.36 & 56.95 & 13.75 & 62.61 & 21.61 \\
OLMoE & 66.43 & 8.93 & 72.47 & 16.44 & 74.01 & 21.39 & 75.89 & 29.91 \\
\midrule
\multicolumn{9}{c}{Multiple choice} \\
\midrule
Qwen3 & 59.38 & 12.73 & 64.93 & 15.24 & 68.77 & 18.13 & 73.74 & 24.93 \\
Qwen3.5 & 60.40 & 16.32 & 65.77 & 20.57 & 68.49 & 23.88 & 72.56 & 30.93 \\
Qwen3.6 & 61.00 & 17.42 & 66.28 & 20.71 & 68.88 & 23.77 & 72.75 & 30.61 \\
GLM-4.7 & 65.81 & 18.76 & 70.08 & 20.27 & 72.14 & 23.18 & 75.16 & 28.53 \\
Qwen1.5 & 55.60 & 12.85 & 56.62 & 14.00 & 58.79 & 16.80 & 63.03 & 22.70 \\
OLMoE & 70.11 & 17.01 & 73.30 & 22.06 & 74.74 & 25.29 & 77.09 & 31.55 \\
\bottomrule
\end{tabularx}
\end{table}

All 48 model--budget--format configurations (six models, four budgets and two formats) preserve the same-domain advantage under routing-weighted RAMP.
Across all four fractions and six models, the same--cross gap ranges from 40.59 to 57.50 percentage points for generation and from 40.33 to 53.10 for multiple-choice inputs.

\subsection{Full Task Matrices}
\label{app:task-matrices}
Cross-domain averages in \cref{sec:discovery-replication} summarize sharing but leave individual task relationships unresolved.
The full matrices place these pairwise overlaps alongside same-domain discovery--validation overlap under routing-weighted RAMP at both budgets.

For each layer, let $T_d^{\mathrm{disc}}$ and $T_d^{\mathrm{val}}$ denote the discovery and validation sets.
Averaging their normalized intersections equally over observed layers gives $C^{AA}$, $C^{BB}$ and $C^{AB}$.
The first two compare sets within a text split, while $C^{AB}$ compares discovery domains in rows with validation domains in columns.
The displayed matrix $H$ combines $C^{AB}_{dd}$ on the diagonal with $C^{AA}_{dc}$ elsewhere.

A common bucket order and overlap scale allow direct comparison across matrices.

\begin{figure}[!htbp]
\centering
\includegraphics[width=\textwidth]{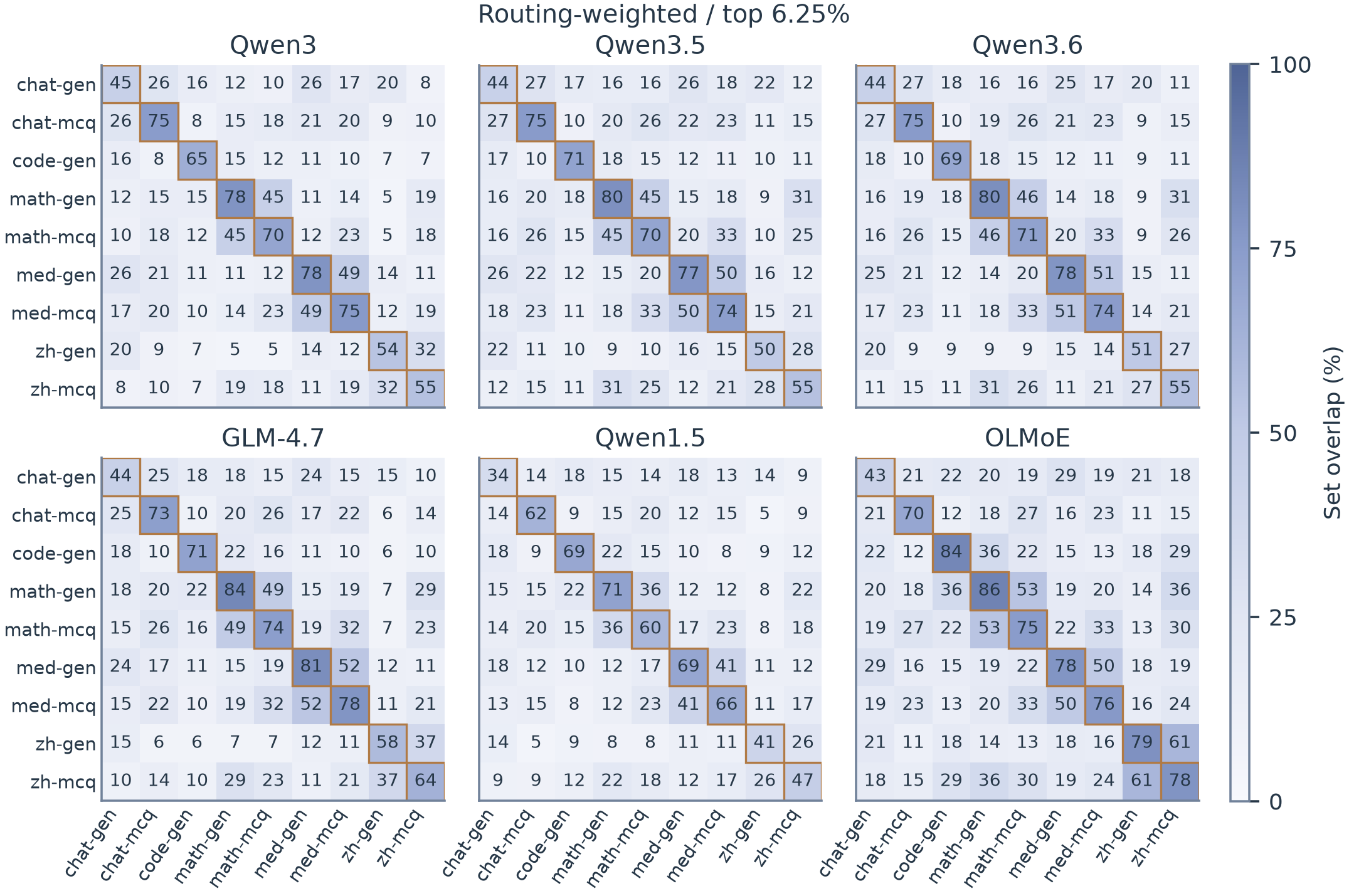}
\caption{Routing-weighted task-set overlap at 6.25\% selection across all six models. Orange outlines identify diagonal entries, which measure same-domain discovery--validation replication. Off-diagonal entries compare discovery sets across domains. Cell annotations are percentages, rounded only for display. All panels use a 0--100\% color scale.}
\label{fig:weighted-matrix}
\end{figure}

\begin{figure}[!htbp]
\centering
\includegraphics[width=\textwidth]{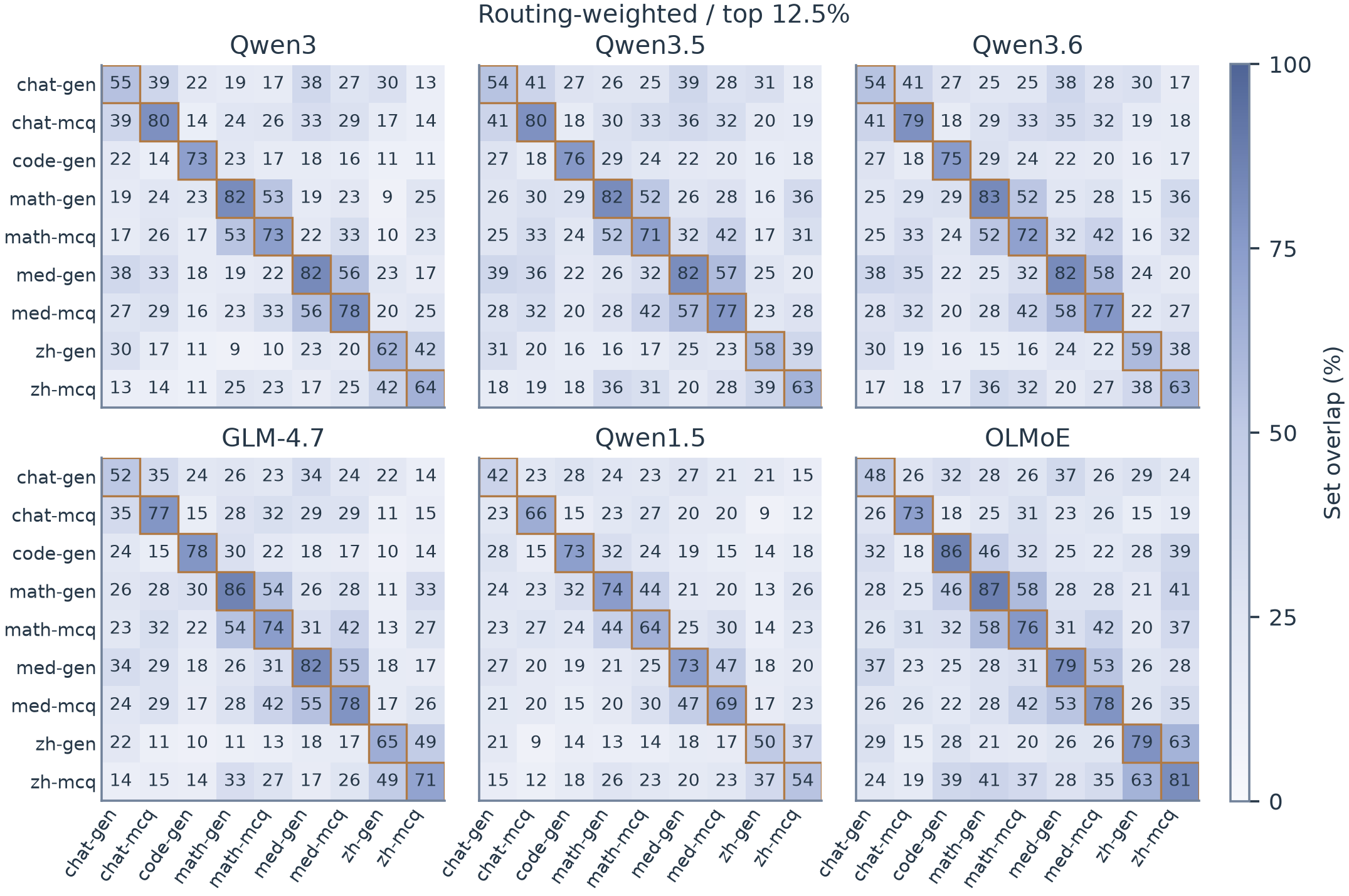}
\caption{Routing-weighted task-set overlap at 12.5\% selection. The uniform-set reference is 12.5\%.}
\label{fig:weighted-matrix-large}
\end{figure}

The 12 displayed composite matrices combine the comparisons of interest; the corresponding $BB$/$AB$ matrices separate comparisons within validation text from those across splits.
The $AA$ matrices follow from $H$ by replacing its diagonal with ones.
Model means computed from $H$ differ from the aggregate overlap means by at most $1.9\times10^{-8}$, consistent with floating-point reduction differences.

At 6.25\% selection, the medical generation and multiple-choice pair has the largest off-diagonal overlap in five models, ranging from 40.65\% to 52.15\%.
In OLMoE, the largest overlap joins the two Chinese buckets at 60.63\%.
The strongest cross-domain sharing in each model therefore connects inputs from the same content or language category across task formats.

\subsection{Dominant-Domain Identity}
\label{app:identity}
The overlapping sets in \cref{sec:discovery-replication} allow neurons to belong to several domains.
Assigning each neuron its highest-scoring domain gives a complementary view.
For neurons with a clear discovery-side preference, group size and validation label agreement distinguish its prevalence from its consistency across texts.

For layer $l$, let $j$ be a flattened expert--neuron index and let $D_d^A$ and $D_d^B$ denote the discovery and validation texts from domain $d$, respectively.
Define the neuron's discovery and validation dominant-domain labels by $u^A_{lj}=\arg\max_d I_{lj}(D_d^A)$ and $u^B_{lj}=\arg\max_d I_{lj}(D_d^B)$.
If $v_1^A$ and $v_2^A$ are its largest and second-largest discovery scores, the strong-margin group is
\begin{equation}
G=\left\{(l,j)\ \middle|\ \frac{v^A_{1,lj}}{\max(v^A_{2,lj},10^{-9})}\ge2\right\}.
\end{equation}
Group size is measured relative to all routed neurons, while label agreement is measured within $G$.
Unlike the equal-layer overlap averages, these proportions pool counts across layers before division.
All neurons enter this analysis independently of the selected-set budget.

The strong-margin group occupies 10.29--24.42\% of all neurons under routing-weighted scoring (\cref{fig:dominant-identity}c).
Validation dominant-label agreement within these groups is 75.24--85.46\% (\cref{fig:dominant-identity}d).
These labels identify each neuron's maximum-scoring domain while allowing responses to other domains.

Qwen3.5 and Qwen3.6 each have 59,093 all-zero neurons on each split, out of 5,242,880 neurons in total.
Overall agreement includes these neurons, whose ties follow fixed bucket order, while 5,183,787 neurons have positive scores on both splits.

The six routing-weighted identity-transition count matrices and 30 group summaries include zero and tie diagnostics for each model.
Matrix totals count all neurons, and diagonal totals count those retaining the same label.

\subsection{Spatial Organization and Count-Preserving Replication}
\label{app:spatial}
The spatial analysis in \cref{sec:discovery-spatial} separates the concentration of a domain's selected-neuron counts across experts from membership within each expert.
Similar expert-level counts can arise from different selected neurons, so both distributions and membership need to be compared across text samples.

\paragraph{Ranked selected-neuron coverage.}
For a fixed domain, layer and selection rate, let $c_e^{\mathrm{disc}}$ count all neurons in the discovery top set $T_d^{\mathrm{disc}}$ that belong to expert $e$.
The denominator is the entire selected set, $k_{\mathrm{sel}}=|T_d^{\mathrm{disc}}|=\sum_e c_e^{\mathrm{disc}}$, including neurons also selected for other domains; it is not the domain-exclusive subset $U_d$.
Sort the $E$ counts within that layer as $c_{(1)}^{\mathrm{disc}}\ge\cdots\ge c_{(E)}^{\mathrm{disc}}$ and define
\begin{equation}
 G(k)=\frac{\sum_{j=1}^{k}c_{(j)}^{\mathrm{disc}}}{k_{\mathrm{sel}}},
 \qquad k=0,\ldots,E.
 \label{eq:selected-neuron-coverage}
\end{equation}
Here $k$ counts ranked experts, whereas $k_{\mathrm{sel}}$ counts selected neurons.
The horizontal axis is $k/E$ and the vertical axis is the covered fraction $G(k)$ of the domain's selected set, with $G(0)=0$ and $G(E)=1$.
The uniform-allocation reference $G_{\mathrm{uniform}}(k)=k/E$ corresponds to spreading the selected set evenly over all experts.
Counts are sorted separately for every domain--layer unit before averaging coverage equally over the nine buckets and native layers; expert identities are never pooled across layers.
Zero-count experts remain in the population, and ties can follow expert ID without changing cumulative coverage.

\begin{wraptable}{r}{0.45\textwidth}
\vspace{-12pt}
\caption{Routing-weighted selected-neuron coverage (\%) of the top $h=\lceil0.1E\rceil$ experts.
$h/E$ is the actual expert fraction and also the coverage under uniform allocation.}
\label{tab:top-coverage}
\centering\footnotesize
\setlength{\tabcolsep}{3pt}
\rowcolors{2}{tabrow}{}
\begin{tabularx}{\linewidth}{lRRRRR}
\toprule
\rowcolor{tabhead}Model & $h/E$ & 1\% & 3.125\% & 6.25\% & 12.5\% \\
\midrule
Qwen3 & 13/128 & 76.53 & 72.32 & 67.22 & 57.36 \\
Qwen3.5 & 26/256 & 82.72 & 78.68 & 73.24 & 61.60 \\
Qwen3.6 & 26/256 & 83.70 & 79.69 & 74.16 & 62.27 \\
GLM-4.7 & 7/64 & 75.73 & 73.59 & 68.73 & 58.74 \\
Qwen1.5 & 6/60 & 53.27 & 47.53 & 42.47 & 35.98 \\
OLMoE & 7/64 & 74.58 & 70.71 & 64.29 & 53.79 \\
\bottomrule
\end{tabularx}
\vspace{-8pt}
\end{wraptable}
The nominal top-10\% summary uses $h=\lceil0.1E\rceil$ and reports $G(h)$ together with its actual expert fraction and uniform reference $h/E$.
For Qwen3 this is $13/128=10.15625\%$; for Qwen3.5 and Qwen3.6 it is $26/256=10.15625\%$; for GLM-4.7 and OLMoE it is $7/64=10.9375\%$; and for Qwen1.5 it is $6/60=10\%$.
Rounding upward changes the expert fraction by $0\le h/E-0.1<1/E$, so the reference must not be replaced by a common hard 10\%.
The neuron selection rates of 6.25\% and 12.5\% are separate from these expert fractions.

\begin{figure}[!htbp]
\centering
\includegraphics[width=\textwidth]{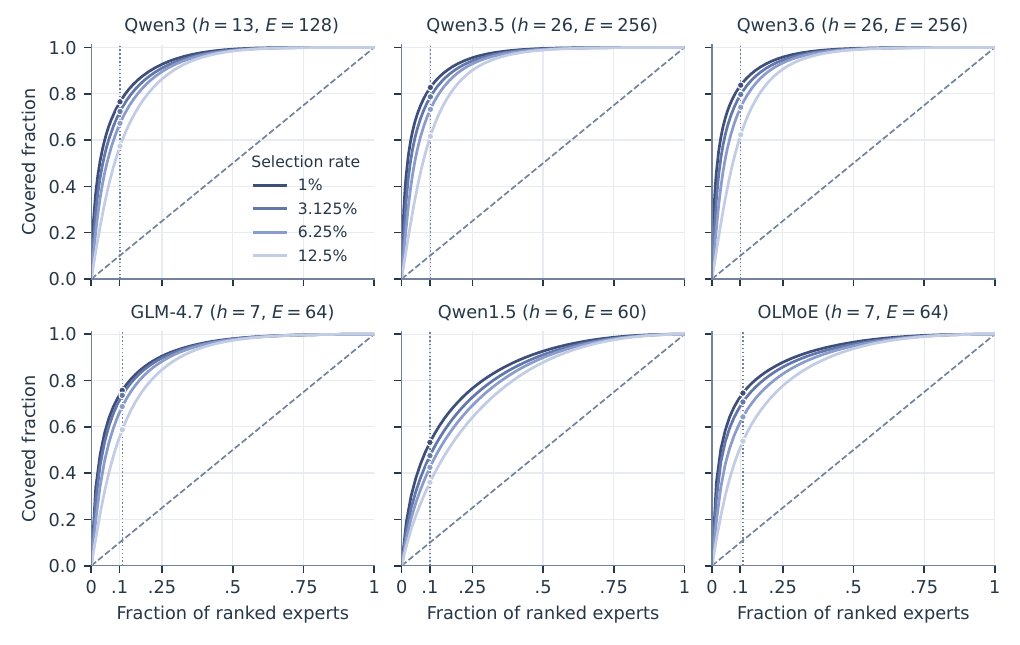}
\caption{Ranked selected-neuron coverage curves for the routing-weighted discovery sets.
The horizontal axis is the fraction of experts included after sorting by selected-neuron count, and the vertical axis is the covered fraction $G$ of the domain's selected set.
Darker blue indicates a smaller selection rate.
Dots mark the top-10\% point at the actual fraction $h/E$, indicated by the dotted vertical line, and the dashed diagonal is the uniform-allocation reference.
Curves are computed within each domain--layer unit and averaged equally over layers and buckets.}
\label{fig:top-coverage-curves}
\end{figure}

\Cref{tab:top-coverage} lists the top-10\% coverage for every model and selection rate, and \cref{fig:top-coverage-curves} shows the corresponding mean curves.
Only the discovery top sets enter this analysis, while the validation sets are used in the count-preserving replication below.
Each curve rises steeply over the first ranked experts and approaches full coverage well before the last one, so a minority of experts holds most of a domain's selected neurons while many experts contain few or none.
Coverage at the top-10\% point decreases from 1\% to 12.5\% selection in every model, as larger selected sets extend into more experts.
Qwen1.5 shows the weakest concentration with 35.98--53.27\% coverage, whereas Qwen3.5 and Qwen3.6 reach 61.60--83.70\%.

\begin{figure}[t]
\centering
\includegraphics[width=\textwidth]{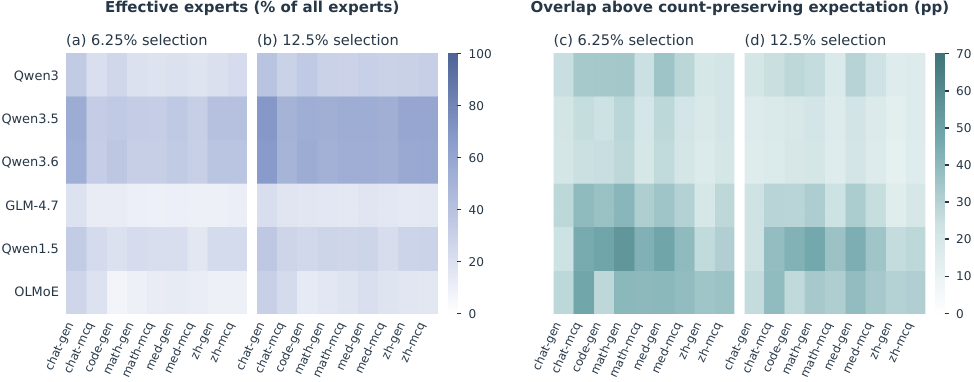}
\caption{Routing-weighted spatial statistics for each model and bucket at 6.25\% and 12.5\% selection, as layer means.
(a, b) Normalized effective expert count, the inverse concentration of selected-neuron shares across experts.
(c, d) Observed overlap above the count-preserving expectation, which is positive in every reported layer--bucket cell at both budgets.}
\label{fig:spatial-grid}
\end{figure}

\paragraph{Effective expert count.}
This auxiliary diagnostic, shown by domain in \cref{fig:spatial-grid}a--b, measures how widely the selected neurons spread across experts.
With $q_e^{\mathrm{disc}}=c_e^{\mathrm{disc}}/k_{\mathrm{sel}}$ the share of expert $e$ in the discovery set, the normalized effective expert count is
\begin{equation}
\frac{N_{\mathrm{eff}}^{\mathrm{disc}}}{E}
=\frac{1}{E\sum_e(q_e^{\mathrm{disc}})^2}.
\label{eq:spatial-auxiliary}
\end{equation}
A perfectly uniform distribution across all experts gives $N_{\mathrm{eff}}^{\mathrm{disc}}/E=1$.
At 6.25\% selection, model-average normalized effective expert counts range from 14.11\% to 37.96\%, compared with this 100\% uniform reference.
The selected neurons therefore span multiple experts but are concentrated within a subset of the expert population.

\paragraph{Count-preserving replication.}
The reference in \cref{eq:expert-count-null} fixes both samples' per-expert selected-neuron counts, $c_e^{\mathrm{disc}}$ and $c_e^{\mathrm{val}}$, and assumes independent uniform selection within each expert.
All nonlinear statistics are computed separately for each native layer and bucket before equal averaging.
The routing-weighted comparison is shown in \cref{fig:spatial-reference}b, with domain-resolved values in \cref{fig:spatial-grid}c--d.

At each of the two budgets, all 1926 routing-weighted layer--bucket cells have positive excess overlap.
At 6.25\% routing-weighted selection, Qwen3 has 66.13\% observed overlap against a 37.70\% count-preserving expectation.
The corresponding values for Qwen1.5 are 57.77\% and 17.66\%.
Membership therefore remains consistent beyond what the allocation of selected-neuron counts across experts alone predicts.
The selected sets contain no zero-score neurons, and cutoff tie counts are recorded separately.

\subsection{Routing Contribution to Domain-Dependent Selection}
\label{app:routing-contribution}
RAMP in \cref{eq:ramp-weighted} accumulates scores only over tokens dispatched to an expert and scales each contribution by the routing weight $p^e(t)$.
Because routers already show domain preferences \citep{muennighoff2025olmoe}, part of the domain dependence of the selected sets could come from routing rather than from the neurons themselves.
This section estimates how much of the observed selection routing alone can account for.

\paragraph{Routing factor shared within an expert.}
Squaring \cref{eq:ramp-weighted} gives
\begin{equation}
 I_i^e(D_d)^2=\left\|w_i^{e,\mathrm{down}}\right\|_2^2\sum_{x\in D_d}\sum_{t\in x}p^e(t)^2a_i^e(t)^2.
\end{equation}
For each token, the dispatch decision and $p^e(t)$ are common to all $N$ neurons of expert $e$, and the column norm does not depend on the domain.
Routing thus determines how much each token counts within an expert, whereas the order of neurons inside the expert follows their activations on those tokens.
If domain dependence came only from routing, neurons within an expert would keep the same relative order for every domain, and domains would differ only in how many neurons each expert contributes to the selected set.

\paragraph{Routing-only bound on cross-domain overlap.}
Under this hypothesis the selected neurons of an expert are nested across domains.
Writing $c_e^{\mathrm{disc}}(d)$ for the number of neurons of expert $e$ in the discovery top set of domain $d$, the cross-domain overlap in \cref{eq:discovery-overlap} would then equal
\begin{equation}
 O_{\mathrm{route}}(d,c)=\frac1k\sum_e\min\!\left(c_e^{\mathrm{disc}}(d),c_e^{\mathrm{disc}}(c)\right).
 \label{eq:routing-only-bound}
\end{equation}
This value is also the largest overlap that any two sets with these per-expert counts can reach, so it is generous to the routing explanation.
The per-expert counts themselves depend on activations and column norms as well as routing, which makes the bound more generous still.
We compute $O_{\mathrm{route}}$ from the complete discovery counts used in Appendix~\ref{app:spatial} and average it over the 36 domain pairs and native layers in the same way as the observed overlap.

\Cref{tab:routing-bound} shows that observed cross-domain overlap stays well below this bound in all 24 model--budget configurations.
Observed values of 8.47--30.88\% correspond to 20.39--74.95\% of bounds ranging from 21.68\% to 53.59\%.
The shortfall is largest at 1\% selection, where the most important neurons are compared.
Observed overlap still exceeds the count-preserving random expectation $\frac{1}{Nk}\sum_e c_e^{\mathrm{disc}}(d)c_e^{\mathrm{disc}}(c)$, which ranges from 1.90\% to 23.65\%, in every configuration.
Domains therefore select largely different neurons even within the experts they both use, sharing only part of them.
Sampling variation between disjoint documents also lowers overlap, yet the same document split leaves same-domain overlap at 50.79--76.42\%, well above the cross-domain values.
The domain dependence of the selected sets therefore extends to which neurons are selected inside each expert, beyond the expert-level allocation that routing can produce.

\begin{table}[!htbp]
\caption{Observed cross-domain overlap (Obs.) and the routing-only bound (Bound) in \cref{eq:routing-only-bound}, in percent.
Both quantities use routing-weighted discovery top sets and are averaged over the 36 domain pairs within each native layer and then over layers.
The bound is the overlap expected if neurons within each expert kept the same order for every domain.}
\label{tab:routing-bound}
\centering\small
\rowcolors{3}{}{tabrow}
\begin{tabularx}{\linewidth}{lRRRRRRRR}
\toprule
\rowcolor{tabhead}& \multicolumn{2}{c}{1\%} & \multicolumn{2}{c}{3.125\%} & \multicolumn{2}{c}{6.25\%} & \multicolumn{2}{c}{12.5\%} \\
\cmidrule(lr){2-3}\cmidrule(lr){4-5}\cmidrule(lr){6-7}\cmidrule(l){8-9}
\rowcolor{tabhead}Model & Obs. & Bound & Obs. & Bound & Obs. & Bound & Obs. & Bound \\
\midrule
Qwen3 & 9.02 & 22.86 & 12.39 & 25.63 & 16.33 & 28.92 & 23.81 & 34.97 \\
Qwen3.5 & 9.28 & 22.51 & 14.46 & 26.92 & 19.52 & 31.20 & 28.42 & 38.17 \\
Qwen3.6 & 9.21 & 21.68 & 14.21 & 26.03 & 19.25 & 30.39 & 27.98 & 37.33 \\
GLM-4.7 & 12.23 & 27.75 & 14.93 & 28.56 & 18.74 & 31.78 & 25.51 & 37.37 \\
Qwen1.5 & 8.47 & 41.52 & 11.28 & 44.92 & 15.23 & 48.51 & 22.47 & 53.59 \\
OLMoE & 12.64 & 30.17 & 19.61 & 35.66 & 23.68 & 39.94 & 30.88 & 46.63 \\
\bottomrule
\end{tabularx}
\end{table}

\paragraph{Removing routing-weight magnitude.}
The unweighted score in \cref{eq:ramp-unweighted} removes $p^e(t)$ while keeping dispatch.
At 6.25\% and 12.5\% selection, it gives same-domain overlap of 67.53--81.03\% and cross-domain overlap of 18.90--36.65\%, so the same-domain advantage persists in all 12 model--budget configurations.
Its observed overlap also exceeds the count-preserving expectation in \cref{eq:expert-count-null} in every layer--bucket cell, with model-average excess of 25.91--47.61 percentage points at 6.25\%.
The domain dependence of the selected sets therefore does not require routing-weight magnitude.
Dispatch remains in both scores because a neuron contributes only on tokens routed to its expert, and the routing-only bound above tests whether dispatch alone can produce the observed selection.

\subsection{Token Identity and Domain Dependence}
\label{app:shared-token}
Routing has been linked to token identity \citep{pmlr-v235-xue24c,wang2026mythexpertspecializationmoes}, while experts also show domain preferences \citep{muennighoff2025olmoe}.
Since domains differ in vocabulary, the domain dependence of the selected sets could partly reflect which token types each domain contains.
Holding token types fixed across domains separates these two contributions.

We take the token types that occur in all 18 text groups, the discovery and validation splits of the nine buckets, without relaxing this intersection.
Only 6 to 11 token types remain, covering 8.70--16.27\% of the token positions processed in the forward passes.
RAMP is then accumulated over these positions alone, and the same-domain and cross-domain overlaps in \cref{eq:discovery-overlap} are recomputed at 6.25\% selection.

\Cref{tab:shared-token} compares both overlaps under all tokens and under shared tokens.
Restricting to shared tokens raises cross-domain overlap in every model, for example from 23.68\% to 48.24\% in OLMoE, so common token types account for part of the neurons that domains share.
Same-domain overlap rises as well, and the same--cross gap remains 37.35--42.85 percentage points, 73.9--90.6\% of its value under all tokens.
The same token types therefore select different neurons in different domains, which places domain dependence in the context in which tokens appear.

Both observations fit the neuron-level organization in \cref{sec:neuron-organization}.
Token identity contributes to the neurons shared across domains, as token-level accounts of routing suggest, while domain-specific neurons within mixed experts carry the remaining dependence and correspond to the routing preferences in \cref{sec:exclusive-enrichment}.
A purely token-level account would leave little same--cross gap on shared tokens, and a purely expert-level account would not explain why the experts holding these neurons serve several domains.

\begin{table}[!htbp]
\caption{Same-domain and cross-domain overlap at 6.25\% selection under all tokens and under token types shared by all 18 text groups, in percent.
Gap retained is the shared-token same--cross gap divided by the all-token gap, and the last two columns give the shared tokens' share of token positions and their number of types.}
\label{tab:shared-token}
\centering\small
\rowcolors{3}{}{tabrow}
\begin{tabularx}{\linewidth}{lRRRRRRR}
\toprule
\rowcolor{tabhead}& \multicolumn{2}{c}{All tokens} & \multicolumn{2}{c}{Shared tokens} & & & \\
\cmidrule(lr){2-3}\cmidrule(lr){4-5}
\rowcolor{tabhead}Model & Same & Cross & Same & Cross & Gap retained & Share & Types \\
\midrule
Qwen3 & 66.13 & 16.33 & 71.67 & 28.82 & 86.0 & 12.13 & 10 \\
Qwen3.5 & 66.18 & 19.52 & 72.67 & 30.48 & 90.4 & 16.27 & 11 \\
Qwen3.6 & 66.40 & 19.25 & 72.76 & 30.05 & 90.6 & 16.27 & 11 \\
GLM-4.7 & 69.72 & 18.74 & 75.03 & 32.97 & 82.5 & 8.70 & 6 \\
Qwen1.5 & 57.77 & 15.23 & 73.62 & 36.27 & 87.8 & 12.13 & 10 \\
OLMoE & 74.33 & 23.68 & 85.66 & 48.24 & 73.9 & 10.66 & 8 \\
\bottomrule
\end{tabularx}
\end{table}

\subsection{Per-Dispatch Importance}
\label{app:dispatch-normalized}
RAMP in \cref{eq:ramp-weighted} accumulates contributions over the tokens dispatched to an expert, so an expert's neurons gain importance both from how strongly they respond to each token and from how many tokens of a domain the expert receives.
The second factor is a genuine part of a neuron's role in the model, yet separating it from the first shows whether the organization in \cref{sec:neuron-organization} already appears in the contribution per processed token.

We therefore divide each neuron's score by the square root of the number of tokens of the same text dispatched to its expert,
\begin{equation}
 \bar I_i^e(D_d)=\frac{I_i^e(D_d)}{\sqrt{C_d^e(D_d)}},
 \label{eq:ramp-per-dispatch}
\end{equation}
which is the root-mean-square contribution per dispatched token and leaves experts receiving no tokens at zero.
Discovery and validation scores are each normalized by their own dispatch counts, so the two splits remain independent.
We then reselect the top 6.25\% of neurons in every layer and recompute expert coverage, same-domain and cross-domain overlap, the enriched lists and their correspondence with validation routing.
The routing measures $\Delta r$ and $\Delta_H$ keep their original definition, because routing preference is the quantity being compared.
No additional forward passes are needed, and the unnormalized statistics reproduce the values in the main text.

\Cref{tab:dispatch-normalized} reports both versions.
Domain specialization is preserved under per-dispatch importance.
Same-domain overlap exceeds cross-domain overlap in all six models, with gaps of 27.1--39.6 percentage points, so most of the domain dependence resides in how strongly neurons respond to each token.
Selected neurons remain concentrated, as the top 10\% of experts still cover 36.40--49.95\% of each domain's selected neurons, 3.6--5.0 times the uniform reference.
The enriched experts also continue to track routing, since the median correlation between $F$ and $\Delta r$ is positive in every model at 0.159--0.379, and the enriched lists still receive 1.93--15.40 percentage points more dispatch share from their own domain than from other domains.

Normalization removes the differences in how often experts process each domain, which is itself a part of domain-dependent computation, so these values form a conservative estimate of the organization.
That all three properties remain under this estimate shows that domain-specific neurons are distinguished by their contribution to each processed token, and that the correspondence between neuron composition and routing reflects this per-token organization in addition to how often experts are used.

\begin{table}[!htbp]
\caption{Statistics at 6.25\% selection with RAMP (Raw) and with per-dispatch importance in \cref{eq:ramp-per-dispatch} (Norm.).
Coverage is the share of each domain's selected neurons held by the top 10\% of experts, overlaps and coverage are in percent, $\widetilde\rho$ is the median correlation between $F$ and $\Delta r$, and $\Delta_H$ is in percentage points.}
\label{tab:dispatch-normalized}
\centering\small
\rowcolors{3}{}{tabrow}
\begin{tabularx}{\linewidth}{lRRRRRRRRRR}
\toprule
\rowcolor{tabhead}& \multicolumn{2}{c}{Coverage} & \multicolumn{2}{c}{Same} & \multicolumn{2}{c}{Cross} & \multicolumn{2}{c}{$\widetilde\rho$} & \multicolumn{2}{c}{$\Delta_H$} \\
\cmidrule(lr){2-3}\cmidrule(lr){4-5}\cmidrule(lr){6-7}\cmidrule(lr){8-9}\cmidrule(l){10-11}
\rowcolor{tabhead}Model & Raw & Norm. & Raw & Norm. & Raw & Norm. & Raw & Norm. & Raw & Norm. \\
\midrule
Qwen3 & 67.22 & 41.21 & 66.13 & 49.21 & 16.33 & 21.45 & 0.667 & 0.277 & 26.01 & 12.50 \\
Qwen3.5 & 73.24 & 49.95 & 66.18 & 51.69 & 19.52 & 23.96 & 0.567 & 0.320 & 19.19 & 15.03 \\
Qwen3.6 & 74.16 & 49.18 & 66.40 & 50.17 & 19.25 & 23.11 & 0.568 & 0.296 & 20.71 & 15.40 \\
GLM-4.7 & 68.73 & 47.97 & 69.72 & 59.54 & 18.74 & 27.34 & 0.650 & 0.379 & 19.61 & 13.98 \\
Qwen1.5 & 42.47 & 36.40 & 57.77 & 55.23 & 15.23 & 15.61 & 0.368 & 0.159 & 3.65 & 1.93 \\
OLMoE & 64.29 & 39.40 & 74.33 & 64.36 & 23.68 & 26.66 & 0.636 & 0.200 & 13.51 & 6.58 \\
\bottomrule
\end{tabularx}
\end{table}

\subsection{Exclusive-Set Concentration and Domain Enrichment}
\label{app:exclusive-enrichment}
Many selected neurons from a domain may reflect an expert's high selected-neuron count overall.
To distinguish concentration from domain enrichment, as in \cref{sec:exclusive-enrichment}, we compare exclusive membership with its parent set and each expert's domain composition.

The analysis uses routing-weighted RAMP on the same 64 discovery documents per bucket, retaining the nine original buckets and all native routed layers.
At each of the four selection rates, exclusivity is recomputed from the corresponding discovery top sets.
This membership criterion differs from the dominant-domain argmax assignment in Appendix~\ref{app:identity}: it identifies coordinates selected for exactly one of the nine buckets, rather than assigning every coordinate a label.
All 1926 domain--layer exclusive sets are nonempty at each rate.

\subsubsection{Conditional Domain Composition}
An expert may hold many exclusive neurons from several domains.
The common-pool comparison in \cref{sec:exclusive-enrichment} accounts for this overall count when assessing whether a particular domain is disproportionately represented.

Four hypergeometric references make this distinction explicit.
Writing $c^e$ and $x^e$ for parent and exclusive counts, and $k=\sum_e c^e$ and $m=\sum_e x^e$, the full-set uniform reference is $\operatorname{Hypergeom}(EN,N,k)$ evaluated at $c^e$.
Restricting attention to exclusive neurons gives $\operatorname{Hypergeom}(EN,N,m)$ at $x^e$, while conditioning on their parent top set gives $\operatorname{Hypergeom}(k,c^e,m)$ at $x^e$.
For domain composition, the reference is $\operatorname{Hypergeom}(S,s^e,m_d)$ at $x_d^e$, where $s^e$ pools mutually disjoint exclusive sets across domains and $S=\sum_e s^e$.
This common pool includes the target domain.

Upper-tail probabilities $\Pr(Z\ge z)$ are BH-adjusted separately for each score, model, selection rate and reference over all finite domain--layer--expert tests.
The family is not restricted to a layer or filtered by validation routing; finite probabilities remain in it even where $F$ is undefined.
Composition-enriched experts satisfy $F_d^e>1$ and $q_d^e<0.05$, defining the discovery list $H_d$ and its carried share
\begin{equation}
M_{dl}=\frac{\sum_{e\in H_d}x_d^e}{m_d}.
\end{equation}
$M$ is undefined when $m_d=0$ and zero for an empty list with positive $m_d$.
The list fraction $|H_d|/E$ has a different denominator and is reported separately.

\subsubsection{Selection-Rate Sensitivity}
\Cref{tab:enrichment-rates} shows how the carried share changes with the selection threshold.
All 1926 units contain an enriched expert at each rate, while model-average $M$ increases from 58.18--78.29\% at 1\% selection to 66.90--85.89\% at 12.5\%.
The exclusive fraction moves in the opposite direction, from 46.50--63.53\% to 18.66--24.79\%.
A stricter threshold therefore yields a more exclusive selected set without necessarily strengthening its expert-composition or routing association.

\subsubsection{Concentration Relative to Parent Sets}
Concentration asks a different question from domain composition: how unevenly does a set span experts?
The parent and exclusive Herfindahl--Hirschman indices are $H_T=\sum_e(c^e/k)^2$ and $H_U=\sum_e(x^e/m)^2$.
For a uniform size-$m$ subset of the parent set, the expected concentration is
\begin{equation}
\mathbb E H_0=H_T+(1-H_T)\frac{k-m}{m(k-1)}.
\end{equation}
At 6.25\% selection, the mean within-unit ratios $H_U/\mathbb E H_0$ are 1.545, 1.617, 1.657, 1.384, 1.081 and 1.119 in the model order of \cref{tab:exclusive-enrichment}.
The highest-count $\lceil0.1E\rceil$ experts carry 43.57--79.36\% of exclusive neurons on average; these count-selected experts need not be the composition-enriched list.
The accompanying concentration test uses 999 multivariate-hypergeometric size-matched samples and upper-tail probability $(1+\#\{H_b\ge H_U-10^{-14}\})/1000$.
Its BH family spans domain--layer units within each model and rate, separately from the expert-composition tests and without selecting units for routing analysis.

\subsection{Domain Enrichment and Validation Routing Preferences}
\label{app:router-link}
Neuron composition alone does not establish whether an expert is preferentially selected for the same domain.
The comparison in \cref{sec:exclusive-enrichment} relates discovery-side enrichment to validation routing, using raw neuron counts as a reference for what composition adds.

\subsubsection{Fixed Discovery Structure and Routing Measurements}
All top sets, exclusive counts, enrichment values and lists are defined on discovery text and remain fixed during validation.
Each bucket contributes 64 documents on each side, with native layers and input processing as in Appendix~\ref{app:discovery-protocol}.
The primary routing measure is dispatch-edge share in \cref{eq:domain-routing-preference}.
An auxiliary measure replaces dispatch counts with summed effective routing weights, normalizing within each domain before averaging the other eight domains; it does not change the discovery membership.

Enrichment requires positive $s^e$, $m_d$ and $S$.
A zero target count with positive denominators gives $F=0$, whereas an empty expert exclusive pool gives undefined $F$.
At 6.25\%, there are 130, 628, 636, 51, 1 and 9 empty-pool layer--experts in the six models, respectively.
Domain-exclusive and layer-pool totals are positive.

\subsubsection{Association Units and Matched Comparisons}
Each domain--layer correlation compares discovery $F$ with validation routing across native experts.
Spearman correlations use average ranks, at least three finite pairs and nonconstant vectors, leaving undefined values missing.
Model summaries take the median across domain--layer units.
All 1926 primary correlations are valid at 6.25\%; nine Qwen1.5 units have negative relative-preference correlations.
\Cref{tab:router-link-associations} reports interquartile ranges, negative counts and the auxiliary routing-weight measure.

\begin{figure}[!htbp]
\centering
\includegraphics[width=\textwidth]{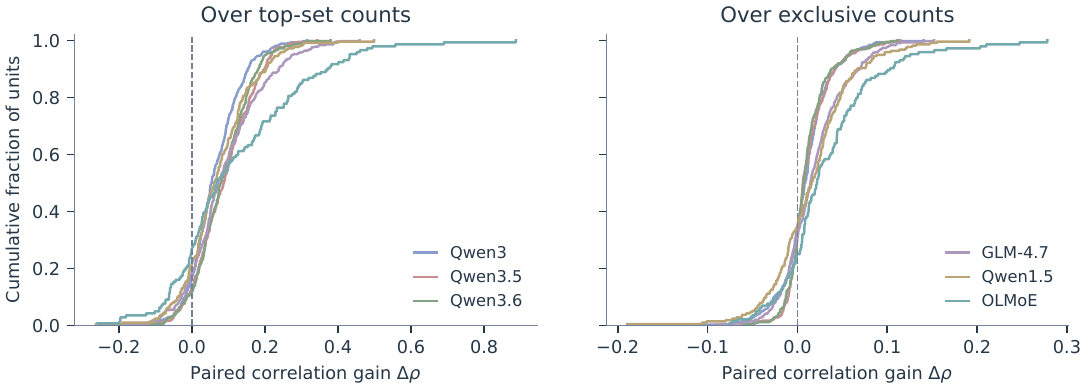}
\caption{Paired Spearman gains over original top-set counts (left) and exclusive counts (right) for validation relative dispatch preference.
Each panel includes all 1926 domain--layer units, using routing-weighted 6.25\% discovery sets and matched finite-$F$ expert support.
ECDFs retain negative gains; no units are selected by their routing outcome.}
\label{fig:app-router-gain-ecdf}
\end{figure}

To compare enrichment with original top-set or exclusive counts, all predictors use the same finite-$F$ expert support within a unit.
Differences are computed within units before summarizing their distribution.
At 6.25\%, model-wise median advantages over top-set counts range from 0.0538 to 0.0874, and those over exclusive counts range from 0.0061 to 0.0218.
Within a domain and layer, dividing a count by its positive set size leaves its Spearman correlation unchanged.
Composition improves the association over top-set counts in 1757 of 1926 units and over exclusive counts in 1343, with the remaining negative differences retained in \cref{fig:app-router-gain-ecdf}.

Holding a native layer--expert fixed instead compares enrichment with validation dispatch share across the nine domains, on common finite-$F$ domain support for all predictors.
The model-wise median correlations range from 0.4000 to 0.7197, with negative correlations in every model.
\Cref{tab:router-link-frozen} gives valid and total expert counts and negative correlations for all models.

\subsubsection{Frozen Enriched Lists}
\begin{figure}[!htbp]
\centering
\includegraphics[width=\textwidth]{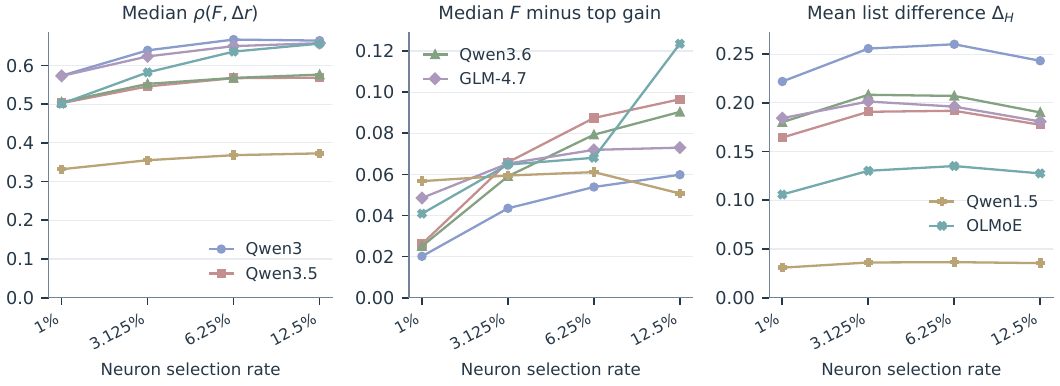}
\caption{Routing-weighted enrichment--routing sensitivity at four selection rates.
Panels show the median $\rho(F,\Delta r)$, median within-unit advantage over top-set counts, and mean frozen-list preference difference.
All six models are shown; the selection rates are discrete measured thresholds, not pruning settings.}
\label{fig:enrichment-routing-rates}
\end{figure}

Aggregating routing over discovery-enriched experts extends the comparison from individual experts to their combined preference for the associated domain.
For each unit, $R_H=\sum_{e\in H_d}r_d^e$, $B_H=\sum_{e\in H_d}b_d^e$ and $\Delta_H=R_H-B_H$ compare the same list on the target and other domains.
Model summaries weight domain--layer units equally, including empty lists with zero shares when present.
All 1926 lists are nonempty at 6.25\%; negative list differences occur in one GLM unit and eight Qwen1.5 units (\cref{tab:router-link-frozen}).
The model means remain positive at all four rates, alongside positive median correlations and paired advantages over top counts (\cref{fig:enrichment-routing-rates}).

\subsubsection{A Matched Scoring Control}
\label{app:weighted-control}
Routing-weighted RAMP includes the routing coefficient, so we check whether the enrichment--routing correspondence depends on it.
At 6.25\%, we compare it with the unweighted score in \cref{eq:ramp-unweighted}, recomputing both correlations on the intersection of their finite-$F$ experts within each domain--layer unit (\cref{tab:router-paired-scoring}).
Both scores retain positive model medians, while the median paired weighted-minus-unweighted change is negative in every model, ranging from $-0.0774$ to $-0.0264$.
The positive enrichment--routing correspondence therefore holds under either score and does not rely on the routing-weight factor, which slightly lowers this particular association; the unweighted control still retains dispatch selection and frequency.

\subsubsection{Whole-Document Resampling}
Whole-document resampling measures variation in validation routing while keeping discovery membership and enriched lists fixed.
The bootstrap draws 64 documents with replacement independently within each domain for 200 repeats.
Document multiplicities are shared across layers and experts, with matching document-slot weights across models and scores.
Each repeat reaccumulates dispatch counts, renormalizes domain shares and recomputes the other-domain mean.
Resampling with replacement duplicates documents and adds noise to dispatch counts, which attenuates Spearman correlations, so 96.5--100\% of replicate median correlations fall below the full-sample value while mean $\Delta_H$ replicates stay centered on it.
Percentile intervals would therefore shift below the estimate, and \cref{tab:router-link-bootstrap} instead reports full-sample estimates with $\pm1.96$ bootstrap standard errors, which remain far from zero in every model.
Every model-level repeat retains all its units.
These intervals describe variation conditional on the fixed discovery structure and current validation pool; no additional correlation $p$ values or BH adjustments are introduced.


\begin{table}[!htbp]
\centering
\caption{Routing-weighted enrichment at four selection rates. $M$ is the mean exclusive-neuron share carried by composition-enriched experts. The last two columns report mean list size and exclusive fraction at 6.25\% only. All entries are percentages averaged equally over domain--layer units; each rate covers all 1926 units with nonempty enriched lists.}
\label{tab:enrichment-rates}
\begingroup\small
\setlength{\tabcolsep}{3pt}
\renewcommand{\arraystretch}{1.12}
\rowcolors{2}{tabrow}{}
\begin{tabularx}{\linewidth}{lRRRRRR}
\toprule
\rowcolor{tabhead}Model & $M_{1\%}$ & $M_{3.125\%}$ & $M_{6.25\%}$ & $M_{12.5\%}$ & $(|H|/E)_{6.25\%}$ & $(|U|/k)_{6.25\%}$ \\
\midrule
Qwen3 & 76.21 & 80.92 & 83.11 & 85.06 & 16.18 & 38.03 \\
Qwen3.5 & 75.52 & 80.63 & 83.02 & 85.68 & 13.61 & 35.04 \\
Qwen3.6 & 76.49 & 80.93 & 83.40 & 85.89 & 13.39 & 34.92 \\
GLM-4.7 & 78.29 & 81.83 & 83.29 & 85.06 & 18.12 & 34.29 \\
Qwen1.5 & 58.18 & 64.76 & 66.40 & 66.90 & 23.26 & 39.51 \\
OLMoE & 67.11 & 72.15 & 72.43 & 72.95 & 17.47 & 27.75 \\
\bottomrule
\end{tabularx}
\endgroup
\end{table}

\begin{table}[!htbp]
\centering
\caption{Domain--layer enrichment--routing associations at routing-weighted 6.25\% selection. $r$ is validation dispatch share and $\Delta r=r-b$ is relative domain preference. Correlations report medians and [25th, 75th percentiles] across units. $N$ is valid coverage, equal to the total; $n_-^r/n_-^{\Delta}$ counts negative correlations. $\widetilde{\Delta\rho}_{F-T}$ is the median within-unit gain over original top counts on matched finite-$F$ support. The last column uses effective routing-weight shares without changing discovery membership.}
\label{tab:router-link-associations}
\begingroup\scriptsize
\setlength{\tabcolsep}{3pt}
\renewcommand{\arraystretch}{1.12}
\rowcolors{2}{tabrow}{}
\begin{tabularx}{\linewidth}{lRRRRRR}
\toprule
\rowcolor{tabhead}Model & $N$ & $\rho(F,r)$ & $\rho(F,\Delta r)$ & $n_-^r/n_-^{\Delta}$ & $\widetilde{\Delta\rho}_{F-T}$ & $\widetilde\rho(F,\Delta r^w)$ \\
\midrule
Qwen3 & 432 & \shortstack{0.7018\\{[0.625, 0.761]}} & \shortstack{0.6668\\{[0.586, 0.721]}} & 0/0 & 0.0538 & 0.6718 \\
Qwen3.5 & 360 & \shortstack{0.5562\\{[0.461, 0.627]}} & \shortstack{0.5671\\{[0.495, 0.628]}} & 0/0 & 0.0874 & 0.5715 \\
Qwen3.6 & 360 & \shortstack{0.5808\\{[0.498, 0.647]}} & \shortstack{0.5677\\{[0.509, 0.630]}} & 0/0 & 0.0793 & 0.5712 \\
GLM-4.7 & 414 & \shortstack{0.6419\\{[0.566, 0.703]}} & \shortstack{0.6498\\{[0.568, 0.713]}} & 0/0 & 0.0719 & 0.6714 \\
Qwen1.5 & 216 & \shortstack{0.3548\\{[0.193, 0.503]}} & \shortstack{0.3682\\{[0.228, 0.499]}} & 13/9 & 0.0611 & 0.6582 \\
OLMoE & 144 & \shortstack{0.5574\\{[0.430, 0.635]}} & \shortstack{0.6357\\{[0.520, 0.706]}} & 3/0 & 0.0681 & 0.6616 \\
\bottomrule
\end{tabularx}
\endgroup
\end{table}

\begin{table}[!htbp]
\centering
\caption{Fixed-expert and frozen-list validation checks at routing-weighted 6.25\% selection. $\rho_{\rm cross}$ correlates $F$ with dispatch share across nine domains within a native layer--expert. $s^e=0$ counts empty exclusive pools. $\bar R_H$, $\bar B_H$ and $\overline\Delta_H$ are domain--layer means for the discovery-frozen list $H=\{e:F>1,\ q<0.05\}$. Shares are fractions. All 1926 lists are nonempty; $n_-^H$ counts negative list differences.}
\label{tab:router-link-frozen}
\begingroup\scriptsize
\setlength{\tabcolsep}{3pt}
\renewcommand{\arraystretch}{1.12}
\rowcolors{2}{tabrow}{}
\begin{tabularx}{\linewidth}{lRRRRRRRR}
\toprule
\rowcolor{tabhead}Model & \shortstack{Cross-domain\\valid/total} & $\widetilde\rho_{\rm cross}$ & $n_-^{\rm cross}$ & $s^e=0$ & $\bar R_H$ & $\bar B_H$ & $\overline\Delta_H$ & $n_-^H$ \\
\midrule
Qwen3 & 6014/6144 & 0.7197 & 125 & 130 & 0.4012 & 0.1410 & 0.2601 & 0 \\
Qwen3.5 & 9612/10240 & 0.6333 & 443 & 628 & 0.3084 & 0.1165 & 0.1919 & 0 \\
Qwen3.6 & 9604/10240 & 0.6390 & 420 & 636 & 0.3214 & 0.1143 & 0.2071 & 0 \\
GLM-4.7 & 2893/2944 & 0.7030 & 93 & 51 & 0.3582 & 0.1621 & 0.1961 & 1 \\
Qwen1.5 & 1439/1440 & 0.4000 & 237 & 1 & 0.2653 & 0.2289 & 0.0365 & 8 \\
OLMoE & 1015/1024 & 0.6946 & 38 & 9 & 0.3007 & 0.1655 & 0.1351 & 0 \\
\bottomrule
\end{tabularx}
\endgroup
\end{table}

\begin{table}[!htbp]
\centering
\begin{minipage}[t]{0.48\linewidth}
\vspace{0pt}
\caption{Paired scoring control at 6.25\% selection with unweighted ($U$) and routing-weighted ($W$) scores. Paired $\Delta$ is the median unit-wise $W-U$ difference, negative in $n_-$ of $N$ units.}
\label{tab:router-paired-scoring}
\centering\footnotesize
\setlength{\tabcolsep}{2.5pt}
\rowcolors{2}{tabrow}{}
\begin{tabularx}{\linewidth}{lRRRRR}
\toprule
\rowcolor{tabhead}Model & $N$ & $\widetilde\rho_U$ & $\widetilde\rho_W$ & Paired $\Delta$ & $n_-$ \\
\midrule
Qwen3 & 432 & 0.6918 & 0.6716 & -0.0264 & 312 \\
Qwen3.5 & 360 & 0.6008 & 0.5732 & -0.0275 & 298 \\
Qwen3.6 & 360 & 0.6064 & 0.5747 & -0.0302 & 289 \\
GLM-4.7 & 414 & 0.7108 & 0.6503 & -0.0531 & 359 \\
Qwen1.5 & 216 & 0.4196 & 0.3682 & -0.0774 & 173 \\
OLMoE & 144 & 0.6840 & 0.6368 & -0.0620 & 118 \\
\bottomrule
\end{tabularx}
\end{minipage}\hfill
\begin{minipage}[t]{0.49\linewidth}
\vspace{0pt}
\caption{Document-bootstrap intervals at 6.25\% selection. Entries are full-sample estimates $\pm1.96$ standard errors from 200 replicates, conditional on the fixed discovery structure and the validation pool.}
\label{tab:router-link-bootstrap}
\centering\footnotesize
\setlength{\tabcolsep}{2.5pt}
\rowcolors{2}{tabrow}{}
\begin{tabularx}{\linewidth}{lRR}
\toprule
\rowcolor{tabhead}Model & Median $\rho(F,\Delta r)$ & Mean $\Delta_H$ \\
\midrule
Qwen3 & 0.6668 [0.6548, 0.6788] & 0.2601 [0.2481, 0.2722] \\
Qwen3.5 & 0.5671 [0.5541, 0.5802] & 0.1919 [0.1825, 0.2012] \\
Qwen3.6 & 0.5677 [0.5554, 0.5801] & 0.2071 [0.1971, 0.2172] \\
GLM-4.7 & 0.6498 [0.6377, 0.6620] & 0.1961 [0.1867, 0.2056] \\
Qwen1.5 & 0.3682 [0.3390, 0.3973] & 0.0365 [0.0341, 0.0389] \\
OLMoE & 0.6357 [0.6112, 0.6603] & 0.1351 [0.1279, 0.1424] \\
\bottomrule
\end{tabularx}
\end{minipage}
\end{table}

\subsection{Ablating Domain Neuron Sets}
\label{app:domain-ablation}
\begin{table}[!htbp]
\caption{Task scores (\%) after zeroing the top-10\% neuron set $T_d^{(10\%)}$ of one domain, compared with the original model and with RAMPr at 90\% retention, which removes the same number of neurons. Each block lists the tasks matched to one domain; gen and mcq denote the generation and multiple-choice buckets of that domain. Shaded rows ablate the domain matched to the task.}
\label{tab:domain-ablation}
\centering\scriptsize
\setlength{\tabcolsep}{4pt}
\begin{tabularx}{\textwidth}{llRRRRRR}
\toprule
\rowcolor{tabhead}Task & Setting & Qwen3 & Qwen3.5 & Qwen3.6 & GLM-4.7 & Qwen1.5 & OLMoE \\
\midrule
WinoGrande & Original & 70.32 & 75.06 & 74.66 & 73.16 & 65.59 & 67.72 \\
 & RAMPr 90\% & 71.59 & 74.19 & 74.43 & 73.01 & 65.98 & 67.40 \\
\rowcolor{tabrow} & Ablate chat-gen & 51.38 & 72.06 & 71.43 & 62.04 & 55.64 & 51.70 \\
\rowcolor{tabrow} & Ablate chat-mcq & 49.17 & 63.93 & 64.96 & 52.64 & 51.78 & 52.80 \\
\midrule
HellaSwag & Original & 77.59 & 82.64 & 83.52 & 80.13 & 78.90 & 80.75 \\
 & RAMPr 90\% & 77.40 & 82.35 & 83.51 & 80.34 & 78.68 & 80.17 \\
\rowcolor{tabrow} & Ablate chat-gen & 34.35 & 77.01 & 77.80 & 62.43 & 47.00 & 44.84 \\
\rowcolor{tabrow} & Ablate chat-mcq & 37.89 & 69.45 & 70.68 & 60.40 & 45.30 & 48.12 \\
\midrule
HumanEval & Original & 87.20 & 98.17 & 96.34 & 91.46 & 43.90 & 37.20 \\
 & RAMPr 90\% & 89.63 & 95.12 & 97.56 & 88.41 & 44.51 & 34.15 \\
\rowcolor{tabrow} & Ablate code-gen & 0.00 & 0.00 & 0.00 & 0.00 & 0.00 & 0.00 \\
\midrule
MBPP & Original & 88.00 & 86.60 & 91.80 & 80.60 & 33.20 & 28.80 \\
 & RAMPr 90\% & 87.20 & 89.60 & 91.40 & 82.20 & 31.80 & 28.80 \\
\rowcolor{tabrow} & Ablate code-gen & 0.00 & 0.00 & 0.40 & 0.00 & 0.00 & 0.00 \\
\midrule
GSM8K & Original & 76.42 & 97.12 & 96.66 & 92.19 & 19.94 & 68.23 \\
 & RAMPr 90\% & 76.42 & 96.97 & 96.66 & 93.25 & 21.46 & 69.52 \\
\rowcolor{tabrow} & Ablate math-gen & 0.00 & 0.99 & 0.91 & 0.08 & 0.00 & 0.00 \\
\rowcolor{tabrow} & Ablate math-mcq & 0.00 & 0.61 & 1.06 & 0.15 & 0.00 & 0.00 \\
\midrule
MATH-500 & Original & 96.00 & 96.80 & 97.40 & 88.40 & 22.60 & 26.00 \\
 & RAMPr 90\% & 95.80 & 97.60 & 97.40 & 88.20 & 24.00 & 23.00 \\
\rowcolor{tabrow} & Ablate math-gen & 0.40 & 2.60 & 2.20 & 1.20 & 0.00 & 0.00 \\
\rowcolor{tabrow} & Ablate math-mcq & 0.00 & 7.60 & 7.20 & 2.00 & 1.00 & 0.00 \\
\midrule
PubMedQA & Original & 69.20 & 65.40 & 76.40 & 53.40 & 14.60 & 60.60 \\
 & RAMPr 90\% & 68.60 & 66.00 & 76.60 & 54.00 & 15.80 & 60.80 \\
\rowcolor{tabrow} & Ablate med-gen & 36.00 & 52.80 & 63.40 & 48.60 & 35.40 & 49.40 \\
\rowcolor{tabrow} & Ablate med-mcq & 41.80 & 52.20 & 38.40 & 42.80 & 46.40 & 44.20 \\
\midrule
CEval & Original & 83.43 & 84.40 & 83.58 & 69.24 & 76.30 & 37.44 \\
 & RAMPr 90\% & 82.99 & 83.73 & 83.80 & 69.32 & 74.00 & 36.63 \\
\rowcolor{tabrow} & Ablate zh-gen & 29.72 & 76.82 & 76.37 & 46.29 & 28.01 & 24.07 \\
\rowcolor{tabrow} & Ablate zh-mcq & 26.75 & 37.37 & 39.97 & 27.41 & 26.82 & 24.22 \\
\midrule
CMMLU & Original & 83.77 & 85.21 & 85.19 & 72.24 & 77.47 & 36.40 \\
 & RAMPr 90\% & 83.34 & 84.82 & 85.11 & 71.48 & 75.40 & 36.46 \\
\rowcolor{tabrow} & Ablate zh-gen & 30.33 & 76.88 & 76.14 & 49.36 & 26.22 & 25.57 \\
\rowcolor{tabrow} & Ablate zh-mcq & 25.48 & 43.79 & 46.61 & 30.96 & 25.62 & 25.09 \\
\bottomrule
\end{tabularx}
\end{table}

The overlap, spatial and enrichment results in \cref{sec:neuron-organization} are correlational.
They show which neurons RAMP ranks highest for each domain, but not whether the model needs these neurons to solve that domain's tasks.
This subsection asks that question directly: if a domain's top-ranked neurons are removed, does the model lose the matching capability?
For each model and each of the nine buckets, we select the top 10\% of routed neurons in every layer by routing-weighted RAMP (\cref{eq:ramp-weighted}) on the discovery split, denoted $T_d^{(10\%)}$ to distinguish it from the 6.25\% sets of the overlap analyses.
We zero the gate and up rows and the down columns of these neurons, as in the zero-masked pruning of \cref{sec:pruning}.
Each domain is ablated separately, and neurons shared with other domains remain in its set.

As a reference of the same size, RAMPr at 90\% retention removes the 10\% of routed neurons with the lowest multi-domain importance, and Appendix~\ref{app:random-removal} compares both with random removal.
Each ablated model is evaluated on the tasks matched to its domain under the protocols of the main table (Appendix~\ref{app:evaluation-settings}), and \cref{tab:domain-ablation} reports all scores.

\textbf{Answer.} Yes. Removing a domain's top-ranked neurons causes large losses on the matched tasks, whereas removing the same number of low-importance neurons with RAMPr causes small ones.
The neurons that RAMP ranks highest for a domain are therefore needed for that domain's tasks, not only correlated with them.
Ablating the code set reduces HumanEval and MBPP to at most 0.40\% in all six models, and ablating either mathematics set reduces GSM8K to at most 1.06\% and MATH-500 to at most 7.60\%, whereas RAMPr at 90\% retention stays within five points of the original model on these tasks.

The Chinese multiple-choice set reduces CEval and CMMLU to 24.22--46.61\% across the six models.
In Qwen3.5, Qwen3.6 and GLM-4.7, ablating the Chinese generation set causes smaller losses than ablating the multiple-choice set, leaving 46.29--76.88\% against 27.41--46.61\%, and ablating the chat generation set lowers WinoGrande and HellaSwag by at most 5.72 points in Qwen3.5 and Qwen3.6, consistent with the partial overlap between formats in Appendix~\ref{app:cross-format}.
In Qwen3, Qwen1.5 and OLMoE, both Chinese ablations bring CEval and CMMLU to 24.07--30.33\%, close to four-choice chance.
PubMedQA decreases after either medical ablation in five models, by 4.80 to 38.00 points.

Qwen1.5 is the exception on PubMedQA, where accuracy rises from 14.60\% to 35.40\% after the medical generation ablation and to 46.40\% after the multiple-choice ablation.
The original model answers ``maybe'' to 472 of the 500 questions, scoring below the 55.20\% obtained by always answering ``yes''.
After the generation ablation it answers ``no'' to 481 questions, after the multiple-choice ablation it answers mostly ``yes'' or ``no'', and in both cases none of the 55 ``maybe'' questions is answered correctly.
The higher accuracy therefore reflects a shift in label preference rather than better medical answers, and we exclude this entry from the comparison above.

\subsection{Random Removal of the Same Size}
\label{app:random-removal}
\begin{table}[t]
\caption{Score difference (percentage points) between random removal and RAMPr at the same budget: random 10\% against RAMPr at 90\% retention, and random 25\% against RAMPr at 75\%. Negative values mean random removal scores lower.}
\label{tab:random-removal}
\centering\scriptsize
\setlength{\tabcolsep}{3pt}
\rowcolors{4}{}{tabrow}
\begin{tabularx}{\textwidth}{lRRRRRRRRRRRR}
\toprule
\rowcolor{tabhead} & \multicolumn{6}{c}{10\% removed} & \multicolumn{6}{c}{25\% removed} \\
\cmidrule(lr){2-7}\cmidrule(lr){8-13}
\rowcolor{tabhead}Task & Qwen3 & Qwen3.5 & Qwen3.6 & GLM-4.7 & Qwen1.5 & OLMoE & Qwen3 & Qwen3.5 & Qwen3.6 & GLM-4.7 & Qwen1.5 & OLMoE \\
\midrule
WinoGrande & $-$2.37 & $-$1.72 & +0.11 & $-$1.03 & $-$0.87 & $-$2.21 & $-$13.42 & $-$1.74 & $-$2.29 & $-$12.08 & +0.47 & $-$1.66 \\
HellaSwag & $-$2.42 & $-$1.59 & $-$1.64 & $-$2.52 & $-$0.90 & $-$2.42 & $-$9.03 & $-$5.19 & $-$4.60 & $-$16.56 & $-$1.82 & $-$5.34 \\
CEval & $-$2.08 & $-$0.89 & $-$0.59 & $-$4.83 & $-$3.05 & $-$2.01 & $-$8.54 & $-$4.16 & $-$4.68 & $-$11.74 & $-$7.13 & $-$4.90 \\
CMMLU & $-$2.35 & $-$1.20 & $-$1.21 & $-$3.51 & $-$2.23 & $-$1.46 & $-$9.20 & $-$4.65 & $-$3.89 & $-$9.10 & $-$5.33 & $-$4.55 \\
PubMedQA & $-$5.80 & $-$5.80 & $-$0.20 & $-$1.60 & $-$1.20 & $-$0.60 & $-$7.20 & $-$0.40 & $-$1.40 & $-$13.40 & +0.80 & $-$27.40 \\
HumanEval & $-$5.49 & $-$1.83 & $-$1.83 & $-$2.78 & $-$4.88 & $-$12.80 & $-$83.54 & $-$10.98 & $-$9.15 & $-$22.56 & $-$23.17 & $-$28.66 \\
MBPP & $-$2.60 & $-$4.60 & $-$3.80 & $-$9.40 & $-$2.80 & $-$7.40 & $-$80.60 & $-$13.00 & $-$8.60 & $-$30.60 & $-$10.40 & $-$17.80 \\
GSM8K & $-$0.76 & $-$0.38 & $-$1.29 & $-$4.78 & $-$2.20 & $-$15.16 & $-$70.28 & $-$2.20 & $-$5.69 & $-$28.43 & $-$27.60 & $-$42.68 \\
MATH-500 & $-$2.40 & $-$1.80 & $-$1.60 & $-$6.00 & $-$8.40 & $-$7.60 & $-$95.00 & $-$6.00 & $-$8.80 & $-$32.40 & $-$21.60 & $-$20.60 \\
\midrule
Mean & $-$2.92 & $-$2.20 & $-$1.34 & $-$4.05 & $-$2.95 & $-$5.74 & $-$41.87 & $-$5.37 & $-$5.46 & $-$19.65 & $-$10.64 & $-$17.07 \\
\bottomrule
\end{tabularx}
\end{table}

The losses in Appendix~\ref{app:domain-ablation} could arise simply because 10\% of the routed neurons are removed, regardless of which neurons they are.
If so, removing the same number of neurons at random should cause losses of similar size, and the RAMP ranking would carry little information.
We therefore remove 10\% and 25\% of the routed neurons uniformly at random in every layer, zero them as above, and evaluate each model on the nine tasks of \cref{tab:domain-ablation}.

\textbf{Answer.} No. Random removal of 10\% lowers task scores by at most 15.86 points relative to the original model, far less than ablating a domain's top 10\%, which drives code and mathematics scores to near zero.
Across the 96 matched pairs of domain ablation and task, the targeted ablation scores below random removal in 94, and the two exceptions are the Qwen1.5 PubMedQA entries explained above.
The large losses in \cref{tab:domain-ablation} therefore come from which neurons are removed, not from how many.

The same comparison also shows that the ranking is informative at the low-importance end (\cref{tab:random-removal}).
At 10\% removal, random neurons score below RAMPr in 53 of the 54 model--task pairs, with model means from $-1.34$ to $-5.74$ points, and the largest gaps fall on generation tasks, down to $-15.16$ points on OLMoE GSM8K.
At 25\% removal the gap widens to model means of $-5.37$ to $-41.87$ points, and random removal scores below RAMPr in 52 of the 54 pairs.
Random removal reduces Qwen3's HumanEval, MBPP and MATH-500 to at most 0.61\%, whereas RAMPr keeps 80.60--95.00\%.
The three pairs where random removal is not lower involve tasks that barely respond to any removal in that model, so their differences of at most 0.80 points do not separate the two methods.
On Qwen3.6 WinoGrande ($+0.11$ at 10\%), the original model, random removal and RAMPr score 74.66\%, 74.54\% and 74.43\%, and even RAMPr at 50\% retention loses at most 2.99 points.
On Qwen1.5 WinoGrande ($+0.47$ at 25\%), RAMPr at 75\% retention matches the original 65.59\% exactly, and RAMPr at 50\% loses at most 2.13 points.
On Qwen1.5 PubMedQA ($+0.80$ at 25\%), every condition scores between 11.40\% and 15.80\%, below the 55.20\% obtained by always answering ``yes'', because of the label preference described in Appendix~\ref{app:domain-ablation}, so this task does not measure the model's medical ability.
RAMP thus identifies both the neurons a domain depends on and the neurons whose removal costs least.

\section{Additional Training-Dynamics Results}
\label{app:evolution-protocol}

Tracing how neuron-level specialization develops requires following neuron importance through training, rather than inferring its stability from expert aggregates.

\subsection{Adjacent-Checkpoint Comparisons Across Training Stages}
\label{app:evolution-stage-level}
A stage comparison needs the same unit of change in both groups: an adjacent checkpoint pair.
We compute neuron- and expert-level correlations within pre-training step pairs and within post-training release pairs, then compare the groups at each level.
The primary analysis neither compares the first and last pre-training checkpoints nor averages checkpoint importance vectors before computing a correlation.
Pre-training includes 33 adjacent pairs from 34 OLMoE checkpoints and six from seven OpenMoE checkpoints.
Post-training includes Base--SFT, SFT--DPO and DPO--Instruct for OLMoE, and Base--SFT, SFT--DPO and DPO--Think for Instella.
Across eleven domains and all native MoE layers, these 45 pairs yield $33\times11\times16=5808$, $6\times11\times4=264$, $3\times11\times16=528$ and $3\times11\times26=858$ cells, totaling 7,458.
OLMoE stage2 and Instella Pretrain--Midtrain--Base belong to the additional interval analyses below.

The main comparison uses \texttt{validation/all}: 64 validation documents per domain, without transition filtering.
Discovery and validation each have all/A/B views (64/32/32 documents), totaling 44,748 records; views are not pooled.
Token contributions accumulate within each domain, layer and checkpoint before rank comparison, with domain--layer--pair cells weighted equally in the summaries.

\paragraph{Neuron importance and expert aggregation.}
For checkpoint $t$, $I_{li}^e(D_d;t)$ is routing-weighted RAMP (\cref{eq:ramp-weighted}), using native executed dispatch edges.
The neuron-level ranking and its expert aggregate can then be compared within cell $c=(d,l,a,b)$:
\begin{equation}
\begin{aligned}
 J^{(t)}_{dle}&=\sum_i I_{li}^e(D_d;t),\\
 \rho_{E,c}&=\operatorname{Spearman}\!\left(J^{(a)}_{dl,:},J^{(b)}_{dl,:}\right),\\
 \rho_{N,c}&=\operatorname{Spearman}\!\left(\operatorname{vec}\!\left(\bigl[I_{li}^e(D_d;a)\bigr]_{e,i}\right),\operatorname{vec}\!\left(\bigl[I_{li}^e(D_d;b)\bigr]_{e,i}\right)\right),\\
 \Delta\rho_c&=\rho_{E,c}-\rho_{N,c}.
\end{aligned}
\end{equation}
Here $\operatorname{vec}$ flattens the expert--neuron matrix in the same coordinate order at both checkpoints.
Expert sums (denoted $S_e$ below) remain within domains and layers; neuron ranks span the full layer, not individual experts.
Ties receive average ranks; constant or insufficient vectors give NA.
All 7,458 cells have valid correlations at both levels.

\begin{table}[!htbp]
\caption{Unfiltered validation/all cell medians and positive-gap counts.
Paired-gap medians summarize within-cell differences, not differences between marginal medians.}
\label{tab:evolution-stage-level}
\centering\small
\rowcolors{2}{tabrow}{}
\begin{tabularx}{\linewidth}{lLRRRRR}
\toprule
\rowcolor{tabhead}Stage & Model & Cells & Median $\rho_E$ & Median $\rho_N$ & Median $\Delta\rho$ & $\Delta\rho>0$ \\
\midrule
Pre & OLMoE & 5808 & 0.989789 & 0.939291 & 0.048084 & 5775/5808 \\
Pre & OpenMoE & 264 & 0.916415 & 0.841811 & 0.053208 & 201/264 \\
Post & OLMoE & 528 & 0.998077 & 0.996918 & 0.000715 & 365/528 \\
Post & Instella & 858 & 0.998947 & 0.997497 & 0.001597 & 737/858 \\
\bottomrule
\end{tabularx}
\end{table}

\Cref{tab:evolution-stage-level} reports the fraction with strictly positive $\Delta\rho$; reverse counts are 33, 63, 163 and 121, with no zeros.
The larger pre-training gaps indicate neuron-level reorganization that is less apparent in expert aggregates, whereas post-training largely preserves both rankings.
\Cref{tab:evolution-stage-level-distributions} shows the spread behind these medians, while \cref{tab:evolution-post-stage-medians} separates Base--SFT adjustment from the greater stability of later pairs.

\begin{table}[!htbp]
\centering
\begin{minipage}[t]{0.51\linewidth}
\vspace{0pt}
\setlength{\tabcolsep}{2pt}
\caption{Same-cell distribution summaries for all four groups.
Means and medians weight cells equally; p10/p90 are linearly interpolated percentiles.
Each $\Delta\rho$ is computed before summarization.}
\label{tab:evolution-stage-level-distributions}
\centering\small
\rowcolors{2}{tabrow}{}
\begin{tabularx}{\linewidth}{lLRRRR}
\toprule
\rowcolor{tabhead}Group & Metric & Mean & Median & p10 & p90 \\
\midrule
Pre OLMoE & $\rho_E$ & 0.984780 & 0.989789 & 0.969780 & 0.995742 \\
 & $\rho_N$ & 0.933405 & 0.939291 & 0.889235 & 0.970663 \\
 & $\Delta\rho$ & 0.051376 & 0.048084 & 0.021692 & 0.085220 \\
\midrule
Pre OpenMoE & $\rho_E$ & 0.841709 & 0.916415 & 0.594992 & 0.985528 \\
 & $\rho_N$ & 0.810303 & 0.841811 & 0.670846 & 0.935030 \\
 & $\Delta\rho$ & 0.031405 & 0.053208 & -0.123104 & 0.134631 \\
\midrule
Post OLMoE & $\rho_E$ & 0.994372 & 0.998077 & 0.985458 & 0.999634 \\
 & $\rho_N$ & 0.988550 & 0.996918 & 0.967078 & 0.999691 \\
 & $\Delta\rho$ & 0.005822 & 0.000715 & -0.000486 & 0.020462 \\
\midrule
Post Instella & $\rho_E$ & 0.968869 & 0.998947 & 0.893915 & 0.999725 \\
 & $\rho_N$ & 0.948620 & 0.997497 & 0.828626 & 0.999693 \\
 & $\Delta\rho$ & 0.020249 & 0.001597 & -0.000074 & 0.067943 \\
\bottomrule
\end{tabularx}
\end{minipage}\hfill
\begin{minipage}[t]{0.46\linewidth}
\vspace{0pt}
\setlength{\tabcolsep}{2pt}
\caption{Post-training cell medians used in \cref{fig:evolution-stage-level}d, distinct from interval means below.
Each pair contains 176 OLMoE or 286 Instella domain--layer cells.}
\label{tab:evolution-post-stage-medians}
\centering\small
\rowcolors{2}{tabrow}{}
\begin{tabularx}{\linewidth}{lLRRR}
\toprule
\rowcolor{tabhead}Model & Stage pair & \shortstack{Median\\$\rho_E$} & \shortstack{Median\\$\rho_N$} & \shortstack{Median\\$\Delta\rho$} \\
\midrule
OLMoE & Base--SFT & 0.988553 & 0.971534 & 0.015946 \\
OLMoE & SFT--DPO & 0.998329 & 0.997987 & 0.000212 \\
OLMoE & DPO--Instruct & 0.999313 & 0.999351 & 0.000012 \\
Instella & Base--SFT & 0.914858 & 0.853142 & 0.054966 \\
Instella & SFT--DPO & 0.999405 & 0.998732 & 0.000517 \\
Instella & DPO--Think & 0.999313 & 0.998312 & 0.000785 \\
\bottomrule
\end{tabularx}
\end{minipage}
\end{table}

Checkpoint spacing differs; OLMoE uses 0924 pre-training and 0125 post-training series, while OpenMoE and Instella are complementary families, not a matched trajectory.

\subsection{Checkpoint Coverage and Evaluation Settings}
\label{app:evolution-coverage}
The fixed-text design in \cref{sec:evolution-design} lets us follow corresponding neuron coordinates across checkpoints within each model family.
Coverage extends beyond the primary 45-pair comparison to the additional interval analyses, with shared execution settings, routed populations and document sampling.

The two parallel studies cover 56 checkpoints, comprising 43 OLMoE, seven OpenMoE and six Instella checkpoints.
OLMoE's 0924 pre-training series has 34 points between step 5000 and step 1200000.
Its five 0125 stage2 checkpoints use a separate step axis between 5000 and 23842.
The four 0125 releases are Base, SFT, DPO and Instruct.
The final stage2 checkpoint and released Base model are separate observations.
OpenMoE checkpoints carry token labels 200B, 400B, 600B, 800B, 890B, 1T and 1.1T.
Instella releases are Pretrain, Midtrain, Base, SFT, DPO and Think.
Each family retains its own checkpoint and stage axes, with OLMoE pre-training, stage2 and released-stage series reported separately.
Historical OpenMoE Chat summaries are outside this 56-checkpoint routed analysis.

Routed layers contain 64 experts of width 1024 in OLMoE, 32 of width 8192 in OpenMoE, and 64 of width 1408 in Instella.
Their respective numbers of MoE layers are 16, 4 and 26, with native top-$K$ values of 8, 2 and 6.
OLMoE covers layers 0--15.
OpenMoE covers native decoder layers 5, 11, 17 and 23, preserving capacity-limited dispatch and counting only post-capacity assignments.

Instella covers layers 1--26, excluding its first dense layer from routed-neuron statistics.
Its configuration's two shared experts are implemented as one shared MLP of width 2816.
This branch remains active and is analyzed separately from routed-expert statistics and decomposition.
Instella retains native Gated MLA and FarSkip behavior, and its model card specifies ResearchRAIL.
The model descriptions follow OpenMoE \citep{pmlr-v235-xue24c} and the Instella-MoE technical report \citep{liu2026instellamoe}.

The eleven text buckets are the nine discovery buckets plus finance-mcq and law-mcq.
Each contains 128 records shuffled with seed 1234.
The first 64 form discovery and the remaining 64 form the validation set, giving 704 documents per side.
Each side is divided in order into two fixed 32-document halves per bucket (32/32).
All checkpoints reuse these document sets.
Discovery and validation text are disjoint under exact-text and whitespace-normalized matching.
Raw text is right-truncated to 2048 tokens without added special tokens or a chat template.
Statistics use non-padding input tokens.

Forward passes predominantly use BF16 with FP32 statistics.
OpenMoE retains its specialized ColossalAI runtime and capacity rules.
Instella preserves native FP32 routing buffers and casts its originally FP32 Midtrain weights to BF16 for forward evaluation.
OLMoE stage2 uses the published 0125 Base tokenizer through an explicit loading configuration that resolves the vocabulary mismatch without changing model weights.
Input token sequences are identical across checkpoints within each family, and raw texts are shared across families.

\subsection{Neuron and Expert Statistics over Additional Intervals}
\label{app:evolution-intervals}
Long pre-training intervals, the separate OLMoE stage2 run and the full release series allow us to examine neuron reorganization beyond the primary adjacent-pair comparison.
These summaries use domain--layer means for each specified interval, not the unfiltered adjacent-pair medians in \cref{tab:evolution-stage-level,tab:evolution-post-stage-medians}.
Comparing neuron rankings with expert utilization and response magnitudes helps distinguish changes at these different levels.

Routed-neuron importance follows \cref{eq:ramp-weighted} with checkpoint-specific weights and activations.
Squared routing-weighted activations are accumulated across tokens and documents before taking the square root and multiplying by the corresponding down-projection column norm.
For fixed domain $d$, layer and checkpoint, summing neuron importance over an expert yields $S_e=\sum_i I_i^e(D_d)$.
Complete-output energy is
\begin{equation}
 E_e=\sum_{x,t}\left\|p_e(t)f_e(x_l(t))\right\|_2^2.
\end{equation}
It includes interactions among neuron outputs and has no outer square root, distinguishing it from both $S_e$ and $\sum_i \bigl[I_i^e(D_d)\bigr]^2$.
Dispatch count $C_e$ counts only executed edges.

For these interval analyses, correlations compare corresponding coordinates within each domain and layer, then average equally over valid cells.
There are 176 domain--layer cells for OLMoE, 44 for OpenMoE and 286 for Instella.
Spearman ties use average ranks, and constant vectors remain NA.
The routing-load shift between checkpoints $a$ and $b$ is $\frac12\sum_e|C_e^a/\sum_{e'}C_{e'}^a-C_e^b/\sum_{e'}C_{e'}^b|$, the fraction of dispatches that would have to move between experts to turn one load distribution into the other, while count-rank correlation measures the ordering of expert loads.
Top-set diagnostics use Jaccard overlap, $|A\cap B|/|A\cup B|$, which differs from the intersection-over-set-size statistic in the discovery section.
Expert top sets are selected by $S_e$, and neuron top sets by importance over all expert--neuron coordinates within a layer.

\begin{figure*}[t]
\centering
\includegraphics[width=\textwidth]{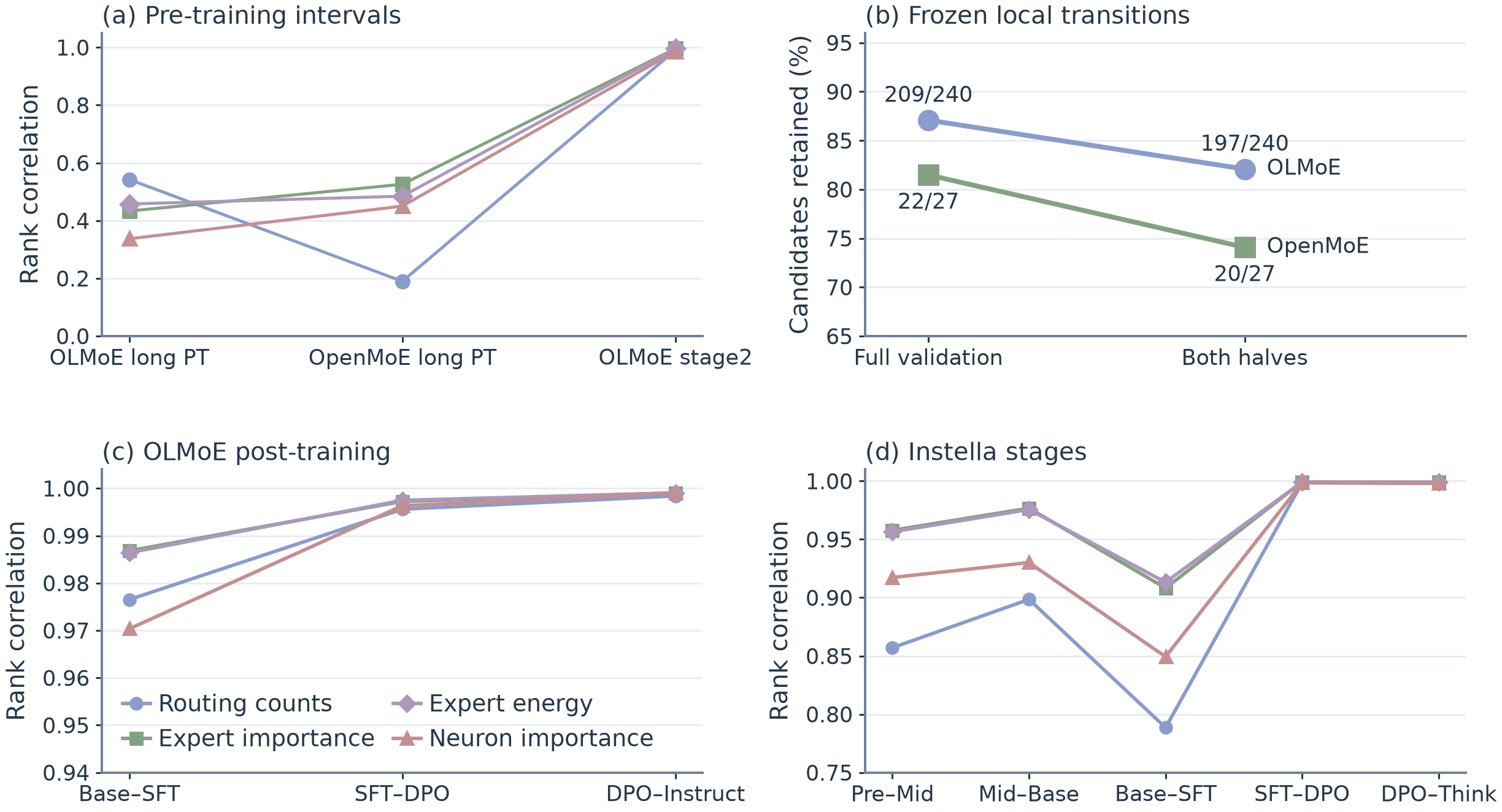}
\caption{Additional multi-granularity interval changes and local replication.
Panels (a), (c) and (d) compare validation rank correlations for routing counts, summed neuron importance, full expert-output energy and neuron importance, using domain--layer means for each interval.
Pre-training endpoint comparisons in (a) are separate intervals.
Panels (c) and (d) show within-family stage comparisons.
Panel (b) reports the original candidate counts retained by the full validation set and both halves.
Connecting lines follow categorical comparisons and do not encode interval duration.
Each panel labels its own vertical scale.}
\label{fig:evolution-main}
\end{figure*}

\begin{table}[!htbp]
\centering
\begin{minipage}[t]{0.46\linewidth}
\vspace{0pt}
\caption{Validation expert-rank correlations over long pre-training intervals and OLMoE stage2.
Columns correspond to dispatch counts, summed neuron importance and complete-output energy.
Each entry is a domain--layer mean.}
\label{tab:expert-endpoints}
\centering\small
\rowcolors{2}{tabrow}{}
\begin{tabularx}{\linewidth}{lRRR}
\toprule
\rowcolor{tabhead}Interval & $C$ & $S$ & $E$ \\
\midrule
OLMoE 20B--5033B & .543 & .433 & .458 \\
OpenMoE 200B--1.1T & .190 & .526 & .485 \\
OLMoE stage2 21B--101B & .991 & .996 & .996 \\
\bottomrule
\end{tabularx}
\end{minipage}\hfill
\begin{minipage}[t]{0.51\linewidth}
\vspace{0pt}
Long pre-training intervals show substantial changes in expert rankings (\cref{tab:expert-endpoints,fig:evolution-main}).
OpenMoE's validation dispatch-count correlation is 0.190 over the full interval, below those for summed importance (0.526) and output energy (0.485).
Rankings are much more stable in OLMoE's separate stage2 interval.
During Instella Base-to-SFT, the validation count correlation is 0.788, compared with 0.908 for summed importance and 0.913 for output energy (\cref{fig:expert-stages}).
The relative stability of routing and expert responses therefore depends on both the statistic and the training interval; routing loads alone do not characterize how expert contributions evolve.
\end{minipage}
\end{table}

\begin{figure}[!htbp]
\centering
\includegraphics[width=\textwidth]{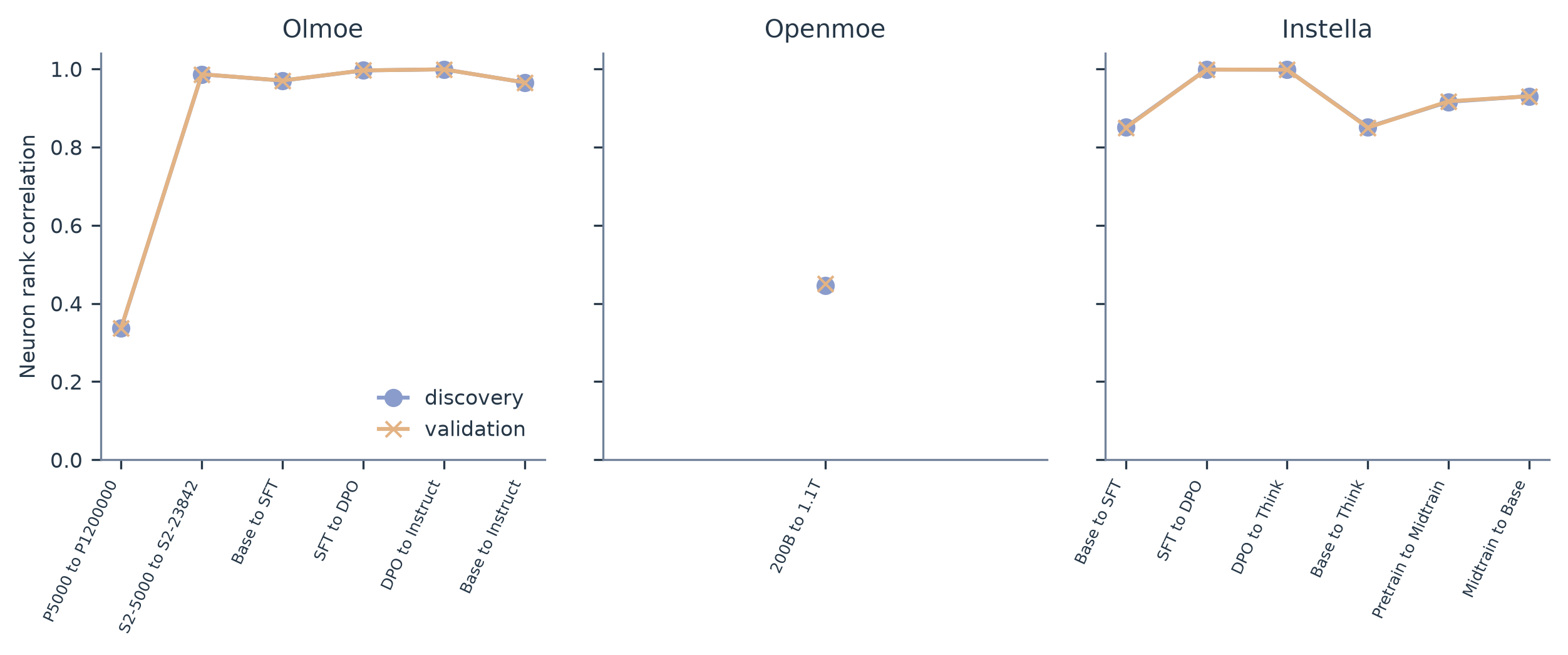}
\caption{Endpoint and released-stage neuron comparisons, including Base-to-final comparisons and Instella's earlier stages.
Points are domain--layer mean Spearman correlations.
Lines join categorical comparisons only.
Discovery and validation markers can overlap when their values are close.}
\label{fig:neuron-stages}
\end{figure}

Neuron rankings also change substantially over long pre-training intervals (\cref{fig:neuron-stages}).
OLMoE's endpoint mean correlation is 0.336 on discovery and 0.337 on validation text; OpenMoE yields 0.446 and 0.450.
OLMoE's separate stage2 run has a mean correlation of 0.986 on both splits.
For post-training, validation Base-to-SFT mean correlations are 0.970 for OLMoE and 0.849 for Instella, increasing to 0.9963 and 0.9985 from SFT to DPO.
OLMoE DPO-to-Instruct yields 0.9990, while Instella DPO-to-Think yields 0.9981.
Instella's earlier Pretrain-to-Midtrain and Midtrain-to-Base means are 0.917 and 0.930.
Base-to-SFT changes neuron rankings more than the subsequent evaluated pairs in both release series.

\begin{figure}[!htbp]
\centering
\includegraphics[width=\textwidth]{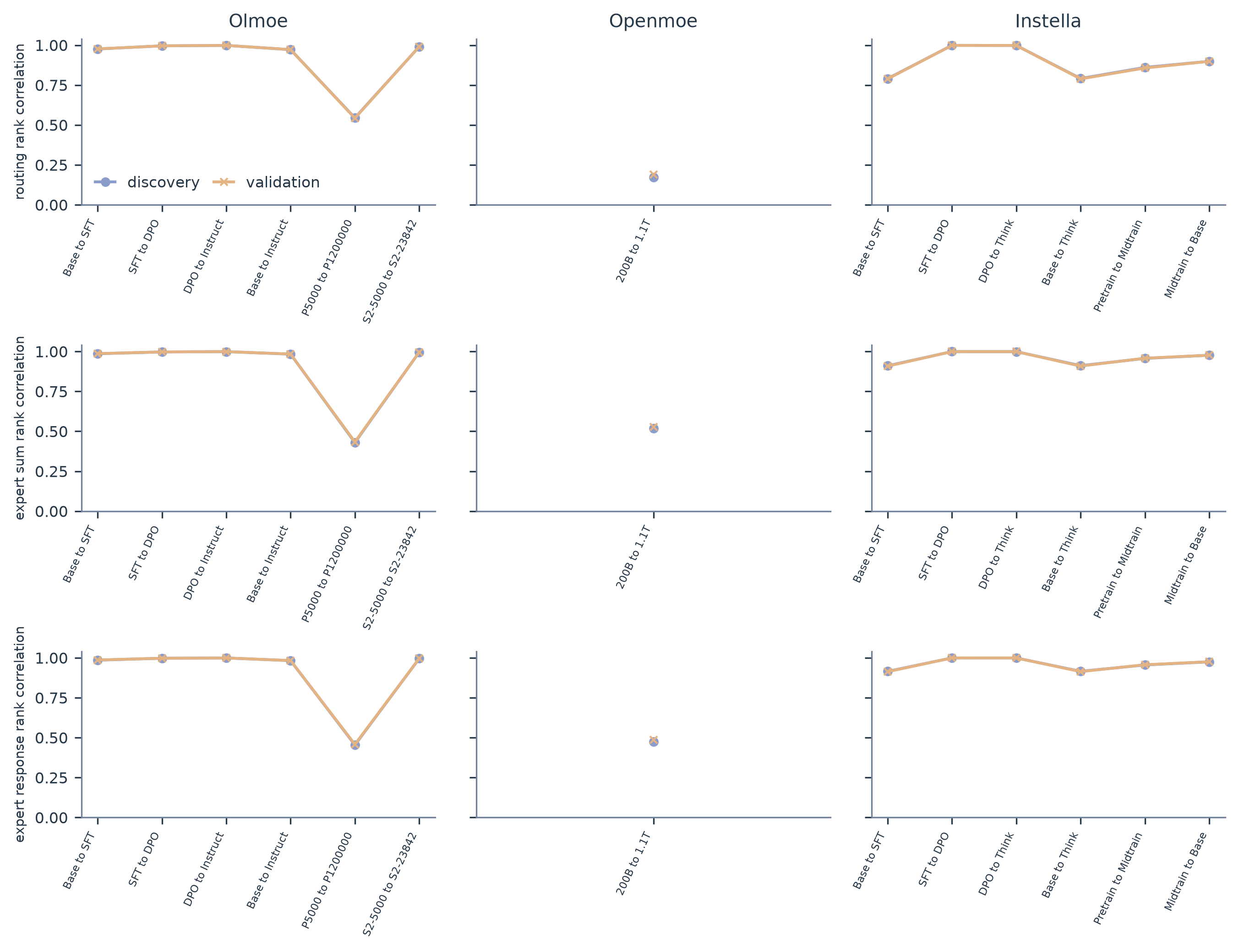}
\caption{Stage comparisons for dispatch counts, summed neuron importance and complete-output energy.
Values are domain--layer mean rank correlations for each interval, computed separately on fixed discovery and validation text.
Stage-pair positions are categorical and do not encode interval length.
OLMoE's P and S2 labels denote its separate pre-training and stage2 step counters.
The three statistics summarize expert-level rank stability.}
\label{fig:expert-stages}
\end{figure}

The accompanying data provide full discovery-side pairwise matrices, valid-cell counts and top-set sensitivity results.
The A/B halves partition the fixed document sets.
At the all-expert endpoint, both top sets contain every expert, giving Jaccard overlap one by construction.

\subsection{Neuron Changes under Nearly Stable Routing}
\label{app:evolution-frozen}
\begin{figure}[t]
\centering
\includegraphics[width=\textwidth]{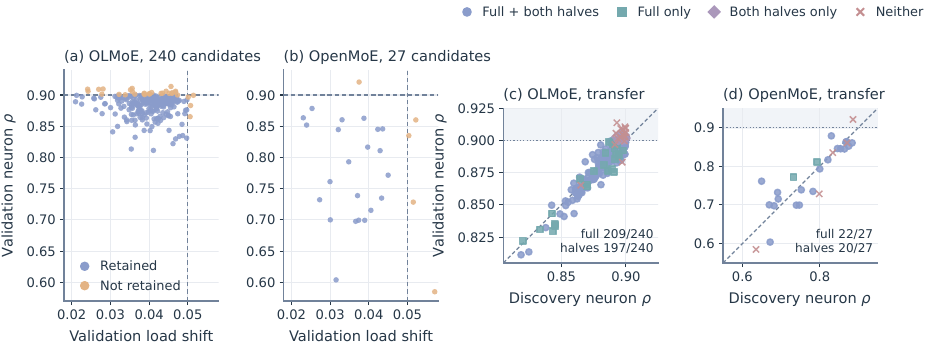}
\caption{Validation outcomes for all 267 frozen domain--layer transitions from OLMoE and OpenMoE pre-training.
(a, b) Validation routing-load shift against validation neuron-rank correlation; dashed lines are the selection thresholds, and orange points fail at least one criterion and remain in the denominator.
(c, d) Discovery versus validation neuron-rank correlation, marking outcomes on the full validation set and on both halves of 32 documents per bucket; the two tests are not nested and the panels use different axis ranges.}
\label{fig:candidate-scatter}
\label{fig:app-candidate-transfer}
\end{figure}

To examine neuron reorganization when routing loads change little, we select adjacent transitions combining nearly stable loads with changing neuron importance ranks.
Unlike the unfiltered comparison in Appendix~\ref{app:evolution-stage-level}, selection uses discovery text and evaluation uses separate validation documents; the scatter plot includes both retained and failed comparisons.

Each candidate is an adjacent pre-training checkpoint pair at a fixed domain and native layer.
The routing-load shift must be at most 0.05 and neuron-rank correlation below 0.90 on the full discovery set and each discovery half.
This selects 240 OLMoE and 27 OpenMoE candidates.
The lists and thresholds are fixed before evaluation on the full validation set and both validation halves.
Instella uses separate released-stage comparisons.

Neuron-rank changes coexist with nearly stable routing (\cref{fig:candidate-scatter}a--b) in 209/240 OLMoE and 22/27 OpenMoE candidates on the full validation set, and in 197/240 and 20/27 on both validation halves.
The 31 and five full-set failures remain in the respective denominators.
\Cref{fig:app-candidate-transfer}c--d relates discovery to validation ranks, distinguishing the non-nested full-set and both-halves outcomes.

\subsection{Exploratory Change Decomposition}
\label{app:evolution-decomposition}
\begin{figure}[t]
\centering
\includegraphics[width=\textwidth]{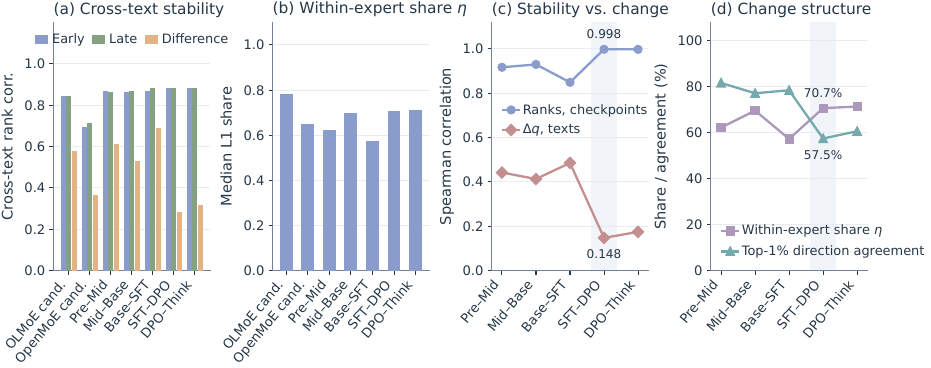}
\caption{Cross-text consistency and within-expert structure of neuron changes.
(a) Early- and late-checkpoint cross-text rank correlations and the correlation of importance-share differences.
(b) Median within-expert L1 share $\eta_c$ on jointly observed experts; pre-training groups are frozen candidates, while Instella groups cover all 286 domain--layer cells per stage pair.
(c) Instella's mean cross-checkpoint rank correlation contrasted with the median cross-text correlation of $\Delta q$.
(d) $\eta_c$ alongside pooled direction agreement of the top-1\% $\Delta q$ changes, retaining unavailable validation coordinates in the denominator.}
\label{fig:evolution-decomposition}
\label{fig:app-stability-vs-change}
\end{figure}

Neuron importance can change through redistribution within experts or shifts in experts' total importance shares.
For the fixed pre-training candidates in Appendix~\ref{app:evolution-frozen} and the Instella release series, a decomposition separates these contributions, while cross-text correlations test how consistently the changes recur.

They cover all 267 frozen pre-training candidates and, separately, all 286 domain--layer cells for each of Instella's five adjacent released-stage pairs.

For each domain--layer--checkpoint-pair cell $c$, normalize importance separately at each checkpoint $t$ and on each text split $s\in\{\mathrm{disc},\mathrm{val}\}$ as
\begin{equation}
\begin{aligned}
 P^{(s)}_{t,e,i}&=\frac{I_i^e(D_d^{(s)};t)}{\sum_{e',i'}I_{i'}^{e'}(D_d^{(s)};t)},\\
 m^{(s)}_{t,e}&=\sum_iP^{(s)}_{t,e,i},\qquad
 q^{(s)}_{t,e,i}=\frac{P^{(s)}_{t,e,i}}{m^{(s)}_{t,e}}.
\end{aligned}
\end{equation}
$D_d^{(s)}$ denotes the domain-$d$ text from split $s$; the domain and layer are fixed by $c$, with the layer index omitted from $I$.
Here $m$ is the expert's layer-importance share and $q$ is its conditional within-expert distribution when its mass is positive.
For zero-mass experts, set $q=0$ to preserve the identity below and exclude them from conditional-distribution comparisons.
With $\Delta P^{(s)}=P^{(s)}_{t_2}-P^{(s)}_{t_1}$ and analogous differences for $m$ and $q$, the exact symmetric decomposition is
\begin{equation}
\begin{aligned}
 \Delta P^{(s)}&=B^{(s)}+W^{(s)},\\
 B^{(s)}&=\Delta m^{(s)}\,\overline q^{(s)},\qquad
 W^{(s)}=\overline m^{(s)}\,\Delta q^{(s)}.
\end{aligned}
\end{equation}
Bars denote two-checkpoint means on the same text split.
Here $B$ measures whole-expert share changes and $W$ measures within-expert redistribution in normalized importance.
The signed components can cancel, so their L1 norms need not add to $\|\Delta P\|_1$.

The joint expert set $\mathcal J_c$ contains exactly the experts with positive executed dispatch count $C^{(s)}_{t,e}>0$ and positive importance mass $m^{(s)}_{t,e}>0$ at both checkpoints $t_1,t_2$ on both discovery and validation text.
Joint-restricted vectors contain all neuron coordinates of these experts.
The discovery-side within-expert norm share for each cell on this set is
\begin{equation}
 \eta_c=\frac{\|W^{(\mathrm{disc})}_{c,\mathcal J_c}\|_1}
 {\|W^{(\mathrm{disc})}_{c,\mathcal J_c}\|_1+\|B^{(\mathrm{disc})}_{c,\mathcal J_c}\|_1}.
\end{equation}
A zero denominator gives NA.
Group summaries are medians of valid per-cell $\eta_c$ values, not ratios of pooled L1 norms.
Cross-text conditional-change correlations compare $\Delta q^{(\mathrm{disc})}$ with $\Delta q^{(\mathrm{val})}$ on $\mathcal J_c$.
Within-component correlations compare $W^{(\mathrm{disc})}$ with $W^{(\mathrm{val})}$ on the same set.
Both are computed within cells and summarized by valid-cell medians.

For OLMoE candidates (\cref{fig:evolution-decomposition}a--b), median early- and late-checkpoint cross-text correlations are 0.84481 and 0.84572, compared with 0.58063 for importance-share differences.
OpenMoE gives 0.69639 and 0.71193, compared with 0.36777.
On the joint expert set, discovery-side within-expert L1 exceeds between-expert L1 in 240/240 and 27/27 cells, with median norm shares of 0.7830 and 0.6491.

Instella's five stage pairs cover 1430 cells, including all domain--layer cells for each pair without applying the pre-training selection criteria.
The discovery-side joint-restricted within-expert component is larger in 266/286, 282/286, 247/286, 286/286 and 284/286 cells in chronological stage-pair order.
Median norm shares are 0.6221, 0.6963, 0.5728, 0.7067 and 0.7138.
The final two pairs have median cross-text conditional-change correlations of 0.1479 and 0.1750.
Late post-training changes thus remain mostly within experts and are small with little consistent direction across texts, consistent with post-training preserving the organization formed earlier.

\subsection{Replication of Large Neuron-Level Changes}
\label{app:evolution-replication}
Overall ranking stability and the cross-text consistency of neuron-level changes are distinct properties, as separated in \cref{fig:app-stability-vs-change}c--d.
Selecting the largest discovery-side changes tests whether they hold on validation text in the fixed pre-training candidates and Instella release series.
Direction agreement and change-rank correlations distinguish consistent directions of adjustment from consistent rankings of their magnitudes.

For each cell and change type, selection uses eligible discovery-side coordinates with nonzero absolute change.
Layer-share changes use $\delta=\Delta P$ over layer coordinates.
Within-expert-share changes use $\delta=\Delta q$ and require positive expert dispatch count and importance mass at both discovery checkpoints.
Validation eligibility is assessed after selection, so this discovery pool is not restricted to $\mathcal J_c$.

For each fraction $f\in\{0.01,0.05,0.10,0.20,1.00\}$, select the $k=\max(1,\operatorname{round}(N_cf))$ coordinates with largest $|\delta^{(\mathrm{disc})}|$ from the cell's $N_c$ eligible nonzero coordinates.
Set $k=0$ when $N_c=0$.
Stable flattened coordinate order breaks ties at the selection boundary.
The fractions correspond to 1\%, 5\%, 10\%, 20\% and 100\%.
The 100\% list contains the full eligible discovery-nonzero pool.
All lists remain fixed for validation evaluation.

For a group $G$ at a fixed fraction and change type, let $n_c^{\mathrm{selected}}$ count all selected coordinates in cell $c$ and $n_c^{\mathrm{same}}$ count those with observable validation changes of the same sign as discovery.
Direction agreement pools these integer counts as
\begin{equation}
 A_G=\frac{\sum_{c\in G}n_c^{\mathrm{same}}}{\sum_{c\in G}n_c^{\mathrm{selected}}}.
\end{equation}
Selected coordinates unobservable on validation text remain in the denominator and contribute zero to the numerator.
Their counts are reported separately.
For $\Delta q$, validation observability requires routing and positive mass at both validation checkpoints.
A selected nonzero discovery change with zero validation change also contributes zero to the numerator.
Coordinates selected in multiple checkpoint-pair cells are counted at each occurrence.
The group rate is coordinate-weighted, not an equal-weight mean of per-cell rates.

On selected coordinates available in both splits, compute separate within-cell Spearman correlations between signed changes $\delta^{(\mathrm{disc})}$ and $\delta^{(\mathrm{val})}$, and between their absolute magnitudes.
Constant vectors or insufficient available coordinates give NA.
Signed-change and absolute-magnitude correlations are each summarized by the median over their valid cells.

\begin{figure}[!htbp]
\centering
\includegraphics[width=0.9\textwidth]{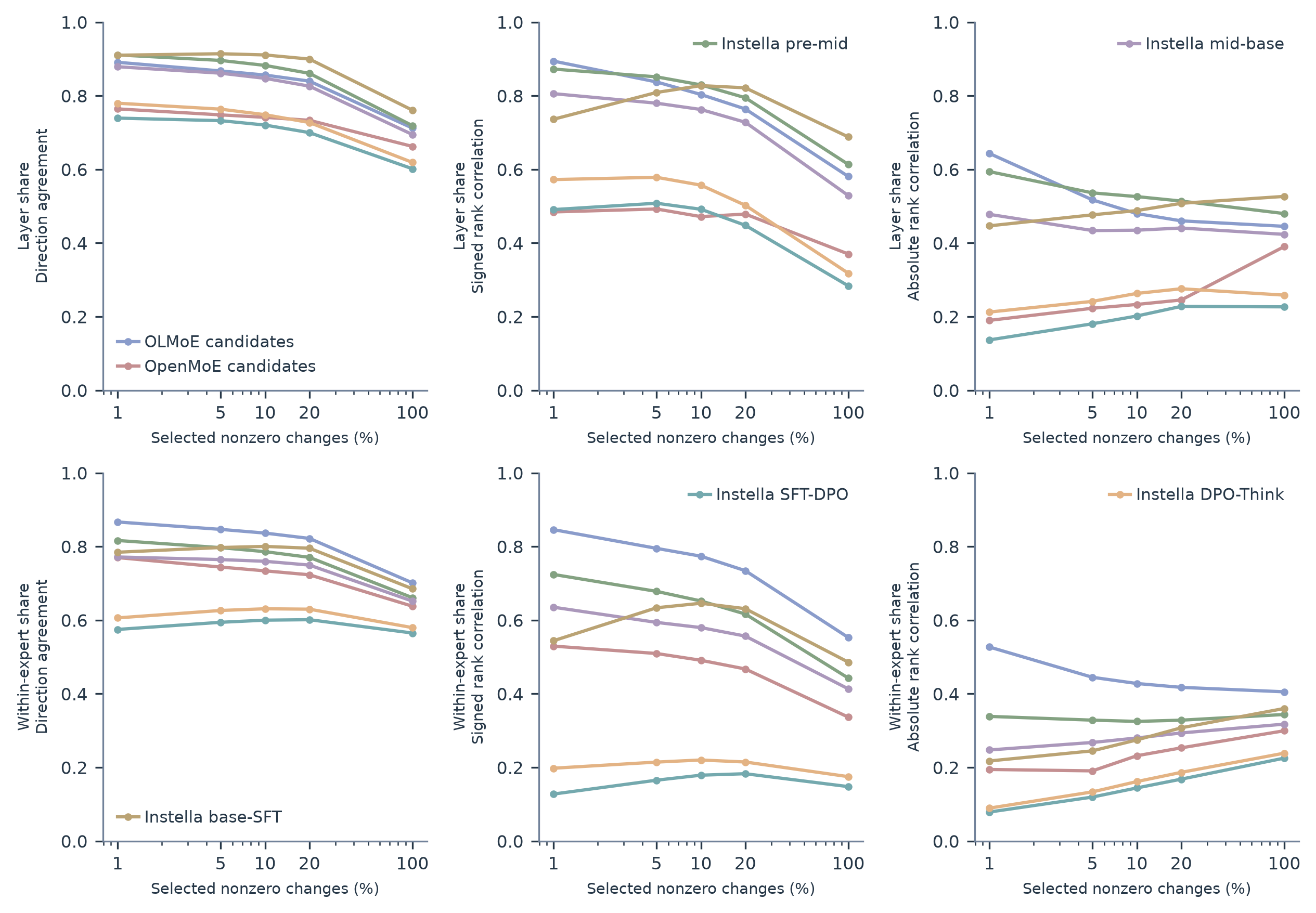}
\caption{Exploratory replication of discovery-selected changes.
All five selection fractions are shown for layer-share and within-expert-share differences.
Direction agreement is the pooled same-direction count divided by the pooled selected count, retaining unavailable validation coordinates in the denominator.
Signed-change and absolute-magnitude Spearman correlations are computed separately on available selected coordinates and summarized by valid-cell medians.}
\label{fig:evolution-large-changes}
\end{figure}

For layer-share changes, OLMoE direction agreement rises from 71.22\% over all eligible discovery-nonzero coordinates to 89.08\% in the largest 1\%.
OpenMoE rises from 66.24\% to 76.43\%, so in both families the largest changes keep their direction across text splits more often than typical changes.
For OpenMoE, the ranking of their magnitudes is less stable, with median absolute-rank correlation falling from 0.3903 to 0.1899.

For Instella SFT-to-DPO within-expert changes, the largest 1\% has direction agreement 57.48\% and median signed-rank correlation 0.1277, compared with 0.1479 over the full eligible discovery-nonzero set.
The corresponding DPO-to-Think values at 1\% are 60.62\% and 0.1979.
The denominators retain validation-unavailable coordinates, including 3448 selected coordinates for SFT-to-DPO at 1\%.
The seven groups, two change types and five fractions yield 70 summaries, with repeated coordinate occurrences counted across candidate comparisons.
Across Instella's five stage pairs, \cref{fig:app-stability-vs-change}c--d separates overall checkpoint rank stability from the cross-text consistency of changes.

\subsection{Domain Neuron Sets Across Checkpoints}
\label{app:evolution-sets}
\begin{figure}[t]
\centering
\includegraphics[width=\textwidth]{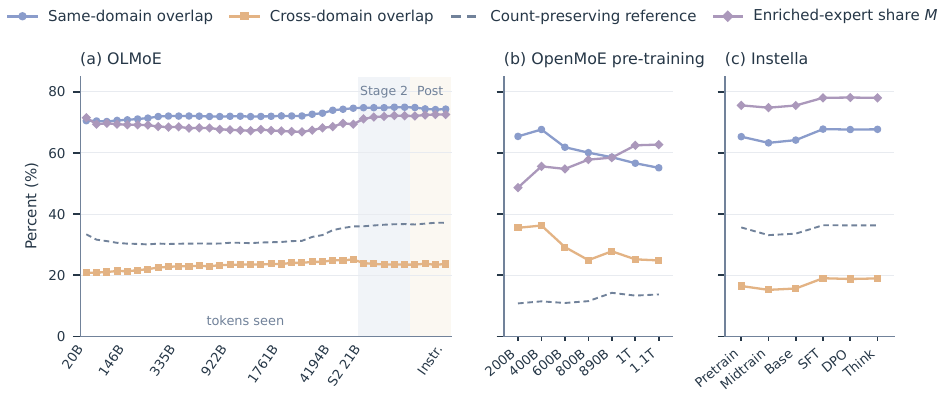}
\caption{Domain neuron-set statistics of \cref{sec:neuron-organization} computed at every checkpoint, at 6.25\% selection under routing-weighted RAMP.
Same- and cross-domain overlaps follow \cref{eq:discovery-overlap}, the count-preserving reference follows \cref{eq:expert-count-null}, and $M$ is the share of each domain's exclusive neurons held by its enriched experts, as in \cref{tab:exclusive-enrichment}.
Values are averaged over domains and layers.
(a) OLMoE pre-training labelled by tokens seen, followed by the stage-2 and post-training checkpoints, ending at the OLMoE-0125-Instruct model of \cref{sec:neuron-organization}.
(b) OpenMoE pre-training. (c) Instella's released stages.}
\label{fig:evolution-sets}
\end{figure}

\Cref{sec:evolution} compares neuron importance rankings between adjacent checkpoints.
A changing or stable ranking does not by itself show whether the domain specialization measured in \cref{sec:neuron-organization} exists at a given checkpoint.
This subsection therefore asks two questions directly: when does domain specialization appear during training, and does it persist afterward?
We therefore apply the set analyses of \cref{sec:discovery-replication,sec:discovery-spatial,sec:exclusive-enrichment} directly at every checkpoint, reusing the neuron importance computed for \cref{sec:evolution} on the same discovery and validation text.
At each checkpoint we select the 6.25\% sets, compute same-domain overlap, cross-domain overlap and the count-preserving reference, and compute the enriched-expert share $M$ from the exclusive neurons, with the definitions and averaging of \cref{sec:neuron-organization}.

\textbf{Answer.} Domain specialization is already present at the earliest available checkpoint of every model and then stays nearly constant through the rest of training (\cref{fig:evolution-sets}).
After 20B tokens of OLMoE pre-training, same-domain overlap is 70.53\% against 20.80\% across domains, a gap of 49.74 points, and over all 36 OLMoE checkpoints through post-training the gap stays between 48.05 and 51.50 points while $M$ stays between 66.89\% and 72.58\%.
At the final checkpoint, the overlaps of 74.33\% and 23.68\% equal those reported for OLMoE in \cref{sec:discovery-replication}.
OpenMoE keeps a gap of 29.91--35.15 points from 200B to 1.1T tokens, while both overlaps decline and $M$ rises from 48.65\% to 62.73\%, so its exclusive neurons concentrate into fewer experts as training proceeds.
Instella keeps a gap of 48.08--48.88 points across its six released stages; Base-to-SFT raises same-domain overlap from 64.20\% to 67.79\% and $M$ from 75.52\% to 78.03\%, and the later stages change each statistic by at most 0.25 points.
Observed overlap exceeds the count-preserving reference at every checkpoint, by 29.69--56.22 points across the three models.

These results separate two things that \cref{sec:evolution} alone cannot.
Which neurons carry each domain keeps changing during pre-training, as OLMoE's neuron ranking correlates at only 0.337 between its first and last pre-training checkpoints (\cref{sec:neuron-pretraining}).
How strongly the neurons specialize by domain does not change, as the same--cross gap stays between 48.05 and 49.74 points over the same pre-training checkpoints.
The post-training results agree with \cref{sec:neuron-posttraining}, as the only visible change in Instella occurs at Base-to-SFT.

\section{Additional Pruning Results and Protocols}
\label{app:pruning-protocol}

Cross-domain sharing and domain-dependent importance motivate the shared neuron selection in \cref{sec:pruning}.
The aim is to preserve broadly important neurons alongside those important to particular domains within a limited budget.

\subsection{Calibration and Score Definitions}
The shared masks in \cref{sec:pruning} require a choice between preserving the cumulative importance scale and giving each domain a comparable scale before selection.
These choices define the two aggregation rules, while calibration budgets distinguish the quality and implementation studies.

The n128 main table uses 128 texts per calibration domain, while n64 paired and efficiency experiments use 64, each truncated to 2048 tokens.
The eleven buckets contain the nine discovery buckets plus finance-mcq and law-mcq.
Each experiment uses its recorded sampling and inference profile.
RAMPr-L sums domain scores after each domain's squared activation statistics have been accumulated and square-rooted, rather than pooling all domain squares before taking a single square root.
RAMPr-D normalizes each domain by its layer-wide importance sum $Z_l(D_d)$, floored at $10^{-12}$, and takes the cross-domain maximum.
Stored dispatch counts do not normalize either score.

Both rules use per-layer top-$k$ selection over routed intermediate neurons.
Zero masking keeps the original weight shapes, while physical layouts store selected rows and columns with possible padding.
Effective parameter count and allocated layout size therefore require separate reporting; neither follows from scaling total model size by the retention ratio.

\subsection{Relation to Other MoE Pruning Methods}
\label{app:pruning-related-scores}
MoE compression methods differ in the unit they remove, the signal that ranks it and the text that estimates this signal (\cref{tab:pruning-score-comparison}).
We group them into expert removal, expert merging and within-expert pruning, and write their scores in the notation of \cref{sec:ramp}, with $\mathcal T_e$ the tokens dispatched to expert $e$, $f^e(t)$ the output of expert $e$ for token $t$, $h(t)$ and $\tilde h(t)$ the hidden states entering and leaving the MoE layer, $\mathcal L$ the language-modeling loss and $h$ the hidden size.

\textbf{Expert removal.}
Early methods search for the expert subset to keep, enumerating subsets that best reconstruct each layer's output on calibration text \citep{lu2024notallexperts}, keeping the experts whose router weights change most during fine-tuning \citep{chowdhury2024provably}, or evolving pruning and merging decisions against task performance \citep{liu2024efficientexpert}, and a unified study compares such dropping criteria across models \citep{jaiswal2025fantastic}.
Score-based methods instead rank experts in one forward pass.
REAP \citep{lasby2026reap} averages the router-weighted output norm over dispatched tokens,
\begin{equation}
S^e_{\mathrm{REAP}}=\frac{1}{|\mathcal T_e|}\sum_{t\in\mathcal T_e}p^e(t)\,\bigl\|f^e(t)\bigr\|_2 .
\end{equation}
RAZOR \citep{song2026razor} measures the output change caused by deleting an expert, using its residual $r^e(t)=f^e(t)-\sum_{e'}p^{e'}(t)f^{e'}(t)$ from the routed mixture and the expert $\hat e$ that the router selects in its place,
\begin{equation}
S^e_{\mathrm{RAZOR}}=\Biggl(\frac{1}{|\mathcal T_e|}\sum_{t\in\mathcal T_e}\biggl[\frac{\bigl\|p^e(t)r^e(t)-p^{\hat e}(t)r^{\hat e}(t)\bigr\|_2}{1-p^e(t)+p^{\hat e}(t)}\biggr]^2\Biggr)^{1/2}.
\end{equation}
EASY-EP \citep{dong2025easyep} prunes experts for a target domain from a few demonstrations, weighting the router-weighted output norm by how much each token is changed by the layer,
\begin{equation}
S^e_{\mathrm{EASY}}(D_d)=\sum_{t\in D_d}p^e(t)\bigl\|f^e(t)\bigr\|_2\bigl[1-\cos\bigl(h(t),\tilde h(t)\bigr)\bigr],
\end{equation}
and serves several domains by summing $S^e_{\mathrm{EASY}}(D_d)/\sum_{e'}S^{e'}_{\mathrm{EASY}}(D_d)$ over domains.
A coding-oriented study prunes half of the experts from routing and activation statistics collected on code \citep{jha2026halfexperts}.

\textbf{Expert merging.}
Merging methods keep the information of removed experts by averaging them into retained ones.
MC-SMoE \citep{li2024mcsmoe} groups experts by the similarity of their router logits, merges each group with activation-frequency weights and further compresses the merged experts.
HC-SMoE \citep{chen2025hcsmoe} clusters experts hierarchically by their average outputs on calibration text before frequency-weighted merging.
DERN \citep{zhou2025dern} drops experts with low routing statistics, splits them into neuron segments, assigns each segment to the most similar retained expert and merges segments within each expert by clustering.

\textbf{Within-expert pruning.}
Hybrid methods combine expert-level and weight-level steps, such as expert pruning by router-weight similarity followed by unstructured pruning \citep{lee2025stun}, or inter-expert pruning followed by intra-expert low-rank decomposition \citep{yang2024moei2}.
MoE-Pruner \citep{xie2024moepruner} removes individual weights by their magnitude times the router-weighted input activation, and SlimMoE \citep{li2025slimmoe} narrows experts over several stages with distillation.
Methods that remove whole neurons rank them as follows.
Fisher pruning \citep{he2026less} uses the squared loss gradients of a neuron's three weight vectors,
\begin{equation}
s^{e}_{i,\mathrm{F}}=\frac1h\sum_{k=1}^{h}\Bigl[\bigl(G^{e}_{\mathrm{gate}}\bigr)_{ik}^2+\bigl(G^{e}_{\mathrm{up}}\bigr)_{ik}^2+\bigl(G^{e}_{\mathrm{down}}\bigr)_{ki}^2\Bigr],
\end{equation}
where $G^e_m=\nabla_{W^e_m}\mathcal L$, and the expert-local rule used in our comparison keeps the same fraction of every expert.
HEAPr \citep{li2026heapr} approximates the loss change from removing a neuron with a second-order term in the expert output space and ranks neurons across all layers,
\begin{equation}
s^{e}_{i,\mathrm{H}}=\frac{1}{|\mathcal T_e|}\sum_{t\in\mathcal T_e}\tfrac12\,u_i^e(t)^{\top}\bar G^{e}\,u_i^e(t),
\end{equation}
where $u_i^e(t)=a_i^e(t)w_i^{e,\mathrm{down}}$ and $\bar G^e$ is the covariance of $\nabla_{f^e}\mathcal L$.
TENP \citep{he2026tenp} keeps the experts ranked highest by an EASY-EP-style score, retaining more of them in deeper layers, and prunes neurons in the remaining experts by their projection onto the expert output,
\begin{equation}
s^{e}_{i,\mathrm{T}}=\frac{1}{|\mathcal T_e|}\sum_{t\in\mathcal T_e}\frac{\bigl\langle u_i^e(t),f^e(t)\bigr\rangle}{\bigl\|f^e(t)\bigr\|_2}.
\end{equation}

\begin{table*}[t]
\caption{MoE compression methods compared by the unit they act on, the operation, the signal that ranks units and the calibration text.}
\label{tab:pruning-score-comparison}
\centering\small
\setlength{\tabcolsep}{4pt}
\rowcolors{2}{tabrow}{}
\begin{tabularx}{\textwidth}{lllLl}
\toprule
\rowcolor{tabhead}Method & Unit & Operation & Ranking signal & Calibration \\
\midrule
NAEE & Expert & Remove & Reconstruction of layer output & General text \\
REAP & Expert & Remove & Router-weighted output norm & General and code \\
RAZOR & Expert & Remove & Output change after deletion & Seven-domain pool \\
EASY-EP & Expert & Remove & Router-weighted output norm, token-weighted & Target domains \\
HC-SMoE & Expert & Merge & Clusters of expert outputs & General text \\
DERN & Expert, neuron & Remove, merge & Routing statistics, segment similarity & General text \\
Fisher & Neuron & Remove & Squared weight gradients & One task \\
HEAPr & Neuron & Remove & Second-order loss change & General text \\
TENP & Expert, neuron & Remove & Expert score, then neuron projection & Target domains \\
RAMPr & Neuron & Remove & Routed contribution per domain & Eleven domains, $S^L$ or $S^D$ \\
\bottomrule
\end{tabularx}
\end{table*}

\textbf{Relation to RAMPr.}
These methods share two observations that RAMPr also relies on: the router weight is part of a unit's contribution to the layer output, and neuron-level units allow finer budgets than whole experts.
RAMPr differs in purpose, since it turns the domain specialization of \cref{sec:neuron-organization} into a pruning rule, and three design choices follow from it.
First, RAMP accumulates a neuron's routed contribution over all tokens of a domain, ranking neurons by their total contribution to that domain, which is the quantity whose domain dependence \cref{sec:neuron-organization} establishes.
Second, most methods estimate one ranking from general text, a single task or a pooled mixture, and EASY-EP and TENP combine domains only at the expert level, whereas RAMPr ranks neurons separately in each of eleven domains across experts, so domain-specific neurons inside mixed experts (\cref{sec:exclusive-enrichment}) are kept even when their expert or their pooled importance is ranked low, and $S^D$ needs a neuron to rank high in only one domain.
Third, because importance is estimated domain by domain, calibration composition becomes a direct control over which capabilities are preserved (Appendix~\ref{app:pruning-calibration}).

\subsection{Task Metrics and Main-Table Results}
\label{app:pruning-metrics}
To assess capability preservation in \cref{sec:pruning}, raw-score averages and Base-relative retention provide complementary views of the same task results.
Individual scores show how gains and losses vary across models and retention budgets.

WinoGrande \citep{sakaguchi2020winogrande}, C-Eval \citep{huang2023ceval}, CMMLU \citep{li2024cmmlu}, MMLU \citep{hendrycks2021mmlu} and PubMedQA \citep{jin2019pubmedqa} use accuracy.
HellaSwag \citep{zellers2019hellaswag} uses normalized accuracy.
MBPP \citep{austin2021mbpp} and HumanEval \citep{chen2021humaneval} use pass@1.
GSM8K \citep{cobbe2021gsm8k} is scored by exact match under the strict-match filter in every row, and MATH-500 uses the MATH benchmark's mathematical problems \citep{hendrycks2021math} with math verification.
GPQA Diamond \citep{rein2024gpqa} and AIME use exact match.
GPQA answers are generated, whereas PubMedQA scores answer options.
Quality evaluation uses vLLM 0.29.0 with lm-eval 0.4.13; runtime-efficiency evaluation uses a separately modified Transformers implementation.
Appendix~\ref{app:evaluation-settings} summarizes the sampling settings and output budgets of the generation tasks.
Each calibration bucket is a fixed file of 128 documents, and every experiment shuffles it once with seed 1234 and takes the first $k$ documents, so the n64 and calibration-size studies use nested subsets of the n128 calibration set.
The nine buckets shared with \cref{sec:discovery-setup} are prepared from training or development splits of the sources in \cref{tab:discovery-sources}, deduplicated within each bucket using the first 256 characters of processed text, and contain no exact duplicates.
Overlap audits against AIME25, AIME26, GPQA Diamond and PubMedQA find no shared questions.
The finance and law buckets are drawn from MMLU professional\_accounting and professional\_law questions that also appear in the MMLU test set, covering 256 of its 14{,}042 items (1.82\%), so their effect on the MMLU score is bounded by this small fraction, and the nine-bucket analyses of \cref{sec:neuron-organization} do not use them.

Mean performance retention is
\begin{equation}
R=\frac{100}{|T|}\sum_{t\in T}\frac{s_{\mathrm{method},t}}{s_{\mathrm{Base},t}},
\end{equation}
where $T$ contains tasks scored for both methods with positive Base scores.
The AIME tasks have zero Base scores for Qwen1.5 and OLMoE and are excluded only from their ratio mean; individual zero scores remain in the task tables.
The separate eleven-task raw macro-average excludes AIME for every model and does not divide by Base.
Partial-coverage Fisher rows are not compared using a full-coverage ratio mean.

\begin{figure}[!htbp]
\centering
\includegraphics[width=\textwidth]{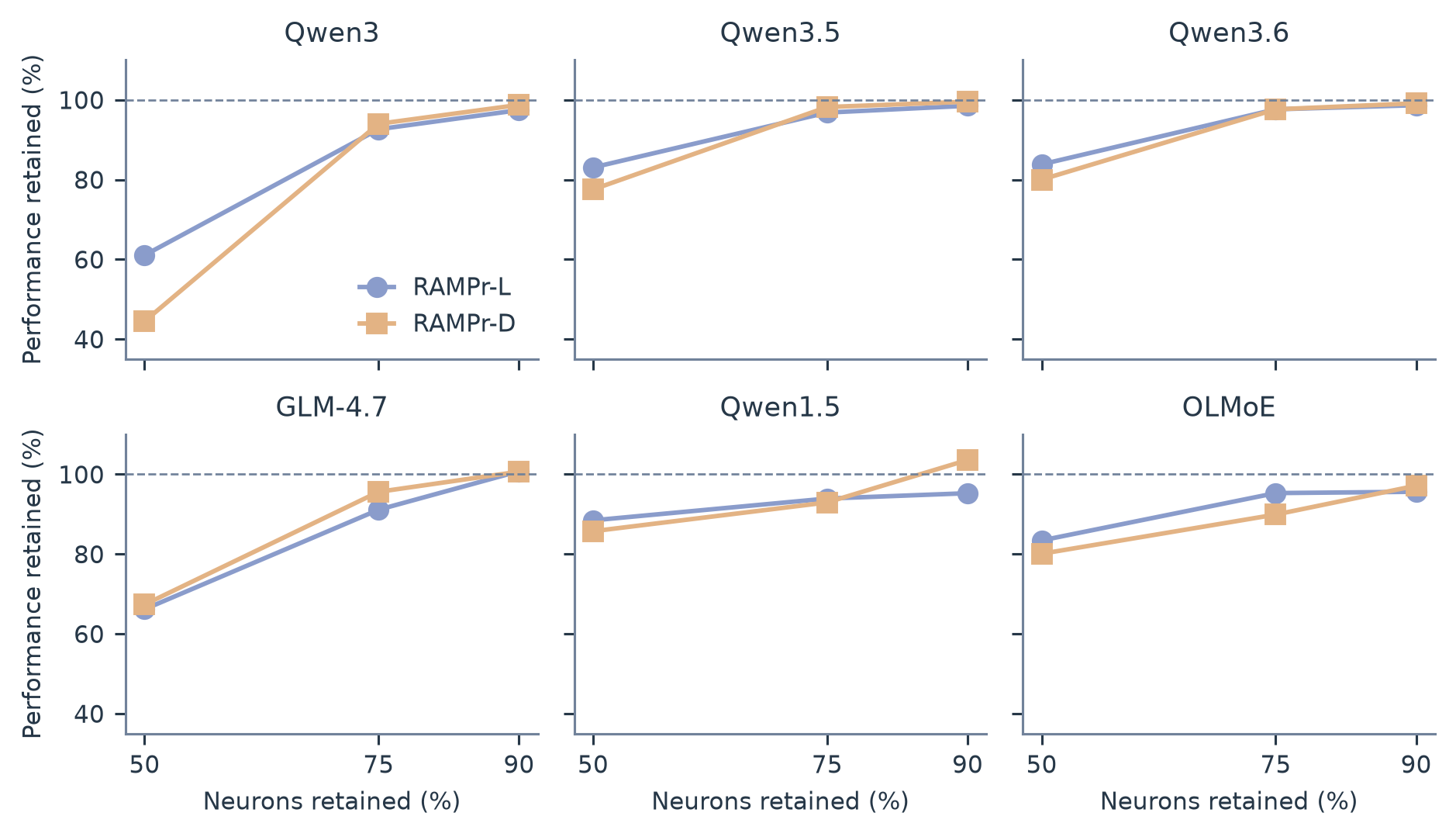}
\caption{n128 mean performance retention for both aggregation rules at all three neuron-retention rates.
Ratios are averaged over tasks with nonzero Base scores.
Qwen1.5 and OLMoE use eleven tasks, and the remaining complete rows use thirteen.
Values exceeding 100\% are retained.
Lines connect only the evaluated budgets.}
\label{fig:pruning-retention}
\end{figure}

\begin{figure}[!htbp]
\centering
\includegraphics[width=\textwidth]{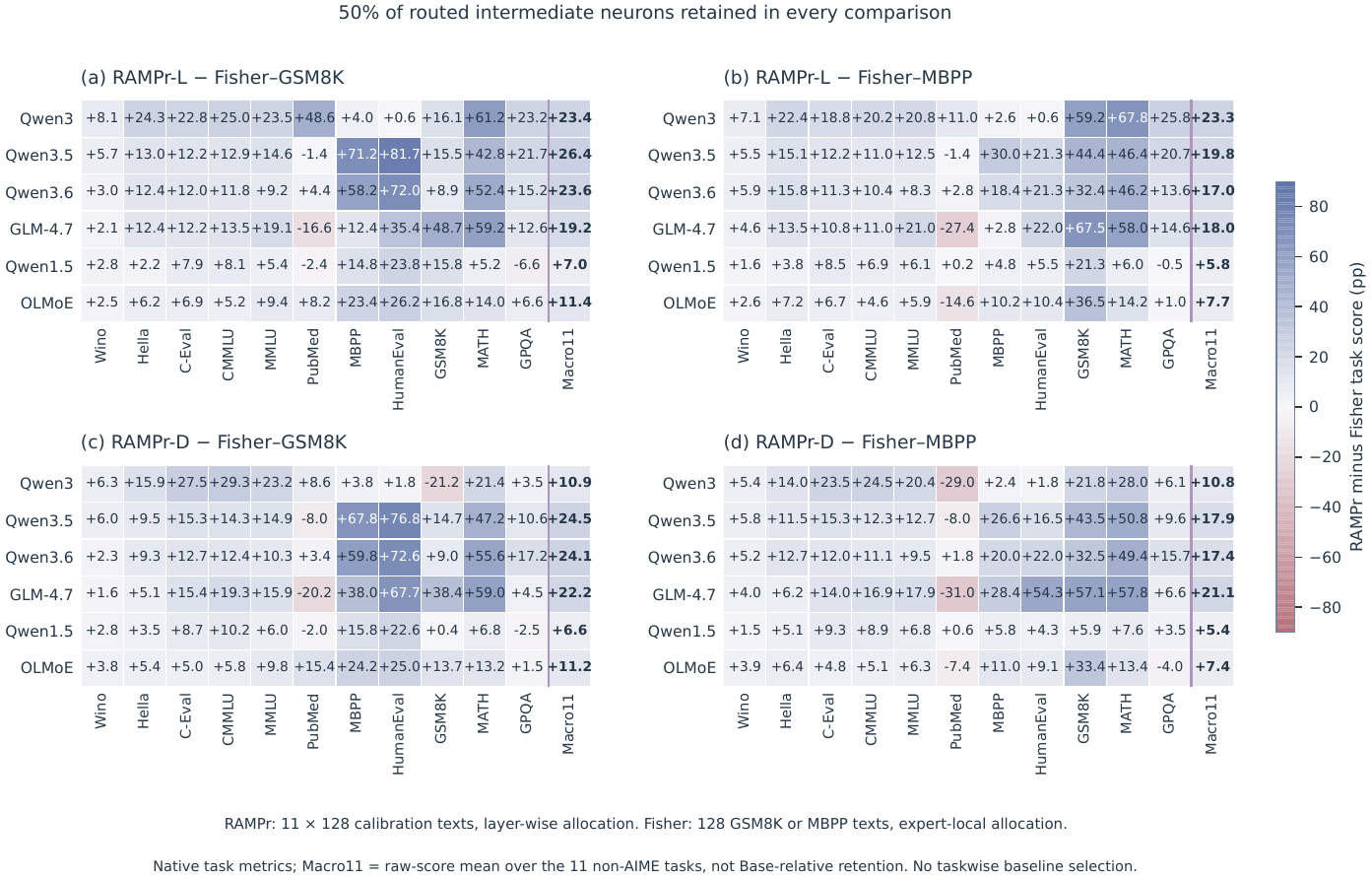}
\caption{All 264 task-score differences (six models, eleven non-AIME tasks and four RAMPr--Fisher comparisons) at 50\% routed-neuron retention, in percentage points.
RAMPr-L/D use $11\times128$ calibration texts and layer-wise allocation, while each Fisher baseline uses 128 GSM8K or MBPP texts and expert-local allocation.
Macro11 averages raw task scores rather than Base-relative retention.
All 24 macro differences are positive, and 246 of 264 task differences are non-negative, with the remaining negative differences retained.}
\label{fig:app-fisher-task-deltas}
\end{figure}

The full main table contains 92 method--model rows and 764 task results.
The two full-task Fisher calibrations use expert-local squared-gradient selection, while RAMPr uses multiple domains and layer-wise allocation.
\Cref{fig:app-fisher-task-deltas} resolves the positive eleven-task macro differences into individual task outcomes, including tasks where Fisher scores higher.

\begin{figure}[!htbp]
\centering
\includegraphics[width=\textwidth]{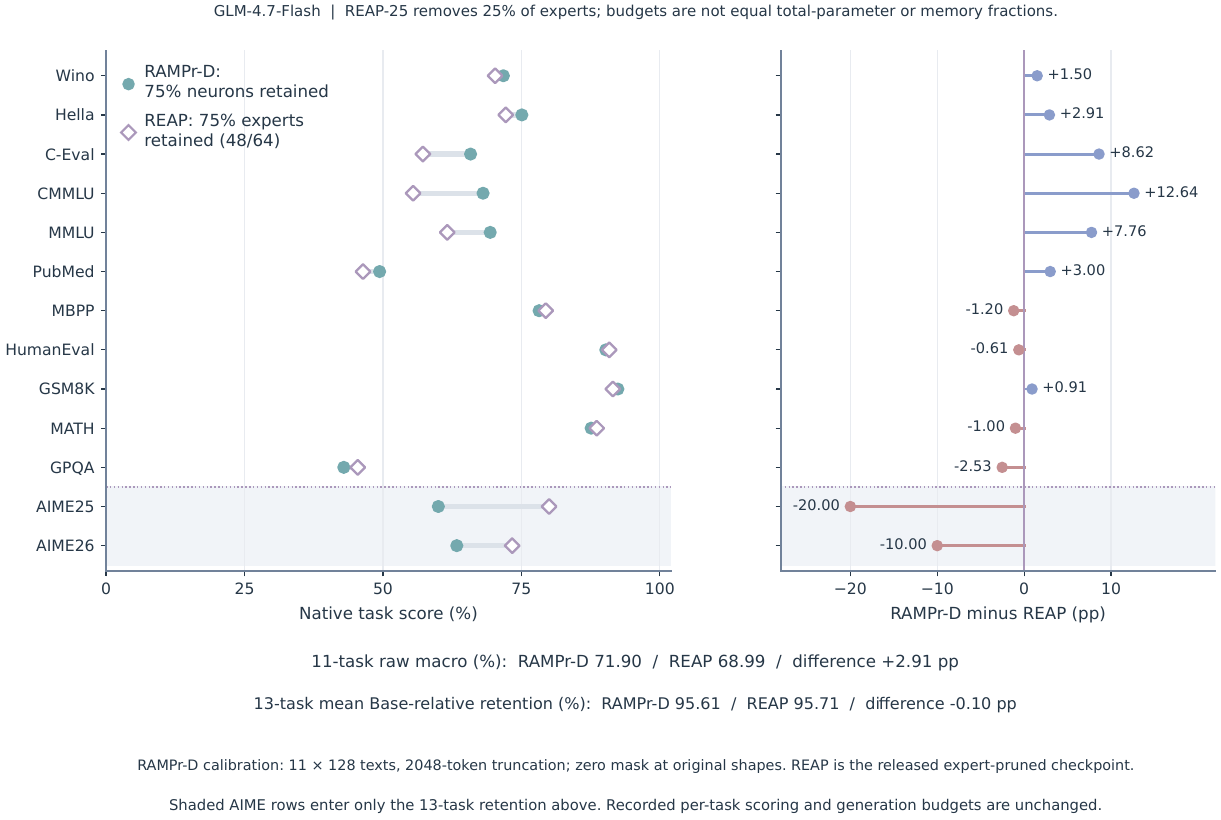}
\caption{GLM-4.7-Flash task scores and RAMPr-D-minus-REAP differences for all thirteen tasks.
RAMPr-D retains 75\% of routed neurons, whereas REAP-25 retains 75\% of experts (48/64).
RAMPr-D has the higher eleven-task raw macro-average (71.90\% versus 68.99\%, a difference of $+2.91$ percentage points).
Its thirteen-task mean Base-relative retention including AIME is nearly identical (95.61\% versus 95.71\%).}
\label{fig:app-glm-reap-tradeoff}
\end{figure}

REAP removes whole experts \citep{lasby2026reap}, defining its 25\% and 50\% compression settings by the fraction of experts pruned per layer.
Qwen3 REAP-50 retains 64 of the original 128 routed experts, each at its original width.
The GLM row, \texttt{reap25}, evaluates \path{cerebras/GLM-4.7-Flash-REAP-23B-A3B} based on \path{zai-org/GLM-4.7-Flash}.
Each of its 46 routed layers retains 48 of the original 64 experts, removing 25\% and retaining 75\%.
Intermediate width 1536, top-$K=4$, and the shared-expert configuration remain unchanged.
The initial dense layer is outside this pruning population.
The released model card labels the checkpoint 23B total and A3B activated.
Its next-token-prediction-layer configuration changes from one to zero, while the main evaluation uses the standard task inference described here.
\Cref{fig:app-glm-reap-tradeoff} shows that RAMPr-D's eleven-task raw-score advantage over REAP on GLM comes mainly from knowledge tasks, while its thirteen-task Base-relative retention stays within 0.10 points of REAP.

\subsection{Neuron Retention and Whole-Expert Selection}
\label{app:width-retention}

\begin{figure}[!htbp]
\centering
\includegraphics[width=0.9\textwidth]{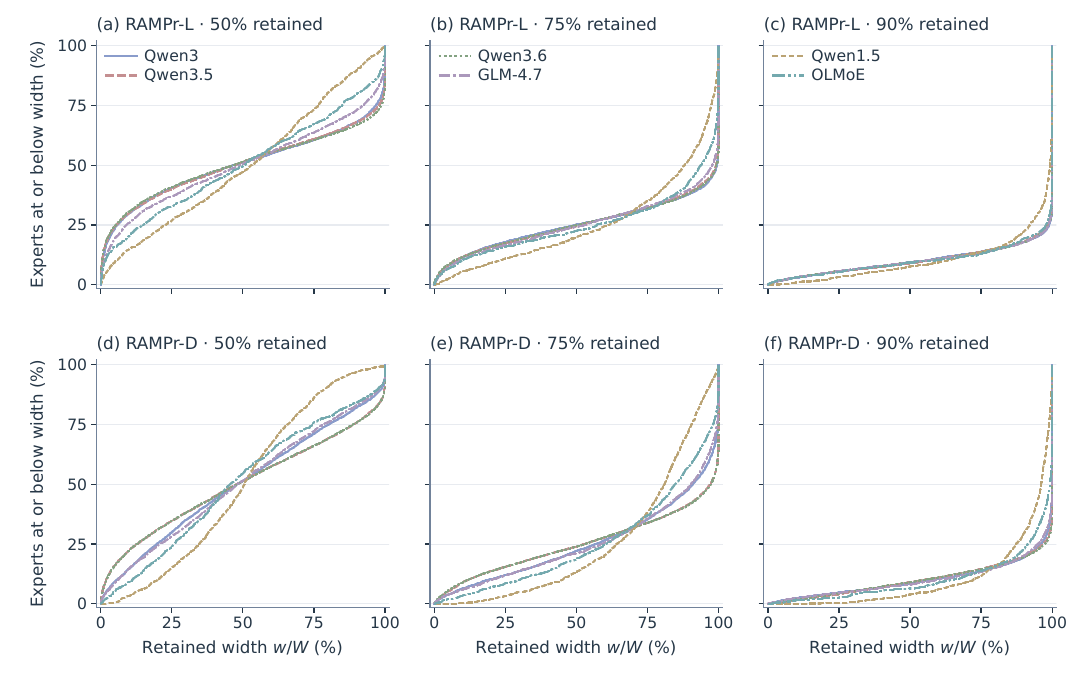}
\caption{Complete empirical distributions of normalized retained width for all 36 n64 masks.
Rows show RAMPr-L/D and columns show neuron-retention budgets; the six-model legend is shared across panels.
Every native integer width is represented without smoothing, including zero-width experts.
Fixed expert counts within each model make pooled and layer-equally weighted ECDFs identical.
Shared experts, GLM's initial dense layer, and layout padding are excluded.}
\label{fig:width-retention-cdf}
\end{figure}

The relation between neuron selection and whole-expert pruning depends on how retained coordinates are distributed across experts.
The comparison covers all 36 fixed n64 masks from six models, both RAMPr variants, and 50\%, 75\%, and 90\% neuron retention.
For expert $e$ in layer $l$, $w_{le}/W$ measures retained intermediate coordinates relative to the original expert width $W$.
The distributions in \cref{fig:width-retention-cdf} show that many experts retain partial neuron sets rather than lying near the deleted or complete endpoints.
At 50\% retention, experts with $0.1<w_{le}/W<0.9$ account for 36.23--74.65\% under RAMPr-L and 53.20--93.54\% under RAMPr-D across models.
RAMPr-D has more experts in this intermediate range in all eighteen model--budget comparisons, consistent with a less polarized allocation.
Even at 90\% retention, 14.50--29.51\% of experts remain in this intermediate range across models and methods.

\begin{figure}[!htbp]
\centering
\includegraphics[width=0.9\textwidth]{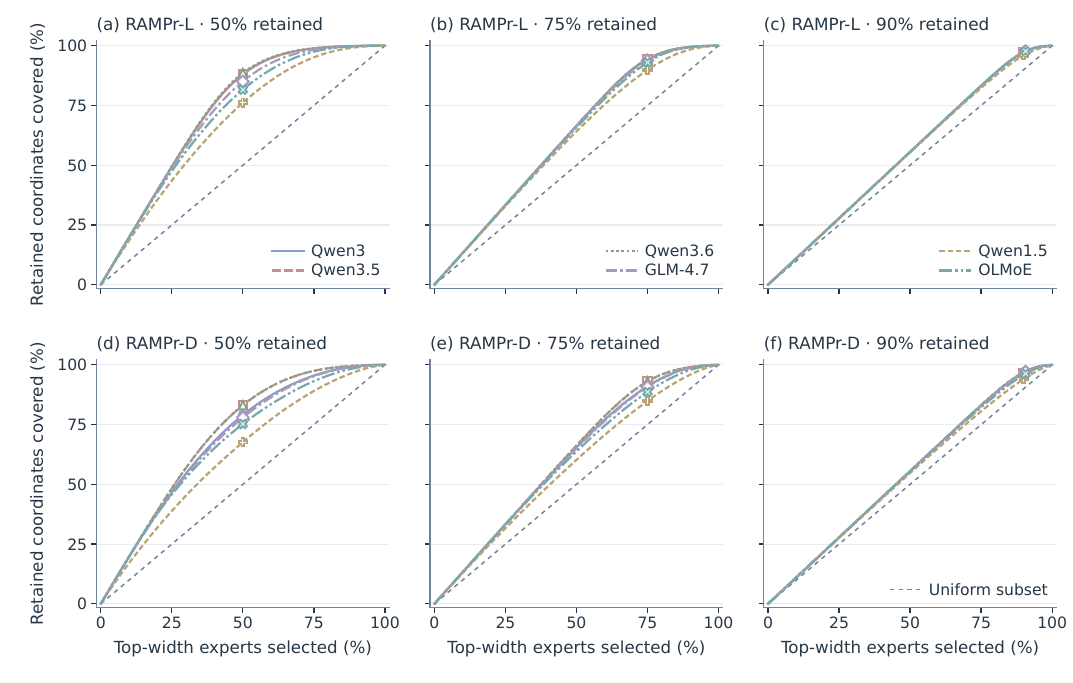}
\caption{Layer-equally weighted retained-coordinate coverage, $100G(k)$, at every integer expert count $k$, with the same panel and model ordering as \cref{fig:width-retention-cdf}.
The diagonal gives uniform-subset expected percentage coverage $100k/E$; markers identify $k=\lfloor rE+1/2\rfloor$ for neuron-retention fraction $r$.
Expert budgets are exactly matched at 50\% and 75\%; at nominal 90\%, actual fractions are 89.84375\% for Qwen3/3.5/3.6, 90\% for Qwen1.5, and 90.625\% for GLM-4.7/OLMoE.}
\label{fig:retained-coordinate-coverage}
\end{figure}

For $L$ routed layers with $E$ experts each, sorting retained widths within each layer as $w_{l,(1)}\ge\cdots\ge w_{l,(E)}$ gives the layer-equally weighted coordinate coverage
\begin{equation}
G(k)=\frac{1}{L}\sum_{l=1}^{L}\frac{\sum_{i=1}^{k}w_{l,(i)}}{\sum_{e=1}^{E}w_{le}}.
\end{equation}
Selecting the top $k$ widths maximizes coordinate coverage at that expert count, giving the highest coverage that any whole-expert selection can reach.
\Cref{fig:retained-coordinate-coverage} measures how much of the neuron selection whole experts can cover at comparable budgets.
At matched 50\% budgets, coverage is 76.16--88.58\% for RAMPr-L and 67.65--83.08\% for RAMPr-D, leaving 11.42--32.35\% of retained coordinates outside the selected experts.
RAMPr-D, whose allocation is less polarized, falls further outside whole-expert coverage in every matched model--budget comparison.
Even at the nearest-integer 90\% expert budgets, where coverage reaches 94.06--97.80\%, whole experts gain only 4.06--7.49 percentage points over a uniform subset.
Concentration thus explains why whole-expert selection can cover many retained coordinates, while the partial sets show what its coarser granularity misses.

\subsection{Estimated Active Parameters and Computation}
\label{app:estimated-active}

Retained coordinates determine storage, but their contribution to per-token computation also depends on how often their experts are dispatched.
For hidden dimension $h_l$ and retained expert width $w_{le}$, the bias-free gate/up/down projections have logical routed storage
\begin{equation}
P_{\mathrm{routed,stored}}=\sum_{l,e}3h_lw_{le}.
\label{eq:routed-stored}
\end{equation}
Let $C_{dle}$ count actual original-model dispatches to expert $e$ in layer $l$ on domain $d$, and let $T_{dl}$ count valid input tokens, not token--expert edges.
Weighting retained coordinates by these dispatch counts gives the fixed-routing estimate
\begin{equation}
\bar P_{\mathrm{active},d}=\sum_l3h_l\frac{\sum_e C_{dle}w_{le}}{T_{dl}},
\qquad
\bar P_{\mathrm{active}}=\frac{1}{9}\sum_d\bar P_{\mathrm{active},d}.
\label{eq:routed-active}
\end{equation}
Dispatch counts, rather than gate probabilities, determine matrix dimensions; the recorded routes satisfy $\sum_e C_{dle}=K_lT_{dl}$.
Active parameters here mean routed-FFN weights invoked per token, not nonzero activations, the batch-wide union of weights, or full-model active parameters.
Shared experts, dense FFNs, attention, routers, embeddings, output heads and auxiliary MTP branches are outside this active count.

The probe uses 64 validation documents from each of nine domains: chat-gen, chat-mcq, code-gen, math-gen, math-mcq, med-gen, med-mcq, zh-gen and zh-mcq.
Models share document IDs and order, with native tokenization, raw text and a 2048-token cap; per-domain token-normalized means are averaged equally.
This probe is separate from mask calibration: \cref{tab:active-rampr-n64} covers all 36 n64 RAMPr configurations, \cref{tab:active-rampr-n128} lists four n128 supplements, and \cref{tab:active-fisher} lists 24 task-conditioned n128 Fisher masks.
All RAMPr active values apply fixed retained widths to original-model dispatch offline.

\begin{figure}[!hbp]
\centering
\includegraphics[width=\textwidth]{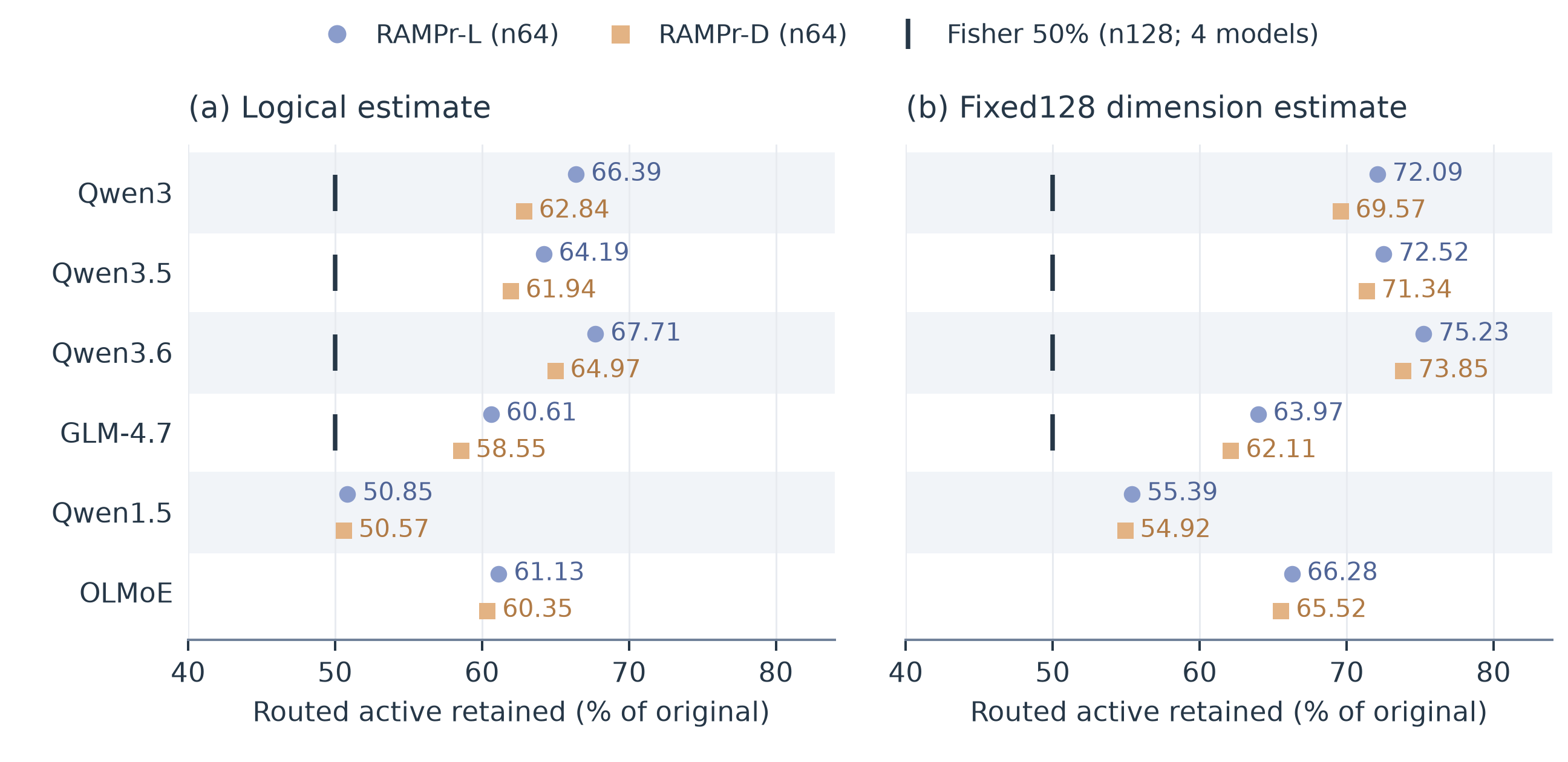}
\caption{Routed-FFN active retention at 50\% neuron retention: (a) logical and (b) fixed128 execution-dimension estimates for RAMPr-L/D with n64 calibration.
Both panels reuse original-model dispatch on the nine-domain validation probe; neither reports post-pruning active measurements.
Fisher's 50\% reference is shown only for the four models covered by the available task-conditioned n128 masks, not for Qwen1.5 or OLMoE; its calibration differs from RAMPr's.
Axes show retained active percentages relative to each original model (100\%), with a restricted display range.
REAP is not plotted across six models; its two inspected configurations are reported in \cref{tab:active-reap-budget}.}
\label{fig:estimated-active}
\end{figure}

Counting a multiply and an add separately gives the matrix-multiply estimate $\bar F_{\mathrm{FFN}}\simeq2\bar P_{\mathrm{active}}$.
It excludes nonlinearities, elementwise operations, routing, data movement, communication and other model components.
For the inspected fixed128 kernel, replacing $w_{le}$ by $128\lceil w_{le}/128\rceil$ (zero when $w_{le}=0$) gives an execution-dimension estimate.
At 50\% neuron retention, logical active retention is 50.57--67.71\%, versus 54.92--75.23\% with this alignment (\cref{fig:estimated-active}).
Keeping dense shapes and merely zeroing weights does not reduce dense matrix-multiply dimensions.

Fisher rule-E fixes every expert at 50\% of its original width; all 24 available masks therefore have exactly 50\% logical active retention, also 50\% under fixed128 because their retained widths are multiples of 128.
RAMPr instead selects neurons across expert boundaries according to domain-wise importance, without imposing the same retention quota on every expert.
Its higher usage-weighted retention is consistent with preserving more neurons in frequently dispatched experts, while Fisher's fixed quota yields lower active counts in these comparisons.
Quality scores are not paired with these estimates here.

For the inspected native no-capacity-drop REAP configurations only, Qwen3 retains 64/128 experts with $K=8$ and width 768, while GLM-4.7 retains 48/64 with $K=4$ and width 1536.
Their unchanged $K$ and full expert widths imply 100\% of original routed active parameters, despite lower expert storage.
\Cref{tab:active-reap-budget} compares backbone parameter budgets, excluding auxiliary MTP tensors and non-parameter buffers.
For GLM, this removes 1,278,094,656 auxiliary elements and 2,944 router-buffer elements from RAMPr, and 2,208 router-buffer elements from REAP: logical RAMPr is 22.997624B versus REAP's 22.996116B, a $+0.0066\%$ difference.
Qwen3 is 16.036608B versus 16.030317B ($+0.0392\%$); fixed128 padding raises RAMPr's storage separately.
Parameter counts exclude added rank/table buffers and are not GPU-memory estimates.

\begin{table}[!hbp]
\caption{All 36 RAMPr n64 configurations. All active values use original-model dispatch on the nine-domain validation probe. Stored counts concern logically sliced routed weights only; $A$ is retained active relative to the original model, $\bar P$ is weights invoked per token, and $\bar F$ is matrix-multiply FLOPs/token. Subscripts log and 128 distinguish logical and fixed128 execution-dimension estimates. B and G denote $10^9$. Keep is the nominal routed-neuron percentage; native integer rounding changes the 90\% routed-storage fraction by at most 0.0007 percentage points.}
\label{tab:active-rampr-n64}
\centering\small
\setlength{\tabcolsep}{3pt}
\renewcommand{\arraystretch}{1.04}
\rowcolors{2}{tabrow}{}
\begin{tabularx}{\textwidth}{llRRRRRRR}
\toprule
\rowcolor{tabhead}Model & Rule & Keep (\%) & $P_{\rm stored}$ (B) & $A_{\rm log}$ (\%) & $A_{128}$ (\%) & $\bar P_{\rm log}$ (B) & $\bar F_{\rm log}$ (G) & $\bar F_{128}$ (G) \\
\midrule
Qwen3 & L & 50 & 14.4955 & 66.39 & 72.09 & 1.2030 & 2.4060 & 2.6123 \\
Qwen3 & L & 75 & 21.7433 & 88.02 & 90.42 & 1.5949 & 3.1898 & 3.2768 \\
Qwen3 & L & 90 & 26.0920 & 96.55 & 97.38 & 1.7495 & 3.4989 & 3.5290 \\
Qwen3 & D & 50 & 14.4955 & 62.84 & 69.57 & 1.1386 & 2.2771 & 2.5212 \\
Qwen3 & D & 75 & 21.7433 & 85.16 & 89.08 & 1.5431 & 3.0861 & 3.2282 \\
Qwen3 & D & 90 & 26.0920 & 95.46 & 96.79 & 1.7297 & 3.4595 & 3.5077 \\
\addlinespace[2pt]
Qwen3.5 & L & 50 & 16.1061 & 64.19 & 72.52 & 0.6461 & 1.2923 & 1.4599 \\
Qwen3.5 & L & 75 & 24.1592 & 85.05 & 89.08 & 0.8561 & 1.7122 & 1.7934 \\
Qwen3.5 & L & 90 & 28.9911 & 95.38 & 96.95 & 0.9601 & 1.9202 & 1.9519 \\
Qwen3.5 & D & 50 & 16.1061 & 61.94 & 71.34 & 0.6235 & 1.2469 & 1.4362 \\
Qwen3.5 & D & 75 & 24.1592 & 83.25 & 87.97 & 0.8380 & 1.6760 & 1.7711 \\
Qwen3.5 & D & 90 & 28.9911 & 94.32 & 96.27 & 0.9494 & 1.8988 & 1.9382 \\
\addlinespace[2pt]
Qwen3.6 & L & 50 & 16.1061 & 67.71 & 75.23 & 0.6816 & 1.3632 & 1.5146 \\
Qwen3.6 & L & 75 & 24.1592 & 87.14 & 90.63 & 0.8772 & 1.7544 & 1.8247 \\
Qwen3.6 & L & 90 & 28.9911 & 96.02 & 97.32 & 0.9666 & 1.9332 & 1.9593 \\
Qwen3.6 & D & 50 & 16.1061 & 64.97 & 73.85 & 0.6540 & 1.3080 & 1.4867 \\
Qwen3.6 & D & 75 & 24.1592 & 85.55 & 89.75 & 0.8612 & 1.7224 & 1.8068 \\
Qwen3.6 & D & 90 & 28.9911 & 95.23 & 96.87 & 0.9586 & 1.9172 & 1.9503 \\
\addlinespace[2pt]
GLM-4.7 & L & 50 & 13.8915 & 60.61 & 63.97 & 1.0525 & 2.1051 & 2.2215 \\
GLM-4.7 & L & 75 & 20.8373 & 83.39 & 85.17 & 1.4481 & 2.8962 & 2.9579 \\
GLM-4.7 & L & 90 & 25.0049 & 94.58 & 95.25 & 1.6424 & 3.2847 & 3.3080 \\
GLM-4.7 & D & 50 & 13.8915 & 58.55 & 62.11 & 1.0167 & 2.0333 & 2.1570 \\
GLM-4.7 & D & 75 & 20.8373 & 81.35 & 83.87 & 1.4126 & 2.8251 & 2.9128 \\
GLM-4.7 & D & 90 & 25.0049 & 93.44 & 94.46 & 1.6226 & 3.2451 & 3.2806 \\
\addlinespace[2pt]
Qwen1.5 & L & 50 & 6.2285 & 50.85 & 55.39 & 0.4223 & 0.8445 & 0.9200 \\
Qwen1.5 & L & 75 & 9.3428 & 75.72 & 79.27 & 0.6288 & 1.2576 & 1.3166 \\
Qwen1.5 & L & 90 & 11.2114 & 90.44 & 92.27 & 0.7511 & 1.5022 & 1.5325 \\
Qwen1.5 & D & 50 & 6.2285 & 50.57 & 54.92 & 0.4200 & 0.8400 & 0.9121 \\
Qwen1.5 & D & 75 & 9.3428 & 75.40 & 79.63 & 0.6261 & 1.2523 & 1.3226 \\
Qwen1.5 & D & 90 & 11.2114 & 90.21 & 93.37 & 0.7492 & 1.4984 & 1.5508 \\
\addlinespace[2pt]
OLMoE & L & 50 & 3.2212 & 61.13 & 66.28 & 0.4923 & 0.9846 & 1.0675 \\
OLMoE & L & 75 & 4.8318 & 83.51 & 86.65 & 0.6725 & 1.3449 & 1.3956 \\
OLMoE & L & 90 & 5.7982 & 94.54 & 95.62 & 0.7613 & 1.5226 & 1.5400 \\
OLMoE & D & 50 & 3.2212 & 60.35 & 65.52 & 0.4860 & 0.9720 & 1.0553 \\
OLMoE & D & 75 & 4.8318 & 82.36 & 86.29 & 0.6633 & 1.3265 & 1.3898 \\
OLMoE & D & 90 & 5.7982 & 93.85 & 95.73 & 0.7558 & 1.5116 & 1.5419 \\
\bottomrule
\end{tabularx}
\end{table}

\begin{table}[!hbp]
\caption{Four supplementary RAMPr n128 configurations, separate from the n64 grid. All active values use original-model dispatch on the nine-domain validation probe. Stored counts concern logically sliced routed weights only; $A$ is retained active relative to the original model, $\bar P$ is weights invoked per token, and $\bar F$ is matrix-multiply FLOPs/token. Subscripts log and 128 distinguish logical and fixed128 execution-dimension estimates. B and G denote $10^9$.}
\label{tab:active-rampr-n128}
\centering\small
\setlength{\tabcolsep}{3pt}
\renewcommand{\arraystretch}{1.04}
\rowcolors{2}{tabrow}{}
\begin{tabularx}{\textwidth}{llRRRRRRR}
\toprule
\rowcolor{tabhead}Model & Rule & Keep (\%) & $P_{\rm stored}$ (B) & $A_{\rm log}$ (\%) & $A_{128}$ (\%) & $\bar P_{\rm log}$ (B) & $\bar F_{\rm log}$ (G) & $\bar F_{128}$ (G) \\
\midrule
Qwen3 & L & 50 & 14.4955 & 66.16 & 71.94 & 1.1988 & 2.3976 & 2.6072 \\
Qwen3 & D & 50 & 14.4955 & 62.77 & 69.47 & 1.1374 & 2.2747 & 2.5176 \\
\addlinespace[2pt]
GLM-4.7 & L & 75 & 20.8373 & 83.28 & 85.02 & 1.4460 & 2.8921 & 2.9528 \\
GLM-4.7 & D & 75 & 20.8373 & 81.40 & 83.87 & 1.4134 & 2.8269 & 2.9128 \\
\bottomrule
\end{tabularx}
\end{table}

\begin{table}[!hbp]
\caption{All 24 available Fisher rule-E masks. Each uses 128 task-conditioned calibration texts; retained width $w$ is exactly half of each expert's original width, unlike RAMPr's cross-expert allocation. Both logical and fixed128 routed-active retention are 50\%; retained widths are multiples of 128, so the two matrix-multiply estimates coincide. Columns use the units and original-routing probe of \cref{tab:active-rampr-n64}.}
\label{tab:active-fisher}
\centering\small
\setlength{\tabcolsep}{3pt}
\renewcommand{\arraystretch}{1.04}
\rowcolors{2}{tabrow}{}
\begin{tabularx}{\textwidth}{llRRRRRR}
\toprule
\rowcolor{tabhead}Model & Calibration task & $w$ & $P_{\rm stored}$ (B) & $A_{\rm log}$ (\%) & $A_{128}$ (\%) & $\bar P_{\rm log}$ (B) & $\bar F_{\rm log}=\bar F_{128}$ (G) \\
\midrule
Qwen3 & C-Eval (valid) & 384 & 14.4955 & 50.00 & 50.00 & 0.9060 & 1.8119 \\
Qwen3 & CMMLU & 384 & 14.4955 & 50.00 & 50.00 & 0.9060 & 1.8119 \\
Qwen3 & GSM8K & 384 & 14.4955 & 50.00 & 50.00 & 0.9060 & 1.8119 \\
Qwen3 & HellaSwag & 384 & 14.4955 & 50.00 & 50.00 & 0.9060 & 1.8119 \\
Qwen3 & MBPP & 384 & 14.4955 & 50.00 & 50.00 & 0.9060 & 1.8119 \\
Qwen3 & \texttt{minerva\_math} & 384 & 14.4955 & 50.00 & 50.00 & 0.9060 & 1.8119 \\
Qwen3 & MMLU & 384 & 14.4955 & 50.00 & 50.00 & 0.9060 & 1.8119 \\
Qwen3 & WinoGrande & 384 & 14.4955 & 50.00 & 50.00 & 0.9060 & 1.8119 \\
\addlinespace[2pt]
Qwen3.5 & C-Eval (valid) & 256 & 16.1061 & 50.00 & 50.00 & 0.5033 & 1.0066 \\
Qwen3.5 & CMMLU & 256 & 16.1061 & 50.00 & 50.00 & 0.5033 & 1.0066 \\
Qwen3.5 & GSM8K & 256 & 16.1061 & 50.00 & 50.00 & 0.5033 & 1.0066 \\
Qwen3.5 & MBPP & 256 & 16.1061 & 50.00 & 50.00 & 0.5033 & 1.0066 \\
Qwen3.5 & MMLU & 256 & 16.1061 & 50.00 & 50.00 & 0.5033 & 1.0066 \\
\addlinespace[2pt]
Qwen3.6 & C-Eval (valid) & 256 & 16.1061 & 50.00 & 50.00 & 0.5033 & 1.0066 \\
Qwen3.6 & CMMLU & 256 & 16.1061 & 50.00 & 50.00 & 0.5033 & 1.0066 \\
Qwen3.6 & GSM8K & 256 & 16.1061 & 50.00 & 50.00 & 0.5033 & 1.0066 \\
Qwen3.6 & HellaSwag & 256 & 16.1061 & 50.00 & 50.00 & 0.5033 & 1.0066 \\
Qwen3.6 & MBPP & 256 & 16.1061 & 50.00 & 50.00 & 0.5033 & 1.0066 \\
Qwen3.6 & \texttt{minerva\_math} & 256 & 16.1061 & 50.00 & 50.00 & 0.5033 & 1.0066 \\
Qwen3.6 & MMLU & 256 & 16.1061 & 50.00 & 50.00 & 0.5033 & 1.0066 \\
Qwen3.6 & WinoGrande & 256 & 16.1061 & 50.00 & 50.00 & 0.5033 & 1.0066 \\
\addlinespace[2pt]
GLM-4.7 & GSM8K & 768 & 13.8915 & 50.00 & 50.00 & 0.8682 & 1.7364 \\
GLM-4.7 & MBPP & 768 & 13.8915 & 50.00 & 50.00 & 0.8682 & 1.7364 \\
GLM-4.7 & \texttt{minerva\_math} & 768 & 13.8915 & 50.00 & 50.00 & 0.8682 & 1.7364 \\
\bottomrule
\end{tabularx}
\end{table}

\begin{table}[!hbp]
\caption{Backbone-parameter reconciliation with the two inspected REAP checkpoints. Qwen3 uses 50\% neuron/expert retention and GLM-4.7 uses 75\%; RAMPr rules L/D are separate. $P_{\rm REAP}$, $P_{\rm log}$ and $P_{128}$ are backbone parameter counts in billions. $\Delta P$ is the logical RAMPr budget difference relative to REAP. $A_{\rm log}$ and $A_{128}$ are original-routing RAMPr active percentages relative to REAP's fixed-$K$, full-width routed structure (100\% of its original model). Auxiliary MTP tensors and non-parameter buffers are excluded.}
\label{tab:active-reap-budget}
\centering\small
\setlength{\tabcolsep}{3pt}
\renewcommand{\arraystretch}{1.04}
\rowcolors{2}{tabrow}{}
\begin{tabularx}{\textwidth}{lRLRRRRRR}
\toprule
\rowcolor{tabhead}Model & $n$ & Rule & $P_{\rm REAP}$ & $P_{\rm log}$ & $\Delta P$ (\%) & $P_{128}$ & $A_{\rm log}$ (\%) & $A_{128}$ (\%) \\
\midrule
Qwen3 & 64 & L & 16.030317 & 16.036608 & +0.0392 & 18.176489 & 66.39 & 72.09 \\
Qwen3 & 64 & D & 16.030317 & 16.036608 & +0.0392 & 18.275580 & 62.84 & 69.57 \\
Qwen3 & 128 & L & 16.030317 & 16.036608 & +0.0392 & 18.206374 & 66.16 & 71.94 \\
Qwen3 & 128 & D & 16.030317 & 16.036608 & +0.0392 & 18.276366 & 62.77 & 69.47 \\
\addlinespace[2pt]
GLM-4.7 & 64 & L & 22.996116 & 22.997624 & +0.0066 & 23.642498 & 83.39 & 85.17 \\
GLM-4.7 & 64 & D & 22.996116 & 22.997624 & +0.0066 & 23.815513 & 81.35 & 83.87 \\
GLM-4.7 & 128 & L & 22.996116 & 22.997624 & +0.0066 & 23.627556 & 83.28 & 85.02 \\
GLM-4.7 & 128 & D & 22.996116 & 22.997624 & +0.0066 & 23.805289 & 81.40 & 83.87 \\
\bottomrule
\end{tabularx}
\end{table}

\subsection{Task Scores under Physical Pruning and Zero Masking}
\label{app:pruning-pairs}
A selected neuron set can be implemented by zeroing discarded weights or removing their rows and columns.
The comparison in \cref{sec:pruning} holds the calibration mask and scoring protocol fixed to separate implementation differences from selection effects.
Question-level outcomes further distinguish score differences from disagreements on individual answers.

The n64 paired study evaluates 36 pruning configurations on five tasks.
HumanEval uses the \texttt{code-answer-v3-prompt-context} scoring protocol for 36 physical models and their 36 zero-masked counterparts, together with two Base configurations.
Each is evaluated on all 164 questions.

\begin{figure}[!htbp]
\centering
\includegraphics[width=\textwidth]{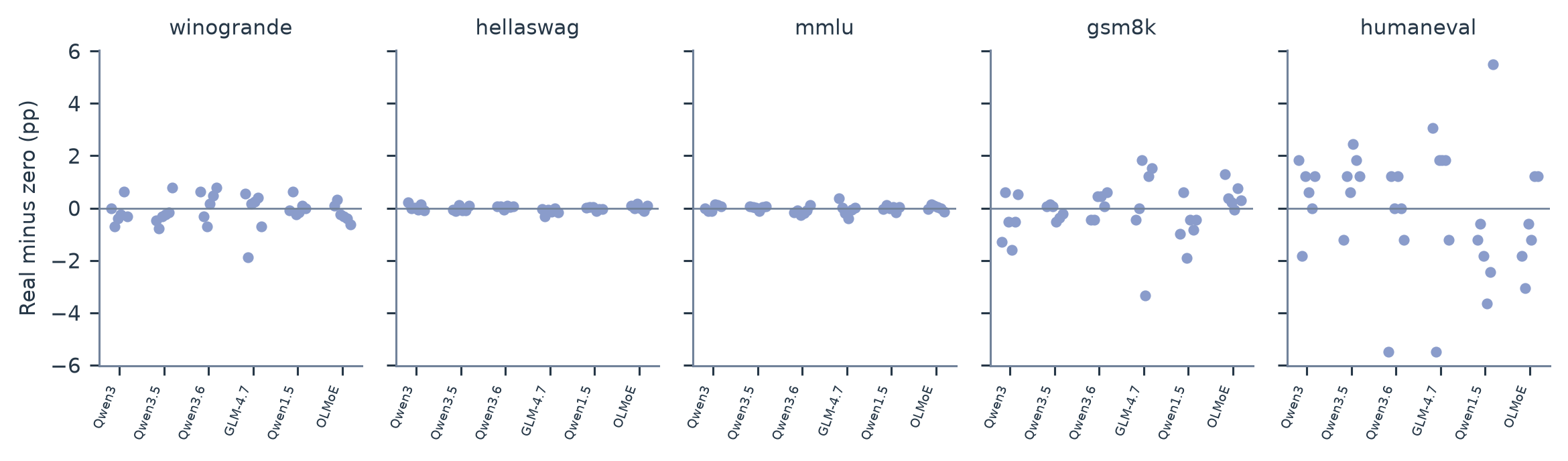}
\caption{Real-minus-zero task-score differences for all 180 n64 comparisons.
Each model has six method--retention settings per task.
HumanEval uses source-matched v3 scores, and the other tasks retain their recorded metrics.
Horizontal offsets separate configurations without implying another quantitative axis.}
\label{fig:pruning-pairs}
\end{figure}

The mean signed differences are $-0.094$, $-0.009$, $-0.020$, $-0.093$ and $-0.051$ percentage points for WinoGrande, HellaSwag, MMLU, GSM8K and HumanEval, respectively.
HumanEval has the largest mean absolute difference, approximately 1.78 points.
For Qwen3.6 RAMPr-L at 50\%, physical and zero implementations score 125/164 and 134/164.
There are 15 questions won only by the physical model and 24 won only by the zero model.
The 39 differing decisions show why the nine-question net gap alone does not capture answer-level agreement.

The n128 quality and n64 implementation studies use different calibration masks and generated-answer sets.
For example, Qwen3 RAMPr-L at 75\% scores 79.8780\% on HumanEval with n128 calibration and 73.7805\% in the n64 zero-mask study.
The supplementary reference-model comparison uses \texttt{code-answer-v4-leading-\allowbreak space-local-import} for four reference models and 18 RAMPr-L zero configurations.
The latter 18 scores equal the v3 scores.
All physical/zero HumanEval comparisons use the common v3 protocol.

\subsection{Efficiency Measurement and Layouts}
\label{app:pruning-efficiency}
Measuring resident memory and inference time connects the neuron selection in \cref{sec:pruning} to practical resource use.
A common backend and fixed layout allow prefill and decode costs to be compared without choosing a different implementation for each workload.

All speedups in the six-model study compare the full and physically pruned models in the same modified Transformers backend, not the vLLM backend used for main-table quality evaluation.
The study uses NVIDIA H20 with BF16 and SDPA, running PyTorch 2.13.0, Transformers 5.17.0 and CUDA 13.3.
Timing excludes model loading and weight conversion.
The original-model baseline is measured before and after each layout run, and its throughput differs by more than 5\% between the two measurements in 17 of 144 scenarios, by at most about 11.88\%.
Native expandable-segment allocation is enabled.
Thirty-six runs cover each model at three retention rates and batch sizes 1 and 32.
Each run measures prompts of 512 and 2048 tokens, with decode fixed to 64 steps.
The layout comparison includes Full and native same-mask zero baselines alongside fixed and adaptive layouts.
The main display uses fixed width 128 rather than selecting the fastest layout separately for each cell.

\begin{figure}[!htbp]
\centering
\includegraphics[width=\textwidth]{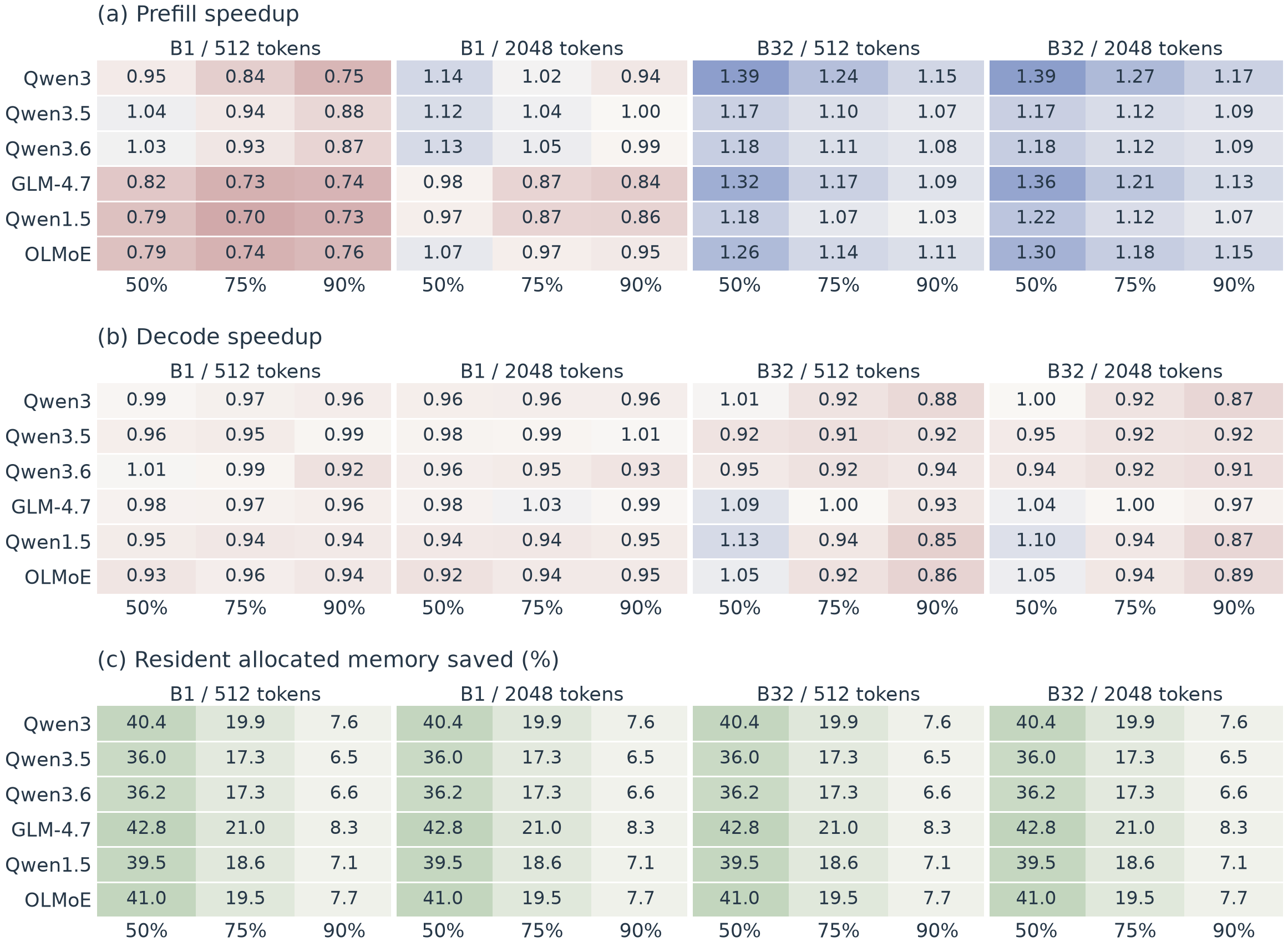}
\caption{Full fixed128 efficiency grid for RAMPr-L with n64 calibration.
Every panel reports all six models at 50\%, 75\% and 90\% neuron retention under four batch--input settings, giving 216 displayed values.
Speedups compare the pruned model with its same-GPU Full baseline and share a color scale centered at one.
Memory reductions use a separate 0--50\% scale.
Numbers are rounded only for display.
Loading and layout conversion are outside inference timing.}
\label{fig:pruning-efficiency-full}
\end{figure}

The complete fixed128 results in \cref{fig:pruning-efficiency-full} cover all retention rates and workloads.

Full and zero each have five timed measurements before and five after the layout runs.
Each pruned layout has six blocks of three repetitions after warmup.
For a fixed model, GPU and workload, let $F_0,F_1$ be the Full time samples and $C$ the pruned samples.
The reported Full-relative speedup is
\begin{equation}
 S_{\mathrm{Full}}=\frac{\operatorname{median}(F_0\cup F_1)}{\operatorname{median}(C)}.
\end{equation}
The prefill and decode panels in \cref{fig:pruning-efficiency} convert this speedup to a percentage change as $100(S_{\mathrm{Full}}-1)$, while resident-memory reductions are already percentages.
For comparison with fixed128, each block gives the ratio of its fixed128 and candidate median times; the final estimate is the median of the six block ratios.
Throughput is summarized separately from per-trial token counts divided by time, whereas reported speedup uses the time samples directly.
Prefill counts input tokens across the batch, while decode counts the fixed output workload.
Inference timing starts after loading, input transfer and layout preparation.

Resident memory is sampled after loading Full, applying the zero mask, or installing each pruned layout, before that model's suite warmup.
The sampler runs garbage collection, releases unused allocator cache, synchronizes the device, and reads \texttt{memory\_allocated()} and \texttt{memory\_reserved()} in GiB.
These values describe active PyTorch allocations and reserved allocator segments in the process.
The current suite input has not yet been created, and preceding trial outputs and KV caches have been deleted.
Each layout uses its last sampled resident snapshot; Full and zero use their before-run snapshots, with after-run values recorded separately.
The memory reduction is $100(1-M_{\mathrm{pruned}}/M_{\mathrm{Full}})$ using resident allocated memory.

Peak allocation statistics are reset after creating the input and attention mask but before prefill.
Their maxima are read after the trial and summarized by the largest non-warmup value.
For decode rows, the peak-memory interval includes prefill and the 64 decode steps, while the time field measures only those decode steps.
All Full, zero and pruned runs use the same PyTorch/Transformers environment.
Pruned models replace native expert modules with the custom ragged/hybrid implementation.

CUDA step percentiles are not request-tail latency.
Energy is measured through cumulative GPU energy without idle subtraction and is not whole-machine energy.
Missing energy and unavailable whole-run tail estimates remain NA.

At batch 32 and 2048 input tokens, the fixed128 prefill speedup ranges are approximately 1.17--1.39 at 50\%, 1.12--1.27 at 75\%, and 1.07--1.17 at 90\% retention.
Corresponding resident-memory reductions are approximately 36--43\%, 17--21\%, and 6--8\%.
Decode is roughly unchanged at 50\% retention, with a geometric-mean speedup of 1.01, and slows at higher retention, where fewer neurons are removed.

\subsection{Calibration Ablations}
\label{app:pruning-calibration}
Neuron importance is estimated from calibration text, so both sample count and domain composition can affect selection.
The ablations in \cref{sec:pruning-ablation} vary these inputs to examine their interaction with score aggregation.

The calibration-size study covers four models with per-domain counts 1, 2, 4, 8, 16, 32 and 64, yielding 2856 seed-level results and 2184 task means.
The n64 HumanEval results use the v3 extraction protocol on the corresponding generated answers; scoring protocols for other calibration sizes follow their individual task records.
Larger calibration sets improve the strongest-compression RAMPr-L endpoint scores across the evaluated models, while intermediate sizes and other configurations vary less regularly.

\begin{figure}[!htbp]
\centering
\includegraphics[width=\textwidth]{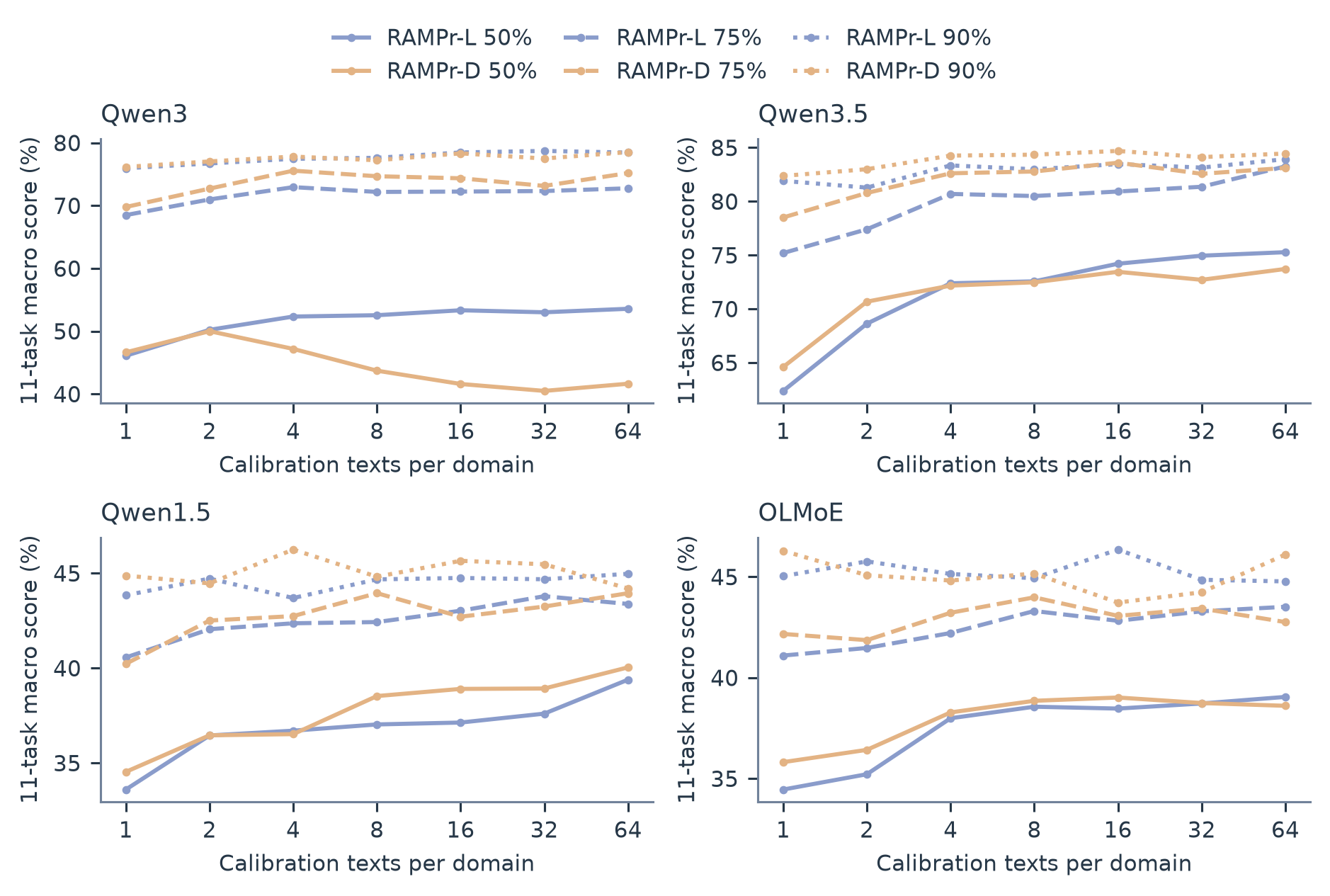}
\caption{Eleven-task raw macro scores over calibration-set size.
Blue denotes RAMPr-L and orange RAMPr-D.
Solid, dashed and dotted lines show 50\%, 75\% and 90\% neuron retention.
The n64 HumanEval scores use the v3 protocol.
Curves show raw task-score averages.}
\label{fig:pruning-size}
\end{figure}

The Qwen3 composition comparison changes calibration from eleven mixed buckets with 1408 texts to five generation buckets with 640 texts.
It preserves the scoring rule and 50\% retention but changes both domain composition and sample count.
Generation-focused calibration thus recovers much of Qwen3's generation ability without changing the method and with fewer than half the texts, raising RAMPr-L overall retention from 61.04\% to 70.12\% and RAMPr-D from 44.47\% to 50.56\%.
Multiple-choice retention decreases in exchange, from 90.69\% to 80.00\% for RAMPr-L.

\begin{figure}[!htbp]
\centering
\includegraphics[width=\textwidth]{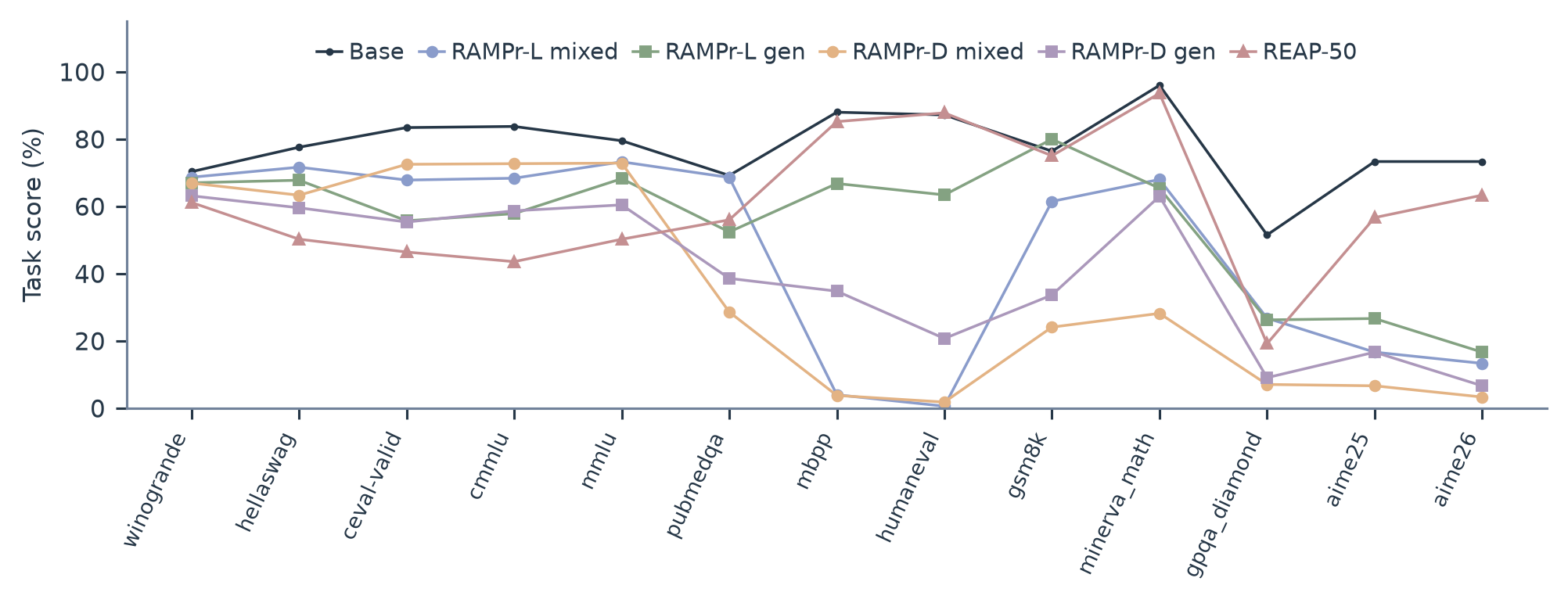}
\caption{All thirteen task scores in the Qwen3 calibration-composition comparison.
Mixed and generation-only calibration contain different total numbers of texts.
Both RAMPr variants and the available full-model and REAP results are shown.
All task scores correspond to the same calibration-composition experiment.}
\label{fig:pruning-composition}
\end{figure}

\subsection{Execution Ablations}
Holding selected neurons fixed separates score-based selection from execution-layout effects in \cref{sec:pruning-ablation}.

Layout ablations preserve effective masks while changing weight packing.
Fixed blocks can pad expert widths; smaller blocks reduce some padding but can exceed the grouped backend's 1023-group limit, requiring segmented calls whose overhead remains in the timing.
Structural adaptive layouts optimize a proxy such as block count, while timing-based choices use a local replay estimate.
Historical Qwen3 runs use their own cohorts and fixed128 denominators and are not combined with the newer six-model Full baselines.

\subsection{Complete Task Tables}
The complete task tables allow the capability-retention comparisons in \cref{sec:pruning} to be read alongside individual outcomes and the settings used to obtain them.

All n128 main-table rows are included, with values rounded only for display and unavailable task results left marked.
\begingroup
\renewenvironment{table*}[1][]{\par\addvspace{10pt}\noindent\begin{minipage}{\textwidth}\captionsetup{type=table}}{\end{minipage}\par}
\begin{table*}[t]
\caption{Qwen3 main-table scores. Missing entries remain NA. Scores are percentages under the recorded task protocols.}
\centering
\setlength{\tabcolsep}{2pt}
\begin{minipage}[t]{0.485\linewidth}
\vspace{0pt}
\centering\scriptsize
\rowcolors{2}{tabrow}{}
\begin{tabularx}{\linewidth}{Lrrrrrrr}
\toprule
\rowcolor{tabhead}Method & Wino & Hella & CEval & cmmlu & mmlu & PubMed & mbpp \\
\midrule
base & 70.32 & 77.59 & 83.43 & 83.77 & 79.53 & 69.20 & 88.00 \\
fisher-gsm8k & 60.62 & 47.39 & 45.02 & 43.39 & 49.71 & 20.00 & 0.00 \\
fisher-mbpp & 61.56 & 49.29 & 49.03 & 48.20 & 52.49 & 57.60 & 1.40 \\
fisher-winogrande & 64.01 & NA & NA & NA & NA & NA & NA \\
fisher-hellaswag & NA & 60.85 & NA & NA & NA & NA & NA \\
fisher-ceval-valid & NA & NA & 65.82 & NA & NA & NA & NA \\
fisher-cmmlu & NA & NA & NA & 63.98 & NA & NA & NA \\
fisher-mmlu & NA & NA & NA & NA & 64.77 & NA & NA \\
fisher-minerva-math & NA & NA & NA & NA & NA & NA & NA \\
RAMPr-L 50 & 68.67 & 71.66 & 67.83 & 68.37 & 73.25 & 68.60 & 4.00 \\
RAMPr-D 50 & 66.93 & 63.32 & 72.51 & 72.70 & 72.87 & 28.60 & 3.80 \\
RAMPr-L 75 & 70.96 & 76.81 & 79.27 & 80.63 & 78.42 & 66.60 & 79.00 \\
RAMPr-D 75 & 69.85 & 75.05 & 79.94 & 80.38 & 77.83 & 64.20 & 80.60 \\
RAMPr-L 90 & 71.19 & 77.32 & 82.91 & 83.34 & 79.63 & 68.00 & 87.20 \\
RAMPr-D 90 & 71.59 & 77.40 & 82.99 & 83.34 & 79.13 & 68.60 & 87.20 \\
reap50 & 61.17 & 50.25 & 46.43 & 43.57 & 50.24 & 56.00 & 85.20 \\
\bottomrule
\end{tabularx}
\end{minipage}\hfill
\begin{minipage}[t]{0.485\linewidth}
\vspace{0pt}
\centering\scriptsize
\rowcolors{2}{tabrow}{}
\begin{tabularx}{\linewidth}{Lrrrrrr}
\toprule
\rowcolor{tabhead}Method & humaneval & gsm8k & MATH & GPQA & aime25 & aime26 \\
\midrule
base & 87.20 & 76.42 & 96.00 & 51.52 & 73.33 & 73.33 \\
fisher-gsm8k & 0.00 & 45.34 & 6.80 & 3.54 & 0.00 & 0.00 \\
fisher-mbpp & 0.00 & 2.27 & 0.20 & 1.01 & 0.00 & 0.00 \\
fisher-winogrande & NA & NA & NA & NA & NA & NA \\
fisher-hellaswag & NA & NA & NA & NA & NA & NA \\
fisher-ceval-valid & NA & NA & NA & NA & NA & NA \\
fisher-cmmlu & NA & NA & NA & NA & NA & NA \\
fisher-mmlu & NA & NA & NA & NA & NA & NA \\
fisher-minerva-math & NA & NA & 11.20 & NA & NA & NA \\
RAMPr-L 50 & 0.61 & 61.49 & 68.00 & 26.77 & 16.67 & 13.33 \\
RAMPr-D 50 & 1.83 & 24.11 & 28.20 & 7.07 & 6.67 & 3.33 \\
RAMPr-L 75 & 79.88 & 61.11 & 95.00 & 43.94 & 66.67 & 60.00 \\
RAMPr-D 75 & 84.15 & 84.38 & 95.00 & 42.42 & 53.33 & 66.67 \\
RAMPr-L 90 & 89.02 & 73.16 & 95.00 & 44.44 & 70.00 & 66.67 \\
RAMPr-D 90 & 89.63 & 76.42 & 95.80 & 52.02 & 66.67 & 66.67 \\
reap50 & 87.80 & 75.06 & 93.60 & 19.19 & 56.67 & 63.33 \\
\bottomrule
\end{tabularx}
\end{minipage}
\end{table*}
\begin{table*}[t]
\caption{Qwen3.5 main-table scores. Missing entries remain NA. Scores are percentages under the recorded task protocols.}
\centering
\setlength{\tabcolsep}{2pt}
\begin{minipage}[t]{0.485\linewidth}
\vspace{0pt}
\centering\scriptsize
\rowcolors{2}{tabrow}{}
\begin{tabularx}{\linewidth}{Lrrrrrrr}
\toprule
\rowcolor{tabhead}Method & Wino & Hella & CEval & cmmlu & mmlu & PubMed & mbpp \\
\midrule
base & 75.06 & 82.64 & 84.40 & 85.21 & 84.96 & 65.40 & 86.60 \\
fisher-gsm8k & 67.32 & 61.67 & 61.96 & 62.23 & 64.11 & 70.80 & 11.20 \\
fisher-mbpp & 67.48 & 59.65 & 61.96 & 64.15 & 66.30 & 70.80 & 52.40 \\
fisher-winogrande & 71.19 & NA & NA & NA & NA & NA & NA \\
fisher-hellaswag & NA & 70.03 & NA & NA & NA & NA & NA \\
fisher-ceval-valid & NA & NA & 72.59 & NA & NA & NA & NA \\
fisher-cmmlu & NA & NA & NA & 72.97 & NA & NA & NA \\
fisher-mmlu & NA & NA & NA & NA & 73.36 & NA & NA \\
fisher-minerva-math & NA & NA & NA & NA & NA & NA & NA \\
RAMPr-L 50 & 73.01 & 74.72 & 74.15 & 75.15 & 78.76 & 69.40 & 82.40 \\
RAMPr-D 50 & 73.32 & 71.19 & 77.27 & 76.50 & 78.97 & 62.80 & 79.00 \\
RAMPr-L 75 & 74.98 & 81.55 & 81.65 & 82.77 & 83.72 & 67.80 & 87.00 \\
RAMPr-D 75 & 74.59 & 81.24 & 82.84 & 83.13 & 83.86 & 68.80 & 89.20 \\
RAMPr-L 90 & 74.19 & 82.35 & 83.73 & 84.82 & 84.46 & 64.60 & 89.60 \\
RAMPr-D 90 & 74.03 & 82.32 & 83.58 & 84.69 & 84.52 & 66.00 & 89.20 \\
\bottomrule
\end{tabularx}
\end{minipage}\hfill
\begin{minipage}[t]{0.485\linewidth}
\vspace{0pt}
\centering\scriptsize
\rowcolors{2}{tabrow}{}
\begin{tabularx}{\linewidth}{Lrrrrrr}
\toprule
\rowcolor{tabhead}Method & humaneval & gsm8k & MATH & GPQA & aime25 & aime26 \\
\midrule
base & 98.17 & 97.12 & 96.80 & 80.81 & 90.00 & 90.00 \\
fisher-gsm8k & 12.80 & 81.12 & 39.40 & 14.65 & 0.00 & 0.00 \\
fisher-mbpp & 73.17 & 52.31 & 35.80 & 15.66 & 0.00 & 0.00 \\
fisher-winogrande & NA & NA & NA & NA & NA & NA \\
fisher-hellaswag & NA & NA & NA & NA & NA & NA \\
fisher-ceval-valid & NA & NA & NA & NA & NA & NA \\
fisher-cmmlu & NA & NA & NA & NA & NA & NA \\
fisher-mmlu & NA & NA & NA & NA & NA & NA \\
fisher-minerva-math & NA & NA & 72.40 & NA & NA & NA \\
RAMPr-L 50 & 94.51 & 96.66 & 82.20 & 36.36 & 43.33 & 43.33 \\
RAMPr-D 50 & 89.63 & 95.83 & 86.60 & 25.25 & 26.67 & 20.00 \\
RAMPr-L 75 & 95.12 & 96.59 & 95.20 & 58.59 & 83.33 & 93.33 \\
RAMPr-D 75 & 96.34 & 96.82 & 96.60 & 60.61 & 90.00 & 93.33 \\
RAMPr-L 90 & 95.12 & 96.82 & 97.60 & 71.21 & 83.33 & 93.33 \\
RAMPr-D 90 & 94.51 & 96.97 & 97.20 & 75.76 & 90.00 & 93.33 \\
\bottomrule
\end{tabularx}
\end{minipage}
\end{table*}
\begin{table*}[t]
\caption{Qwen3.6 main-table scores. Missing entries remain NA. Scores are percentages under the recorded task protocols.}
\centering
\setlength{\tabcolsep}{2pt}
\begin{minipage}[t]{0.485\linewidth}
\vspace{0pt}
\centering\scriptsize
\rowcolors{2}{tabrow}{}
\begin{tabularx}{\linewidth}{Lrrrrrrr}
\toprule
\rowcolor{tabhead}Method & Wino & Hella & CEval & cmmlu & mmlu & PubMed & mbpp \\
\midrule
base & 74.66 & 83.52 & 83.58 & 85.19 & 82.45 & 76.40 & 91.80 \\
fisher-gsm8k & 69.38 & 63.93 & 64.49 & 64.38 & 65.87 & 73.20 & 21.20 \\
fisher-mbpp & 66.46 & 60.50 & 65.16 & 65.77 & 66.73 & 74.80 & 61.00 \\
fisher-winogrande & 71.59 & NA & NA & NA & NA & NA & NA \\
fisher-hellaswag & NA & 71.69 & NA & NA & NA & NA & NA \\
fisher-ceval-valid & NA & NA & 72.88 & NA & NA & NA & NA \\
fisher-cmmlu & NA & NA & NA & 73.09 & NA & NA & NA \\
fisher-mmlu & NA & NA & NA & NA & 72.09 & NA & NA \\
fisher-minerva-math & NA & NA & NA & NA & NA & NA & NA \\
RAMPr-L 50 & 72.38 & 76.33 & 76.45 & 76.17 & 75.05 & 77.60 & 79.40 \\
RAMPr-D 50 & 71.67 & 73.24 & 77.19 & 76.82 & 76.21 & 76.60 & 81.00 \\
RAMPr-L 75 & 74.74 & 82.55 & 82.24 & 83.10 & 81.58 & 77.60 & 89.20 \\
RAMPr-D 75 & 74.03 & 81.95 & 82.76 & 83.17 & 80.69 & 77.40 & 90.00 \\
RAMPr-L 90 & 74.43 & 83.20 & 83.43 & 85.03 & 82.79 & 76.20 & 91.40 \\
RAMPr-D 90 & 74.43 & 83.51 & 83.80 & 85.11 & 82.46 & 76.60 & 90.60 \\
\bottomrule
\end{tabularx}
\end{minipage}\hfill
\begin{minipage}[t]{0.485\linewidth}
\vspace{0pt}
\centering\scriptsize
\rowcolors{2}{tabrow}{}
\begin{tabularx}{\linewidth}{Lrrrrrr}
\toprule
\rowcolor{tabhead}Method & humaneval & gsm8k & MATH & GPQA & aime25 & aime26 \\
\midrule
base & 96.34 & 96.66 & 97.40 & 57.07 & 93.33 & 96.67 \\
fisher-gsm8k & 15.24 & 87.34 & 33.00 & 10.10 & 0.00 & 0.00 \\
fisher-mbpp & 65.85 & 63.84 & 39.20 & 11.62 & 0.00 & 3.33 \\
fisher-winogrande & NA & NA & NA & NA & NA & NA \\
fisher-hellaswag & NA & NA & NA & NA & NA & NA \\
fisher-ceval-valid & NA & NA & NA & NA & NA & NA \\
fisher-cmmlu & NA & NA & NA & NA & NA & NA \\
fisher-mmlu & NA & NA & NA & NA & NA & NA \\
fisher-minerva-math & NA & NA & 45.00 & NA & NA & NA \\
RAMPr-L 50 & 87.20 & 96.29 & 85.40 & 25.25 & 60.00 & 53.33 \\
RAMPr-D 50 & 87.80 & 96.36 & 88.60 & 27.27 & 36.67 & 23.33 \\
RAMPr-L 75 & 96.95 & 96.44 & 96.40 & 54.04 & 86.67 & 86.67 \\
RAMPr-D 75 & 96.34 & 96.36 & 96.80 & 57.07 & 83.33 & 86.67 \\
RAMPr-L 90 & 97.56 & 96.66 & 96.80 & 53.54 & 90.00 & 90.00 \\
RAMPr-D 90 & 95.73 & 96.21 & 97.40 & 58.08 & 86.67 & 93.33 \\
\bottomrule
\end{tabularx}
\end{minipage}
\end{table*}
\begin{table*}[t]
\caption{GLM-4.7 main-table scores. Missing entries remain NA. Scores are percentages under the recorded task protocols.}
\centering
\setlength{\tabcolsep}{2pt}
\begin{minipage}[t]{0.485\linewidth}
\vspace{0pt}
\centering\scriptsize
\rowcolors{2}{tabrow}{}
\begin{tabularx}{\linewidth}{Lrrrrrrr}
\toprule
\rowcolor{tabhead}Method & Wino & Hella & CEval & cmmlu & mmlu & PubMed & mbpp \\
\midrule
base & 73.16 & 80.13 & 69.24 & 72.24 & 72.43 & 53.40 & 80.60 \\
fisher-gsm8k & 58.64 & 51.15 & 41.38 & 38.66 & 41.06 & 42.60 & 1.20 \\
fisher-mbpp & 56.20 & 50.05 & 42.72 & 41.13 & 39.10 & 53.40 & 10.80 \\
fisher-winogrande & 60.38 & NA & NA & NA & NA & NA & NA \\
fisher-hellaswag & NA & 57.88 & NA & NA & NA & NA & NA \\
fisher-ceval-valid & NA & NA & 51.49 & NA & NA & NA & NA \\
fisher-cmmlu & NA & NA & NA & 52.61 & NA & NA & NA \\
fisher-mmlu & NA & NA & NA & NA & 51.10 & NA & NA \\
fisher-minerva-math & NA & NA & NA & NA & NA & NA & NA \\
RAMPr-L 50 & 60.77 & 63.52 & 53.57 & 52.18 & 60.13 & 26.00 & 13.60 \\
RAMPr-D 50 & 60.22 & 56.26 & 56.76 & 58.00 & 56.97 & 22.40 & 39.20 \\
RAMPr-L 75 & 73.24 & 79.21 & 66.27 & 65.65 & 69.79 & 25.60 & 69.60 \\
RAMPr-D 75 & 71.74 & 75.08 & 65.82 & 68.08 & 69.37 & 49.40 & 78.20 \\
RAMPr-L 90 & 73.01 & 80.34 & 69.32 & 70.86 & 71.88 & 52.80 & 82.20 \\
RAMPr-D 90 & 73.01 & 80.00 & 68.35 & 71.48 & 71.54 & 54.00 & 78.60 \\
reap25 & 70.24 & 72.18 & 57.21 & 55.44 & 61.62 & 46.40 & 79.40 \\
\bottomrule
\end{tabularx}
\end{minipage}\hfill
\begin{minipage}[t]{0.485\linewidth}
\vspace{0pt}
\centering\scriptsize
\rowcolors{2}{tabrow}{}
\begin{tabularx}{\linewidth}{Lrrrrrr}
\toprule
\rowcolor{tabhead}Method & humaneval & gsm8k & MATH & GPQA & aime25 & aime26 \\
\midrule
base & 91.46 & 92.19 & 88.40 & 48.48 & 66.67 & 63.33 \\
fisher-gsm8k & 0.61 & 38.89 & 11.80 & 2.53 & 0.00 & 3.33 \\
fisher-mbpp & 14.02 & 20.17 & 13.00 & 0.51 & 0.00 & 0.00 \\
fisher-winogrande & NA & NA & NA & NA & NA & NA \\
fisher-hellaswag & NA & NA & NA & NA & NA & NA \\
fisher-ceval-valid & NA & NA & NA & NA & NA & NA \\
fisher-cmmlu & NA & NA & NA & NA & NA & NA \\
fisher-mmlu & NA & NA & NA & NA & NA & NA \\
fisher-minerva-math & NA & NA & 39.80 & NA & NA & NA \\
RAMPr-L 50 & 35.98 & 87.64 & 71.00 & 15.15 & 53.33 & 46.67 \\
RAMPr-D 50 & 68.29 & 77.26 & 70.80 & 7.07 & 50.00 & 40.00 \\
RAMPr-L 75 & 80.49 & 92.19 & 87.60 & 36.87 & 66.67 & 66.67 \\
RAMPr-D 75 & 90.24 & 92.42 & 87.60 & 42.93 & 60.00 & 63.33 \\
RAMPr-L 90 & 88.41 & 92.72 & 87.80 & 45.96 & 76.67 & 66.67 \\
RAMPr-D 90 & 87.80 & 93.25 & 88.20 & 49.49 & 76.67 & 63.33 \\
reap25 & 90.85 & 91.51 & 88.60 & 45.45 & 80.00 & 73.33 \\
\bottomrule
\end{tabularx}
\end{minipage}
\end{table*}
\begin{table*}[t]
\caption{Qwen1.5 main-table scores. Missing entries remain NA. Scores are percentages under the recorded task protocols.}
\centering
\setlength{\tabcolsep}{2pt}
\begin{minipage}[t]{0.485\linewidth}
\vspace{0pt}
\centering\scriptsize
\rowcolors{2}{tabrow}{}
\begin{tabularx}{\linewidth}{Lrrrrrrr}
\toprule
\rowcolor{tabhead}Method & Wino & Hella & CEval & cmmlu & mmlu & PubMed & mbpp \\
\midrule
base & 65.59 & 78.90 & 76.30 & 77.47 & 59.47 & 14.60 & 33.20 \\
fisher-gsm8k & 60.69 & 66.35 & 52.30 & 51.36 & 49.30 & 16.00 & 11.00 \\
fisher-mbpp & 61.96 & 64.81 & 51.71 & 52.58 & 48.59 & 13.40 & 21.00 \\
fisher-winogrande & 61.96 & NA & NA & NA & NA & NA & NA \\
fisher-hellaswag & NA & 72.75 & NA & NA & NA & NA & NA \\
fisher-ceval-valid & NA & NA & 57.65 & NA & NA & NA & NA \\
fisher-cmmlu & NA & NA & NA & 58.61 & NA & NA & NA \\
fisher-mmlu & NA & NA & NA & NA & 53.81 & NA & NA \\
fisher-minerva-math & NA & NA & NA & NA & NA & NA & NA \\
RAMPr-L 50 & 63.54 & 68.58 & 60.25 & 59.44 & 54.69 & 13.60 & 25.80 \\
RAMPr-D 50 & 63.46 & 69.89 & 61.00 & 61.53 & 55.34 & 14.00 & 26.80 \\
RAMPr-L 75 & 65.59 & 76.51 & 69.47 & 68.47 & 57.61 & 11.40 & 29.80 \\
RAMPr-D 75 & 65.59 & 75.99 & 71.62 & 70.66 & 58.37 & 12.00 & 33.60 \\
RAMPr-L 90 & 65.75 & 78.62 & 72.44 & 73.79 & 58.97 & 13.80 & 31.80 \\
RAMPr-D 90 & 65.98 & 78.68 & 74.00 & 75.40 & 59.50 & 15.80 & 31.40 \\
\bottomrule
\end{tabularx}
\end{minipage}\hfill
\begin{minipage}[t]{0.485\linewidth}
\vspace{0pt}
\centering\scriptsize
\rowcolors{2}{tabrow}{}
\begin{tabularx}{\linewidth}{Lrrrrrr}
\toprule
\rowcolor{tabhead}Method & humaneval & gsm8k & MATH & GPQA & aime25 & aime26 \\
\midrule
base & 43.90 & 19.94 & 22.60 & 13.13 & 0.00 & 0.00 \\
fisher-gsm8k & 4.27 & 26.00 & 11.60 & 9.60 & 0.00 & 0.00 \\
fisher-mbpp & 22.56 & 20.47 & 10.80 & 3.54 & 0.00 & 0.00 \\
fisher-winogrande & NA & NA & NA & NA & NA & NA \\
fisher-hellaswag & NA & NA & NA & NA & NA & NA \\
fisher-ceval-valid & NA & NA & NA & NA & NA & NA \\
fisher-cmmlu & NA & NA & NA & NA & NA & NA \\
fisher-mmlu & NA & NA & NA & NA & NA & NA \\
fisher-minerva-math & NA & NA & 14.40 & NA & NA & NA \\
RAMPr-L 50 & 28.05 & 41.77 & 16.80 & 3.03 & 0.00 & 0.00 \\
RAMPr-D 50 & 26.83 & 26.38 & 18.40 & 7.07 & 0.00 & 0.00 \\
RAMPr-L 75 & 39.02 & 27.60 & 21.60 & 9.09 & 0.00 & 0.00 \\
RAMPr-D 75 & 41.46 & 19.94 & 20.80 & 9.60 & 0.00 & 0.00 \\
RAMPr-L 90 & 42.07 & 17.59 & 20.80 & 12.12 & 0.00 & 0.00 \\
RAMPr-D 90 & 44.51 & 21.46 & 24.00 & 16.67 & 0.00 & 0.00 \\
\bottomrule
\end{tabularx}
\end{minipage}
\end{table*}
\begin{table*}[t]
\caption{OLMoE main-table scores. Missing entries remain NA. Scores are percentages under the recorded task protocols.}
\centering
\setlength{\tabcolsep}{2pt}
\begin{minipage}[t]{0.485\linewidth}
\vspace{0pt}
\centering\scriptsize
\rowcolors{2}{tabrow}{}
\begin{tabularx}{\linewidth}{Lrrrrrrr}
\toprule
\rowcolor{tabhead}Method & Wino & Hella & CEval & cmmlu & mmlu & PubMed & mbpp \\
\midrule
base & 67.72 & 80.75 & 37.44 & 36.40 & 53.77 & 60.60 & 28.80 \\
fisher-gsm8k & 56.51 & 53.89 & 29.35 & 28.01 & 36.16 & 32.40 & 0.80 \\
fisher-mbpp & 56.43 & 52.92 & 29.57 & 28.67 & 39.66 & 55.20 & 14.00 \\
fisher-winogrande & 60.77 & NA & NA & NA & NA & NA & NA \\
fisher-hellaswag & NA & 63.10 & NA & NA & NA & NA & NA \\
fisher-ceval-valid & NA & NA & 32.54 & NA & NA & NA & NA \\
fisher-cmmlu & NA & NA & NA & 32.48 & NA & NA & NA \\
fisher-mmlu & NA & NA & NA & NA & 43.14 & NA & NA \\
fisher-minerva-math & NA & NA & NA & NA & NA & NA & NA \\
RAMPr-L 50 & 59.04 & 60.11 & 36.26 & 33.22 & 45.58 & 40.60 & 24.20 \\
RAMPr-D 50 & 60.30 & 59.28 & 34.32 & 33.76 & 45.97 & 47.80 & 25.00 \\
RAMPr-L 75 & 64.17 & 77.09 & 36.40 & 36.19 & 51.13 & 51.00 & 28.80 \\
RAMPr-D 75 & 64.40 & 76.47 & 36.85 & 36.06 & 51.35 & 41.80 & 27.40 \\
RAMPr-L 90 & 67.40 & 80.17 & 36.63 & 36.46 & 53.21 & 51.80 & 28.80 \\
RAMPr-D 90 & 67.17 & 79.95 & 36.63 & 36.31 & 53.58 & 60.80 & 27.80 \\
\bottomrule
\end{tabularx}
\end{minipage}\hfill
\begin{minipage}[t]{0.485\linewidth}
\vspace{0pt}
\centering\scriptsize
\rowcolors{2}{tabrow}{}
\begin{tabularx}{\linewidth}{Lrrrrrr}
\toprule
\rowcolor{tabhead}Method & humaneval & gsm8k & MATH & GPQA & aime25 & aime26 \\
\midrule
base & 37.20 & 68.23 & 26.00 & 13.13 & 0.00 & 0.00 \\
fisher-gsm8k & 1.22 & 46.17 & 6.20 & 5.05 & 0.00 & 0.00 \\
fisher-mbpp & 17.07 & 26.46 & 6.00 & 10.61 & 0.00 & 0.00 \\
fisher-winogrande & NA & NA & NA & NA & NA & NA \\
fisher-hellaswag & NA & NA & NA & NA & NA & NA \\
fisher-ceval-valid & NA & NA & NA & NA & NA & NA \\
fisher-cmmlu & NA & NA & NA & NA & NA & NA \\
fisher-mmlu & NA & NA & NA & NA & NA & NA \\
fisher-minerva-math & NA & NA & 10.40 & NA & NA & NA \\
RAMPr-L 50 & 27.44 & 62.93 & 20.20 & 11.62 & 0.00 & 0.00 \\
RAMPr-D 50 & 26.22 & 59.89 & 19.40 & 6.57 & 0.00 & 0.00 \\
RAMPr-L 75 & 33.54 & 69.07 & 24.60 & 12.63 & 0.00 & 0.00 \\
RAMPr-D 75 & 34.76 & 67.85 & 20.80 & 9.09 & 0.00 & 3.33 \\
RAMPr-L 90 & 34.15 & 69.52 & 23.00 & 11.62 & 0.00 & 0.00 \\
RAMPr-D 90 & 32.32 & 68.46 & 22.20 & 13.64 & 0.00 & 0.00 \\
\bottomrule
\end{tabularx}
\end{minipage}
\end{table*}

\endgroup

\section{Evaluation Settings}
\label{app:evaluation-settings}
All task scores in Appendix~\ref{app:pruning-metrics} are obtained with lm-eval 0.4.13 on vLLM 0.29.0, in BF16 and with seed 1234.
The multiple-choice tasks, WinoGrande, HellaSwag, C-Eval, CMMLU, MMLU and PubMedQA, are scored by comparing the likelihoods of the answer options, so they involve neither sampling nor a generation budget.
The generation tasks apply each model's chat template, enabling thinking for the models that support it, and use the sampling settings and maximum output lengths in \cref{tab:eval-generation}.
Within each model, the original model and every pruned variant are evaluated under the same settings.

\begin{table}[!htbp]
\caption{Generation settings for the six models. Budgets are maximum output tokens. For Qwen3.5 and Qwen3.6, HumanEval and MBPP use temperature 0.6, top-$p$ 0.95 and top-$k$ 20 without a presence penalty. Qwen1.5 additionally uses the system prompt ``You are a helpful assistant.''}
\label{tab:eval-generation}
\centering\small
\setlength{\tabcolsep}{3.5pt}
\rowcolors{2}{tabrow}{}
\begin{tabularx}{\textwidth}{llLrrrrrr}
\toprule
\rowcolor{tabhead}Model & Thinking & Sampling & AIME & MATH & GPQA & GSM8K & HumanEval & MBPP \\
\midrule
Qwen3 & On & $T{=}0.6$, top-$p$ 0.95, top-$k$ 20 & 38,912 & 16,384 & 16,384 & 8,192 & 8,192 & 8,192 \\
Qwen3.5 & On & $T{=}1.0$, top-$p$ 0.95, top-$k$ 20, presence 1.5 & 81,920 & 16,384 & 16,384 & 8,192 & 8,192 & 8,192 \\
Qwen3.6 & On & $T{=}1.0$, top-$p$ 0.95, top-$k$ 20, presence 1.5 & 81,920 & 16,384 & 16,384 & 8,192 & 8,192 & 8,192 \\
GLM-4.7 & On & $T{=}1.0$, top-$p$ 0.95 & 32,768 & 16,384 & 16,384 & 8,192 & 8,192 & 8,192 \\
Qwen1.5 & Off & $T{=}0.7$, top-$p$ 0.8, top-$k$ 20, repetition 1.05 & 8,192 & 16,384 & 8,192 & 8,192 & 8,192 & 8,192 \\
OLMoE & Off & Greedy & 2,048 & 2,048 & 1,024 & 2,048 & 2,048 & 1,024 \\
\bottomrule
\end{tabularx}
\end{table}

\end{document}